\documentclass[runningheads]{llncs}

\usepackage{eccv}

\usepackage{eccvabbrv}

\usepackage{graphicx}
\usepackage{booktabs}
\usepackage{nicefrac}
\usepackage{wrapfig}
\usepackage{tikz}
\usepackage{multirow}
\usetikzlibrary{calc,spy,backgrounds}
\usepackage{adjustbox}
\usepackage{sidecap}
\usepackage{enumitem}
\usepackage{multirow}
\usepackage{xfrac}

\usepackage[dvipsnames]{xcolor}

\newcommand{\normal}{\mathbf{n}}

\newcommand{\light}{\ell}
\providecommand{\xpos}{\mathbf{x}}
\newcommand{\xposof}[1]{\mathbf{x}\lft(#1\rgt)}
\providecommand{\xorigin}{\mathbf{o}}

\newcommand{\superj}[1]{#1^{(j)}}
\newcommand{\superjk}[1]{#1^{(j, k)}}

\providecommand{\normal}{\mathbf{n}}

\providecommand{\direction}{\boldsymbol{\omega}}
\providecommand{\dirin}{\direction_\mathrm{i}}

\providecommand{\dirout}{\direction_\mathrm{o}}

\providecommand{\mixture}{q}

\providecommand{\ncount}{\mathrm{N}}
\providecommand{\scount}{\mathrm{S}}
\providecommand{\mcount}{\mathrm{M}}
\providecommand{\kcount}{\mathrm{K}}
\providecommand{\lcount}{\mathrm{L}}

\newcommand\lft{\mathopen{}\left}
\newcommand\rgt{\aftergroup\mathclose\aftergroup{\aftergroup}\right}

\usepackage[accsupp]{axessibility}  

\usepackage{hyperref}

\usepackage{orcidlink}

\begin{document}

\title{Flash Cache: Reducing Bias in Radiance Cache Based Inverse Rendering} 

\titlerunning{Flash Cache: Reducing Bias in Radiance Cache Based Inverse Rendering}

\author{Benjamin Attal\inst{1}\orcidlink{0000-0002-0132-5232} 
\and
Dor Verbin\inst{2}\orcidlink{0000-0001-8798-3270} 
\and
Ben Mildenhall\inst{2}\orcidlink{0009-0001-9796-6121} 
\and
Peter Hedman\inst{2}\orcidlink{0000-0002-2182-0185} 
\and
Jonathan T. Barron\inst{2}
\and
Matthew O'Toole\inst{1}\orcidlink{0000-0002-0740-9349} \and
Pratul P. Srinivasan\inst{2}\orcidlink{0009-0008-8268-3285}
}

\authorrunning{B.~Attal et al.}

\institute{$^1$Carnegie Mellon University \quad $^2$Google Research \quad \href{https://benattal.github.io/flash-cache/}{https://benattal.github.io/flash-cache/}}

\maketitle

\begin{abstract}
State-of-the-art techniques for 3D reconstruction are largely based on volumetric scene representations, which require sampling multiple points to compute the color arriving along a ray. Using these representations for more general inverse rendering --- reconstructing geometry, materials, and lighting from observed images --- is challenging because recursively path-tracing such volumetric representations is expensive. Recent works alleviate this issue through the use of radiance caches: data structures that store the steady-state, infinite-bounce radiance arriving at any point from any direction. However, these solutions rely on approximations that introduce bias into the renderings and, more importantly, into the gradients used for optimization. We present a method that avoids these approximations while remaining computationally efficient. In particular, we leverage two techniques to reduce variance for unbiased estimators of the rendering equation: (1) an occlusion-aware importance sampler for incoming illumination and (2) a fast cache architecture that can be used as a control variate for the radiance from a high-quality, but more expensive, volumetric cache. We show that by removing these biases our approach improves the generality of radiance cache based inverse rendering, as well as increasing quality in the presence of challenging light transport effects such as specular reflections.

\end{abstract}

\begin{figure*}[t]
\footnotesize

\newcommand{\teaserwidth}{1.3in}
\centering
\begin{tabular}{@{}cc|cc@{}}
& Ground-Truth Image & Predicted Image & Predicted $\sfrac{\text{Roughness}}{\text{Albedo}}$ \\
\rotatebox{90}{\quad\quad\quad\,\,\, \textit{pitcher}} & \adjincludegraphics[width=\teaserwidth]{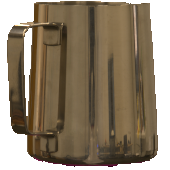} & \adjincludegraphics[width=\teaserwidth]{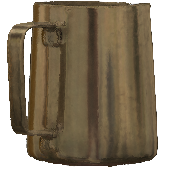} & \adjincludegraphics[width=\teaserwidth]{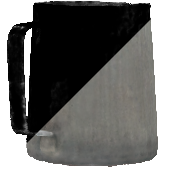} \\
\rotatebox{90}{\quad\quad\quad\quad\,\,\, \textit{cactus}} & \adjincludegraphics[width=\teaserwidth]{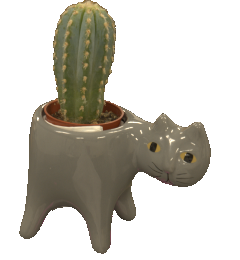} &
\adjincludegraphics[width=\teaserwidth]{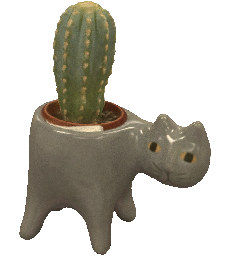} &
\adjincludegraphics[width=\teaserwidth]{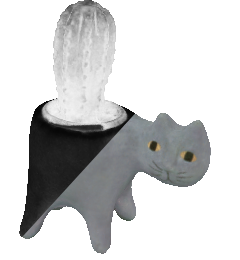}
\end{tabular}
\vspace{-0.1in}
\caption{%
Results on two captured scenes from the Stanford ORB dataset~\cite{StanfordOrb}. Our model synthesizes realistic novel views of real-world objects that exhibit complicated specular materials, and recovers accurate material roughnesses (black represents a perfect mirror, white is perfectly rough) and albedo maps.
}%
\label{fig:teaser}
\vspace{-5mm}
\end{figure*}

\section{Introduction}
\label{sec:intro}

Volume rendering has been remarkably effective for novel view synthesis and 3D reconstruction, as demonstrated by approaches such as Neural Radiance Fields (NeRFs)~\cite{MildeSTBRN2020} and Gaussian Splatting~\cite{kerbl20233d}. However, it is difficult to use volume rendering for the more general inverse rendering problem of recovering materials and lighting in addition to geometry. The issue is largely computational: volume rendering techniques generally query the appearance and density of the volume at many sample points to compute the light arriving along each ray. Combining this with physically-accurate rendering models that simulate global illumination (instead of simple shading models that only consider direct lighting) dramatically increases computational complexity, because evaluating the rendering integral requires recursively casting additional scattered rays from each sample location.

Recent methods~\cite{jin2023tensoir, zhang2022modeling} have leveraged radiance caches to simulate global illumination tractably within an inverse rendering optimization pipeline. Instead of integrating over the domain of $n$-bounce light paths, radiance caches store a representation of the steady-state light arriving at any point in the volume from any direction. This is a powerful paradigm that removes the need for recursively evaluating the rendering integral by considering a single reflection of the incoming steady-state (as opposed to direct) light at any point on the scene geometry towards the direction of the viewer.

Existing works that combine radiance caches with volumetric inverse rendering fall into two categories. The first uses trained NeRF models as radiance caches, and volume renders the NeRF along rays from points within the scene to estimate incoming light. This type of radiance cache has high quality but tends to be computationally expensive to evaluate; methods that use NeRF as a radiance cache, such as TensoIR~\cite{jin2023tensoir}, only compute the reflection integral at a single point (the expected ray termination) along each camera ray. The second category relies on cheap radiance caches such as a small MLPs, as in InvRender~\cite{zhang2022modeling} and NeRO~\cite{liu2023nero}, to evaluate incoming lighting with a single ray query. Though fast, these caches fail to capture high-frequency and near-field illumination. We observe that both of these existing approaches introduce bias into the pipeline: the first category renders a NeRF model at only a single evaluation along each camera ray, which results in a biased estimation of the volume rendering integral, and the second category approximates incoming light with a low-capacity model, which results in an inaccurate and biased estimation of the reflection integral.

We present a system for volumetric inverse rendering based on the basic goal of avoiding introducing bias into the rendering procedure as much as possible, while leveraging variance reduction techniques to keep the rendering model computationally tractable. This is made possible by the following key technical contributions:
\begin{itemize}[noitemsep, topsep=0pt, partopsep=0pt]
    \item A strategy for efficiently estimating the light reflected by volumetric geometry along each camera ray by using a neural field to parameterize a spatially-varying mixture of von-Mises Fisher distributions for light importance sampling. This captures spatially-varying illumination effects such as light occlusion and interreflections and lets our system efficiently estimate the rendering equation with fewer queries of the radiance cache. 
    \item An architecture for a lightweight high-capacity radiance cache that can query the incoming illumination from any direction at any point within the volume, along with a strategy for using this cache as a control variate to avoid introducing rendering bias. This lets us efficiently model high-frequency and near-field illumination without requiring the full volume rendering necessary when using NeRF as a radiance cache. 
\end{itemize}

\noindent We demonstrate that our system recovers improved geometry and materials over prior work, and in particular enables more accurate recovery of material parameters in regions affected by near-field lighting effects such as specular interreflections. Furthermore, since our system does not use a simplified model of volumetric geometry, it preserves the advantages of NeRF-like volumetric representations for capturing thin structures. We validate the benefits of our system with evaluation on existing benchmarks and rendered scenes that exhibit challenging lighting effects.

\section{Related Work}
\label{sec:related}

\paragraph{Inverse rendering.} Inverse rendering aims to recover scene attributes like materials, lighting, and geometry, from a set of images~\cite{ramamoorthi2001signal,yu1999inverse,Barron2013A,debevec2008rendering}. A common approach
is to use a physically-based model of light transport~\cite{pharr2023physically,veach1998robust,kajiya1986rendering} and differentiable rendering so that it is possible to leverage gradient-based optimization for decomposing the scene's appearance into its constituent parts~\cite{chen2019learning,li2018differentiable,Jakob2020DrJit}. 
Recent advances in view synthesis demonstrate the benefit of volume rendering neural field representations for the task of reproducing realistic images of a captured scene from unseen views~\cite{MildeSTBRN2020,VerbiHMZBS2022,barron2023zipnerf}. Neural field representations also improve inverse rendering methods, combining ideas from volume rendering to represent geometry and physically-based rendering to represent appearance~\cite{srinivasan2021nerv,BiXSMSHHKR2020,Munkberg_2022_CVPR,hasselgren2022nvdiffrecmc,Mai_2023_ICCV}.

\paragraph{Variance-reduction for inverse rendering.} Among the different paradigms for inverse rendering using neural fields, cache-based inverse rendering is the most closely related to ours. These methods rely on using a radiance cache for storing the incoming radiance at every point and along every direction. These radiance caches capture global illumination effects, replacing a prohibitively expensive recursive path tracing procedure with a much cheaper cache look-up. This technique has been used extensively in the past for computing indirect illumination and global illumination effects~\cite{ward1988ray,krivanek2022practical}, and consequently used for real-time rendering applications~\cite{scherzer2012pre,silvennoinen2017real,muller2021real}.

In the context of inverse rendering, radiance caches are used for storing the hemisphere of incoming radiance at every point. This hemisphere is then integrated against the local BRDF to yield outgoing radiance, which is optimized to match the observed image pixels. NeRFactor~\cite{zhang2021nerfactor} only considers direct illumination effects for decomposing lights and materials, but caches the visibility of each secondary ray into an MLP to model self occlusions. Neural Incident Light Field (NeILF)~\cite{yao2022neilf} and Neural Reflective Objects (NeRO)~\cite{liu2023nero} and InvRender~\cite{zhang2022modeling} parameterize the radiance cache using an MLP modeling the incident light field. However, they only supervise the outgoing radiance, which does not enforce consistency between the radiance cache and incoming radiance. An improved version of NeILF~\cite{zhang2023neilf++} solves this by penalizing differences between the radiance cache and the corresponding outgoing radiance. These methods all use inexpensive caches which tend to fail to capture high-frequency and near-field illumination. Most recently, concurrent work to ours from Ling \etal~\cite{ling2024nerf} uses a NeRF-based cache, along with an optimizable mixture of Gaussians model for importance sampling the 3D distribution of light sources in the scene.
Our work also uses importance sampling to reduce Monte Carlo noise, but our distributions account for occlusions --- a key part of variance reduction~\cite{veach1998robust}.
Along with importance sampling, we further reduce noise using control variates, two techniques that have been extensively explored in the context of forward rendering~\cite{bitterli20spatiotemporal,mueller2019neural,nicolet2023recursive}.

\section{Preliminaries}

In this work, we consider the problem of inverse rendering of geometry, materials, and lighting from a set of input images. We leverage a combination of volumetric rendering and physically-based Monte Carlo rendering in order to accomplish this task. In particular, we optimize the parameters $\mathrm\Phi$ of a scene representation using a regularization term $\mathcal{L}_{\textit{reg}}$ and a loss term such that the color value $L_\mathrm{i}(\xorigin, \dirout ; \mathrm\Phi)$ incident at a pixel corresponding to a camera ray $(\xorigin, \dirout)$ --- where $\xorigin$ is its origin and $\dirout$ its direction --- matches with the corresponding measured image pixel $I(\xorigin, \dirout)$:
\begin{align}
    \min_{\mathrm\Phi} \sum_{\xorigin, \dirout} \lVert I(\xorigin, \dirout) - L_\mathrm{i}(\xorigin, \dirout ; \mathrm\Phi) \rVert ^ 2 + \mathcal{L}_{\textit{reg}}(\mathrm\Phi)
    \label{eqn:diff-ren}
\end{align}

\subsection{Volume Rendering and Neural Radiance Fields}
\label{sec:neural-volumes}



In the volume rendering setting, the incident radiance $L_\mathrm{i}$ of a ray $(\xorigin, \dirout)$ passing through a neural volume, is computed by integrating the differential outgoing radiance $L_\mathrm{o}$ at every point along that ray:
\begin{align}
    L_\mathrm{i}(\xorigin, \dirout) = \int_{0}^{\infty} L_\mathrm{o}(\xposof{t}, \dirout) T\lft(\xorigin, \xposof{t}\rgt) \sigma(\xposof{t}) dt,
    \label{eqn:volume-rendering}
\end{align}
\noindent where $T\lft(\xorigin, \xposof{t}\rgt) = \exp( -\int_{0}^{t} \sigma(\xposof{s})\, ds )$ denotes the transmittance from $\xorigin$ to $\xposof{t}$, and $\xposof{t}=\xorigin-t\dirout$ is a point along the ray parameterized by $t$. This integral is generally intractable to evaluate analytically, but it can be approximated via quadrature~\cite{max1995optical}, by sampling density and radiance at a set of points $\{\xposof{t_k}\}_{k=1}^\ncount$:
\begin{align}
    \hat{L}_\mathrm{i}(\xorigin, \dirout) &= \sum_{k = 1}^{\ncount} w_k \,
    L_\mathrm{o}\lft(\xposof{t_k}, \dirout \rgt), \label{eqn:quadrature} \\
    w_k &= \lft(1 - e^{-\sigma(\xposof{t_k}) (t_{k+1} - t_k)} \rgt)\exp\lft( - \sum_{j=1}^{k - 1} \sigma(\xposof{t_j})(t_{j+1} - t_j)\rgt)\,. \nonumber
\end{align}

In NeRF~\cite{MildeSTBRN2020}, the volume density $\sigma$ and outgoing radiance $L_\mathrm{o}$ are parameterized using a multi-layer perceptron (MLP). Follow-up work~\cite{chen2022tensorf,fridovich2022plenoxels,MuelleESK2022} relies on optimizable grids or hash tables along with smaller neural networks. In both cases, a function maps every position $\xpos$ and direction $\dirout$ into the volume density $\sigma(\xpos)$ at that location and the outgoing radiance from it in the specified direction $L_\mathrm{o}(\xpos, \dirout)$. Together these give the radiance along the ray using Equation~\ref{eqn:quadrature}.

\subsection{Physics-Based Monte Carlo Light Transport}

We use ideas from physically-based rendering for combining incoming radiance $L_\mathrm{i}$ with a BRDF $f$ to produce outgoing radiance $L_\mathrm{o}$ using the rendering equation:
\begin{align}
    L_\mathrm{o}(\xposof{t}, \dirout) =  \int_{\mathrm{\Omega}} f\lft(\xposof{t}, \dirin, \dirout\rgt) L_\mathrm{i}\lft(\xpos, \dirin\rgt) \lft(\mathbf{n} \cdot \dirin\rgt)\, d \dirin\,,
    \label{eqn:rendering-equation}
\end{align}
where $\mathrm{\Omega} = \{\dirin \colon \mathbf{n}\cdot\dirin > 0\}$ is the positive hemisphere with respect to normal $\normal$.

Like the volume rendering integral in Equation~\ref{eqn:volume-rendering}, evaluating the rendering equation analytically is, in general, intractable. Most approaches therefore use Monte Carlo integration in order to approximate Equation~\ref{eqn:rendering-equation}. In particular, if we randomly sample $\mcount$ incoming directions from some distribution $p$ over incoming directions, $\dirin^{(1)}, \ldots, \dirin^{(\mcount)} \sim p\lft(\dirin\rgt)$,
then we can construct an unbiased estimator:
\begin{align}
    \hat{L}_\mathrm{o}(\xposof{t}, \dirout) =  \frac{1}{\mcount} \sum_{j=1}^{\mcount}  f\lft(\xposof{t}, \superj{\dirin}, \dirout \rgt) L_\mathrm{i}\lft(\xposof{t}, \superj{\dirin}\rgt) \lft(\mathbf{n} \cdot \superj{\dirin} \rgt) \lft/ p\lft(\superj{\dirin}\rgt) \rgt.,
    \label{eqn:rendering-estimator}
\end{align}
where performing optimization using the estimator in Equation~\ref{eqn:rendering-estimator} is, in expectation, equivalent to optimizing using Equation~\ref{eqn:rendering-equation}~\cite{kloek1978bayesian}. 

We can now plug the Monte Carlo-based estimator in Equation~\ref{eqn:rendering-estimator} into the volume rendering estimator of Equation~\ref{eqn:quadrature} to arrive at the combined estimator:
\begin{equation}
\!\!\!\resizebox{0.94\linewidth}{!}{$\displaystyle
    \hat{L}_\mathrm{i}(\xorigin, \dirout) = \sum_{k = 1}^{\ncount} \frac{w_k}{\mcount} \sum_{j=1}^{\mcount} f\lft(\xposof{t_k}, \superjk{\dirin}, \dirout\rgt) \hat{L}_\mathrm{i}\lft(\xposof{t_k}, \superjk{\dirin}\rgt) \lft(\mathbf{n} \cdot \superjk{\dirin}\rgt) \lft/ p\lft(\superjk{\dirin}\rgt) \rgt.
    $}
\label{eqn:general-model}
\end{equation}

Unfortunately, Equation~\ref{eqn:general-model} requires recursive evaluation of both the volume rendering integral and the rendering equation, since $\hat{L}_\mathrm{i}$ is now defined with respect to itself. If one is not careful, the recursive nature of the rendering equation can quickly lead to a rapid increase in computational cost.

To deal with this issue, we return to the discussion in Section~\ref{sec:neural-volumes}. It is possible to define a neural volume that \textit{directly} predicts outgoing radiance, without requiring recursive evaluation. In other words, we can replace the incoming radiance $\hat{L}_\mathrm{i}$ with the output of a neural radiance field $\hat{L}_\mathrm{i}^{\mathit{cache}}$, which relates the incoming and outgoing radiance according to Equation~\ref{eqn:quadrature}.
It is worth remarking that for this to be useful, evaluating the radiance cache $\hat{L}_\mathrm{i}^{\mathit{cache}}$ for a given ray should be cheaper than performing recursive path tracing.

\begin{figure}[t]
\centering
\includegraphics[width=1\linewidth, trim={0mm 8.7in 0.4in 0mm}, clip]{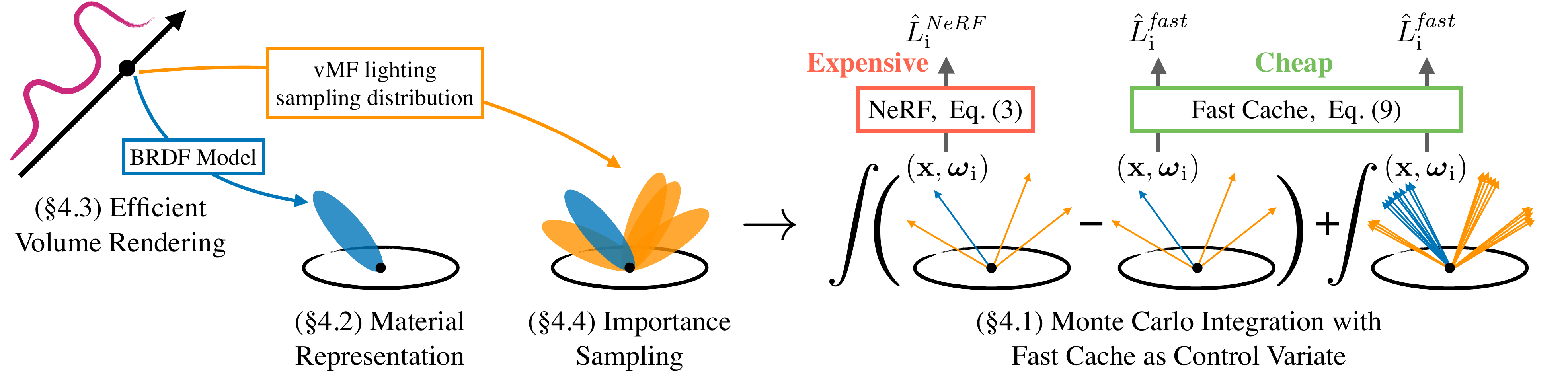}
\caption{
Our pipeline for physically-based inverse rendering combines volume rendering with physically-based Monte Carlo rendering to recover geometry, materials, and lighting. We first use an estimator (Equation~\ref{eqn:fast-volume-render}) to approximate the volume rendering integral (Equation~\ref{eqn:volume-rendering}) with only a single sample point per ray. Given this sample point, we use a spatially-varying mixture-of-vMFs (Section~\ref{sec:method-low-variance}) to importance-sample directions based on the distribution of incoming radiance. We then produce a low-variance estimate of the rendering equation (Equation~\ref{eqn:rendering-equation}) using a NeRF as a radiance cache, and a fast cache approximation of the NeRF (Equation~\ref{eqn:fast-cache}) that is far less expensive to evaluate.
}
\label{fig:pipeline}
\end{figure}

\section{Method}

Our inverse rendering method is based on a volumetric rendering approach to physically-based rendering, as described by Equation~\ref{eqn:general-model}. However, naïvely using the rendering model of Equation~\ref{eqn:general-model} is extremely expensive for two main reasons. First, it requires volumetrically rendering the incoming radiance at every point along the ray. Second, for every such point we need to use Monte Carlo integration, which typically requires many samples of incoming radiance. Our work aims to resolve both issues without introducing bias in the process, which may adversely affect the recovery of materials and lighting.

In this section, we present the design of our system. We begin by describing the incoming radiance cache in Section~\ref{sec:method-cache}, along with a novel ``fast cache'' architecture that is cheap to evaluate. Designing this cache as a control variate for incoming illumination allows us to use many incoming rays, which increases the accuracy of our method without requiring significant compute. In Section~\ref{sec:method-representation}, we discuss our physically-based material representation. Finally, in Sections~\ref{sec:method-volume} and \ref{sec:method-low-variance}, we present our 
efficient sampling schemes based on a simple unbiased estimator for the volumetric rendering integral, and a novel optimizable occlusion-aware spatially-varying importance sampling scheme.

\subsection{Incoming Radiance Cache}
\label{sec:method-cache}

\begin{figure}[t]
\centering
\includegraphics[width=1\linewidth, trim={0mm 1.55in 0in 0mm}, clip]{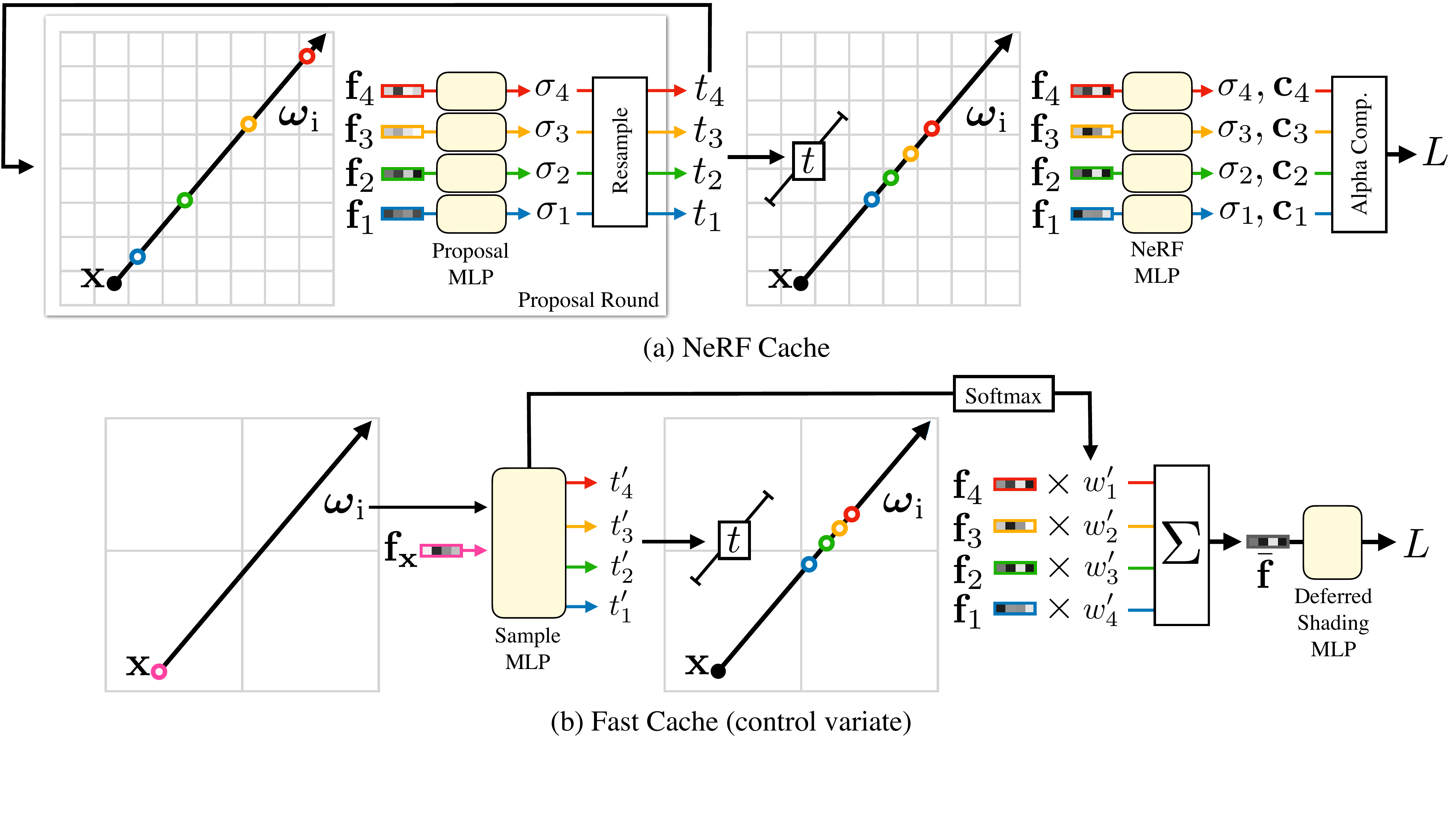}
\vspace{-0.2in}
\caption{
An illustration of our two radiance cache architectures. (a) The NeRF-based cache, which is accurate yet expensive to evaluate, uses proposal MLPs to predict sample point distances $\{t_k\}$ through an iterative resampling procedure. These distances are then used to generate sample points at which we query density and outgoing radiance. These quantities are combined to approximate the volume rendering integral via quadrature~\ref{eqn:quadrature}. (b) Rather than predict sample point distances via proposal sampling, our fast cache directly outputs distances $\{t'_k\}$ as well as weights $\{w'_k\}$ given incoming ray parameters $(\xpos, \dirin)$. The distances are used to generate sample points at which we query features from an NGP grid. The resulting features are averaged together, and then fed through a deferred shading MLP which predicts the final color for the incoming ray. Though our illustration depicts four samples in both cases, the NeRF-based cache (a) uses $64$ samples for the first two resampling rounds and $32$ samples for the final one, while our fast cache only uses $8$ samples overall.
}
\label{fig:architecture}
\end{figure}

Our incoming radiance cache is based on the architecture of Zip-NeRF~\cite{barron2023zipnerf}, and our appearance model is based on Ref-NeRF~\cite{VerbiHMZBS2022}. In particular, we borrow Zip-NeRF's usage of a multi-resolution hash grid as a neural graphics primitive (NGP)~\cite{MuelleESK2022}. See Figure~\ref{fig:architecture}(a) for an illustration of the NeRF-based radiance cache, and see a complete description of it in the supplement.






In order to increase the efficiency of our method, we wish to use an accelerated cache for querying incoming radiance instead of relying on an expensive NeRF-based cache which requires evaluating many samples per incoming ray. We therefore design a ``fast cache'' and supervise it to match the incoming radiance of the expensive, NeRF-based cache. See Figure~\ref{fig:architecture} for a diagram comparing the two caches.

\begin{figure*}[t]
\footnotesize

\newcommand{\qualwidth}{0.98in}
\begin{tabular}{@{}cc@{\!}c@{\!}c@{}}
Camera View & NeRF Cache & Fast Cache & vMFs \\
\multirow{2}{1.6in}[5ex]{
\includegraphics[width=1.5in]{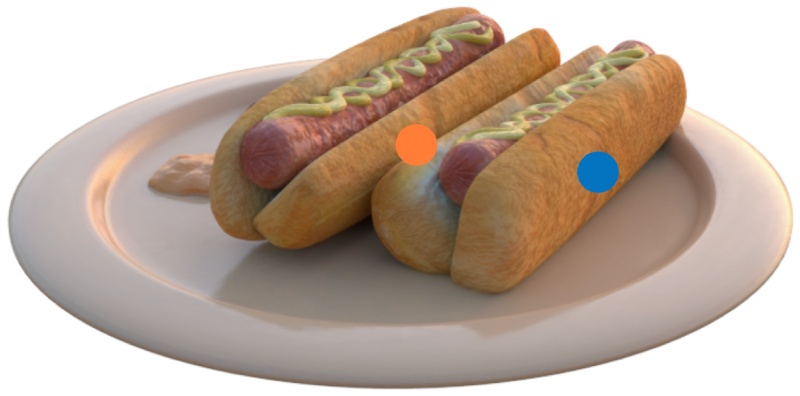} \vspace{1in}
}
&
\adjincludegraphics[width=\qualwidth,cframe=Orange 3pt]{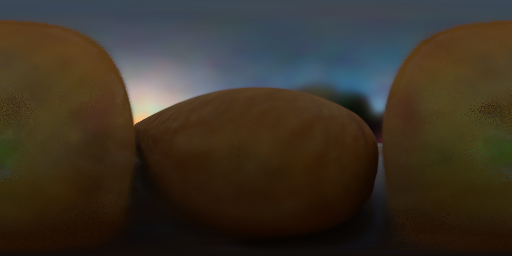}
&
\adjincludegraphics[width=\qualwidth,cframe=Orange 3pt]{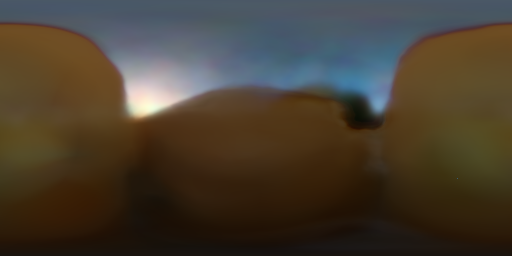}
&
\adjincludegraphics[width=\qualwidth,cframe=Orange 3pt]{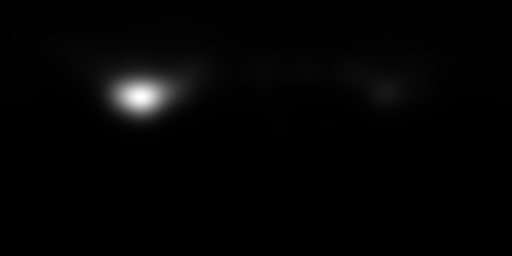}\\[-0.1mm]
&
\adjincludegraphics[width=\qualwidth,cframe=RoyalBlue 3pt]{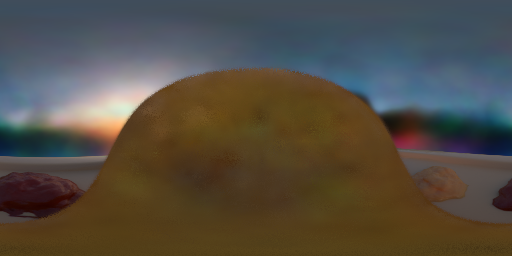}
&
\adjincludegraphics[width=\qualwidth,cframe=RoyalBlue 3pt]{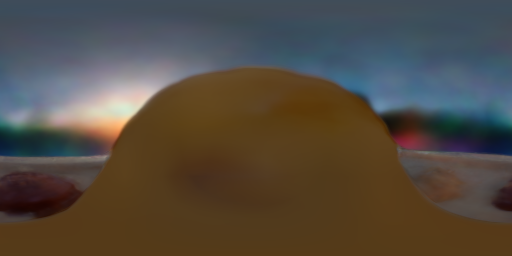}
&
\adjincludegraphics[width=\qualwidth,cframe=RoyalBlue 3pt]{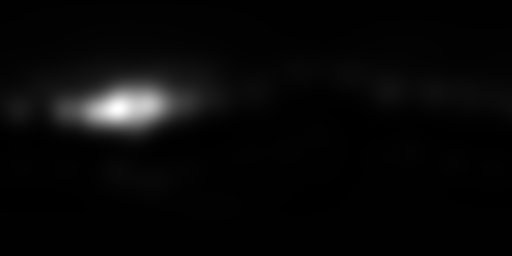}
\end{tabular}
\vspace{-0.1in}
\caption{%
\textit{(left)} We choose two points on the hotdog for which to visualize distributions of incoming radiance. \textit{(right)} We show (a) incoming radiance from the NeRF cache, which is high quality but slow to evaluate, (b) incoming radiance from the fast cache, which is slightly lower quality but fast to evaluate, and (c) the probability density of the spatially-varying mixture of vMFs, which is able to capture occlusions in the incoming radiance distribution.
}%
\label{fig:incoming}
\end{figure*}

Our fast cache works by first passing the incident ray's origin and direction ($\xpos$ and $\dirin$) through an optimizable function $g_{\operatorname{sample}}$, implemented as a low-resolution hash-encoding followed by a small sample MLP. This sampling function returns a set of sample point distances along the ray and their corresponding weights:
\begin{align}
    \{t'_k, w'_k \}_{k=1}^\scount\ = g_{\operatorname{sample}} (\xpos, \dirin).
\end{align}
Each distance $t'_k$ corresponds to a new sample point $\mathbf{x}'_k = \xpos + t'_k \dirin$ along the incident ray. Each of these samples is passed through a separate low-resolution NGP to yield a set of corresponding features $\{\mathbf{f}_k\}_{k=1}^\scount$, which are combined with their corresponding normalized (via a softmax) weights $w'_k$ to yield a ray feature vector which encodes the appearance of that ray. This feature is then decoded into the incident radiance by using a small MLP, similar to the ``deferred shading'' approach of Hedman~\etal~\cite{HedmaSMBD2021}:
\begin{align}
    \hat{L}_{\mathrm{i}}^{\mathit{fast}}  = \operatorname{MLP}_{\operatorname{color}}\lft( \sum_{k=1}^{\scount} w'_k \mathbf{f}_k \rgt)\,.
    \label{eqn:deferred-color}
\end{align}
Even for a small number of samples (we use $\scount = 8$ for all experiments), we find that this design serves as an effective cache for incoming radiance that captures high-frequency near-field illumination. See Figure~\ref{fig:architecture} for an overview of the NeRF-based and fast caches, as well as figure~\ref{fig:incoming} for an example of the output of the fast cache. Our supplement contains more details about both models.

Instead of simply replacing the NeRF-based cache with our fast cache, we combine the accuracy of the NeRF-based cache with the efficiency of the fast cache by treating the 
fast cache as a control variate~\cite{kalos2009monte, nicolet2023recursive} for incoming radiance. 
In order to do this, we sample two sets of incident directions using an importance sampling distribution $p$ (to be introduced in Section~\ref{sec:method-low-variance}): $\mcount'$ samples $\{\superj{\dirin'}\}_{j=1}^{\mcount'}$, and $\mcount\ll\mcount'$ additional independent samples $\{\superj{\dirin}\}_{j=1}^{\mcount}$. We then use the $\mcount'$ first samples to evaluate the fast cache $\hat{L}_\mathrm{i}^{\mathit{fast}}(\xpos, \dirin)$ via Equation~\ref{eqn:deferred-color}, and the
other $\mcount$ samples to evaluate the error between the NeRF-based cache $\hat{L}_\mathrm{i}^{\mathit{NeRF}}(\xpos, \dirin)$ and the fast cache. We then add the two estimates to arrive at an unbiased estimator for the outgoing radiance:
\!\!\!\!\!\!\!\!\!\!
\begin{align}
    \hat{L}_\mathrm{o}^{\mathit{fast}} &= \frac{1}{\mcount'} \sum_{j=1}^{\mcount'} f\lft(\xpos, \superj{\dirin'}, \dirout\rgt) \hat{L}_\mathrm{i}^{\mathit{fast}}\lft(\xpos, \superj{\dirin'}\rgt) \lft(\mathbf{n} \cdot \superj{\dirin'}\rgt) \lft/ p\lft(\superj{\dirin'}\rgt) \rgt., \nonumber \\
    \mathrm{\Delta}\hat{L}_\mathrm{o} &= \frac{1}{\mcount} \sum_{j=1}^{\mcount} f\lft(\!\xpos, \superj{\dirin}\!, \dirout\!\rgt) \lft(\!\hat{L}_\mathrm{i}^{\mathit{NeRF}}\!\lft(\!\xpos, \superj{\dirin}\!\rgt) \!-\! \hat{L}_\mathrm{i}^{\mathit{fast}}\!\lft(\!\xpos, \superj{\dirin}\!\rgt)\!\rgt)\! \lft(\mathbf{n} \cdot \superj{\dirin}\rgt) \!\lft/ \!p\lft(\!\superj{\dirin}\rgt)\rgt., \nonumber \\
    \hat{L}_\mathrm{o} &= \hat{L}_\mathrm{o}^{\mathit{fast}} + \mathrm{\Delta}\hat{L}_\mathrm{o} \,.
    \label{eqn:fast-cache}
\end{align}
Note that this requires $\mcount$ evaluations of the NeRF-based cache and $\mcount+\mcount'$ evaluations of the fast cache. Given sufficiently accurate cache (see Section~\ref{sec:results} for an example) that is indeed cheap to evaluate, this process results in the quality of using $\mcount'$ incoming rays from the accurate NeRF-based cache, by only requiring a fraction of the computation.




On top of the two caches, we also have an explicit model for far-field illumination based on an MLP,
$\hat{L}_{\mathrm{i}}^{\mathit{env}}(\dirin)  = \operatorname{MLP}_{\operatorname{env}}\lft( \dirin\rgt)$,
which is multiplied by the transparency of the ray $1-\sum_{k=1}^{\mathrm{N}} w_k$ and added to the radiance of the cache.

\subsection{Material Representation}
\label{sec:method-representation}




Like Ref-NeRF~\cite{VerbiHMZBS2022}, we extract analytic normals using the negative gradient of the density field, and leverage ``predicted normals'' output by the NGP, which are tied to the analytic normals using an L2 loss. Our material model is based on the Disney-GGX BRDF~\cite{burley2012physically}, which is combined with the predicted normals to obtain the radiance value as in Equation~\ref{eqn:rendering-equation}. The spatially-varying parameters of the BRDF model are output
by a separate NGP, $g_{\operatorname{material}}$:
\begin{align}
    \Big(m(\xpos), r(\xpos), \mathbf{a}(\xpos)\Big) = g_{\operatorname{material}}\lft(\xpos\rgt)\,,
\end{align}
where $m$ is metalness, $r$ is roughness, and $\mathbf{a}$ is albedo.


\subsection{Efficient Gradients for Volume Rendering}
\label{sec:method-volume}

Evaluating the volume rendering integral using the quadrature technique in Equation~\ref{eqn:quadrature} means querying the appearance and geometry of the volume at $\ncount$ sample point locations per ray, each of which requires Monte Carlo integration. 
However, this can be accelerated considerably by replacing the sum with a Monte Carlo estimate along the ray using $\kcount\ll \ncount$ samples~\cite{gupta2023mcnerf}. We draw $\kcount$ indices from a categorical distribution $j_1, \ldots, j_\kcount \sim \operatorname{Cat}(w_1, \ldots, w_\ncount)$, and replace Equation~\ref{eqn:quadrature} with:
\begin{align}
    \hat{L}_\mathrm{i}(\xorigin, \dirout) = \frac{1}{\kcount} \sum_{k=1}^{\kcount} L_\mathrm{o}\lft(\xposof{t_{j_k}}, \dirout\rgt) \,.
    \label{eqn:fast-volume-render}
\end{align}

The estimator of Equation~\ref{eqn:fast-volume-render} is unbiased, \ie, its expected value equals the quadrature value of Equation~\ref{eqn:quadrature}. While this still requires querying volume density for $\ncount$ points, we only need to evaluate the costly physically-based appearance at $\kcount$ points. Furthermore, if the render weight distribution has a single peak (often the case in practice), then this estimator has low variance, even for $\kcount = 1$.


\subsection{Low-Variance Gradients for Monte Carlo Rendering}
\label{sec:method-low-variance}

\begin{SCfigure}[50][t]
\caption{
An illustration of our occlusion-aware importance sampler. We use an NGP to predict the parameters of a mixture of vMFs distribution for each point in space (Equation~\ref{eqn:vmfs-ngp}), which can produce different lobes for different light sources. Because the parameters of these distributions are \textit{spatially-varying}, if a light source is occluded from the perspective of one surface point, then the model can ``turn off'' the lobe for that light source. 
}
\includegraphics[width=0.35\linewidth]{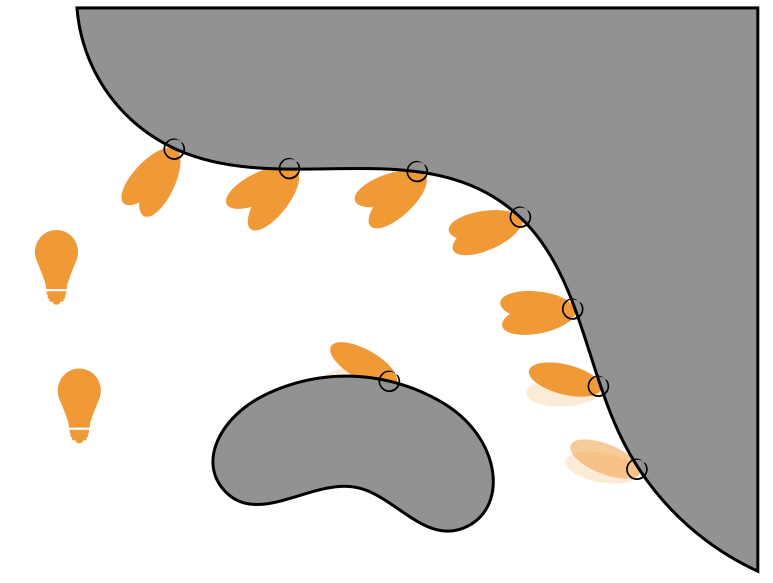}
\label{fig:vmf}
\end{SCfigure}

The unbiased estimator in Equation~\ref{eqn:fast-volume-render} reduces the number of secondary ray samples from $\ncount\cdot \mcount$ to $\kcount \cdot \mcount$. However, a large number of secondary rays $\kcount \cdot \mcount$ may be necessary for convergence, but may be prohibitively expensive. We reduce the number of samples by leveraging occlusion-aware importance sampler.

Importance sampling for incoming radiance is common technique in physically-based forward rendering and inverse rendering. Here, we aim to build a lightweight importance sampler that is capable of approximating the true distribution of incoming radiance, including occlusions, from both direct and indirect light sources. For this purpose, we use an NGP
to map a point $\xpos$ into a set of $5\lcount$ scalars representing the parameters $\lft\{ \boldsymbol{\mu}'_\light, \kappa_\light, \lambda_\light \rgt\}_{\light=1}^\lcount$ which parameterize a mixture of von Mises-Fisher (vMF) distributions as:
\begin{align}
    \mixture(\direction; \xpos) = \frac{1}{Z}\sum_{\light=1}^\lcount \lambda_\ell(\xpos) \operatorname{vMF}\lft(\direction; \boldsymbol{\mu}_\light(\xpos), \kappa_\light(\xpos)\rgt), \; \text{with}\; \boldsymbol{\mu}_\light(\xpos) = \frac{\boldsymbol{\mu}_\light'(\xpos) - \xpos}{\| \boldsymbol{\mu}_\light'(\xpos) - \xpos \|}\,,
    \label{eqn:vmfs-ngp}
\end{align}
where $Z = \sum_{\light=1}^\lcount \lambda_\light(\xpos)$ is the partition function used for normalizing $\mixture$, and the coefficients $\{\lambda_\light\}_{\light=1}^\lcount$ and $\{\kappa_\light\}_{\light=1}^\lcount$ are nonnegative numbers representing the mixture weights and vMF concentration parameters respectively.

The parameterization of the means $\lft\{ \boldsymbol{\mu}'_\light \rgt\}_{\light=1}^\lcount$ in Equation~\ref{eqn:vmfs-ngp} is designed so that a single vMF lobe can be assigned to the location of single (direct or indirect) light source in 3D, which is then projected onto the sphere. 

We optimize the parameters describing the distribution $\mixture$ at every location to match the true distribution of incoming radiance, taking occlusions into account. Concretely, we use the same importance sampling scheme used for evaluating the rendering integral in Equation~\ref{eqn:rendering-estimator}, to minimize the loss:
\begin{equation}
    \mathcal{L}_{\operatorname{vMF}} = \frac{1}{\mcount}\sum_{j=1}^\mcount \lft\| Z\cdot \mixture\lft(\superj{\dirin}; \xpos\rgt) - \bar{L}\lft(\xpos, \superj{\dirin}\rgt) \rgt\|^2 \lft/ p\lft(\superj{\dirin}\rgt)\rgt.,
\end{equation}
where $p$ is a distribution used to generate the samples $\{\superj{\dirin}\}_{j=1}^\mathrm{M}$ (as well as the $\mathrm{M}'$ cheap cache samples used in Section~\ref{sec:method-cache}), which is based the mixture distribution $q$ and a standard material-based importance sampling distribution (see supplement for more details), and $\bar{L}(\xpos, \dirin) = \|\hat{L}_\mathrm{i}^{\mathit{NeRF}}(\xpos, \dirin)\|_2$ is the L2 norm of the incident radiance, taken across channels.


\begin{figure}[t!]
    \centering
    \begin{tabular}{@{}c@{\hspace{3pt}}c@{\hspace{3pt}}c@{}}
        Fast Cache \& vMFs & vMFs & Neither \\
         \includegraphics[trim={10cm 8cm 6cm 13cm},clip,width=0.327\linewidth]{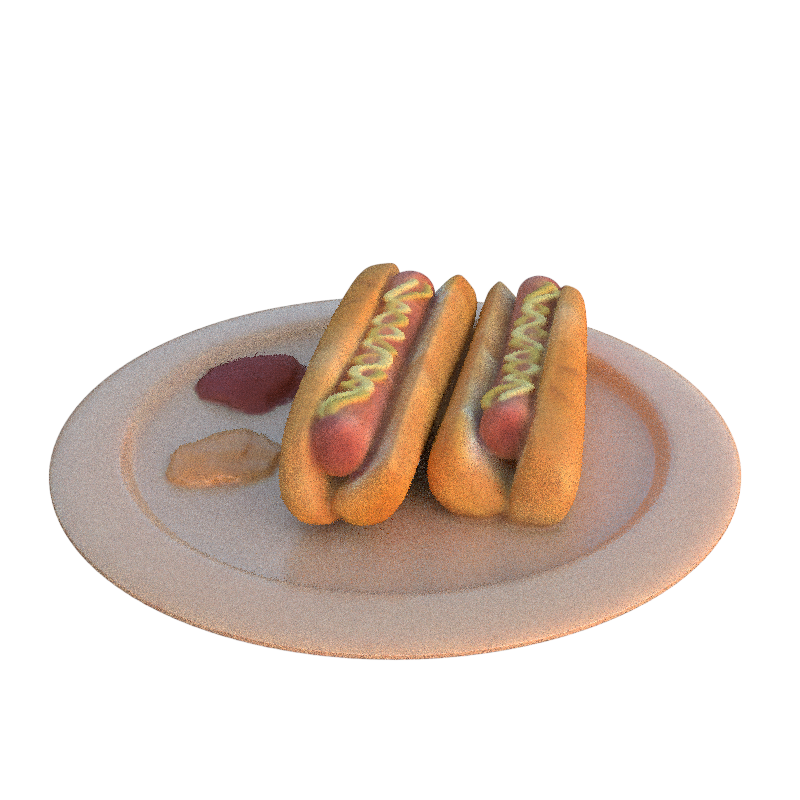} &
         \includegraphics[trim={10cm 8cm 6cm 13cm},clip,width=0.327\linewidth]{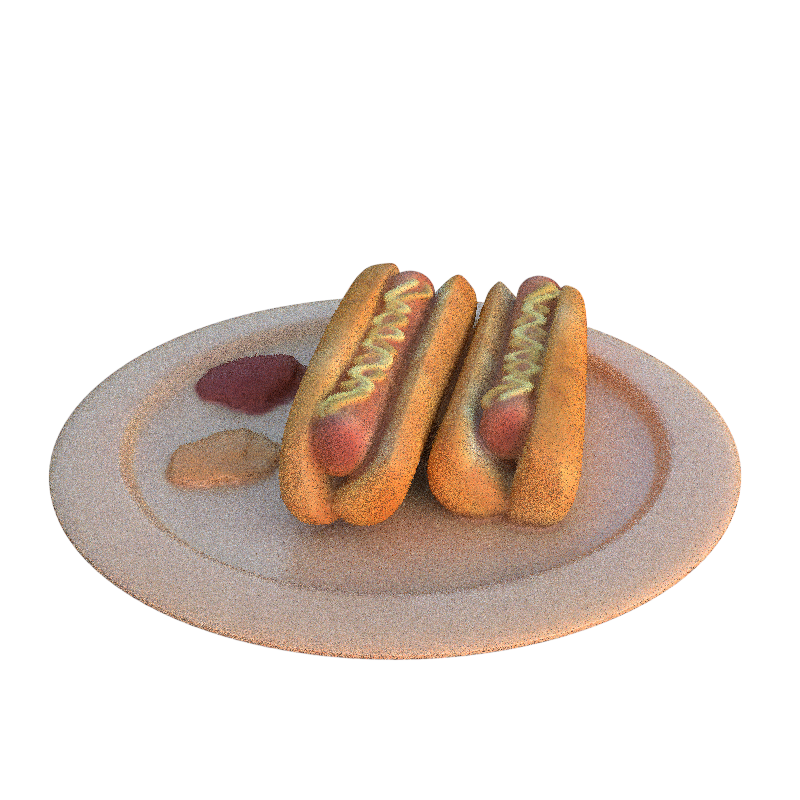} &
         \includegraphics[trim={10cm 8cm 6cm 13cm},clip,width=0.327\linewidth]{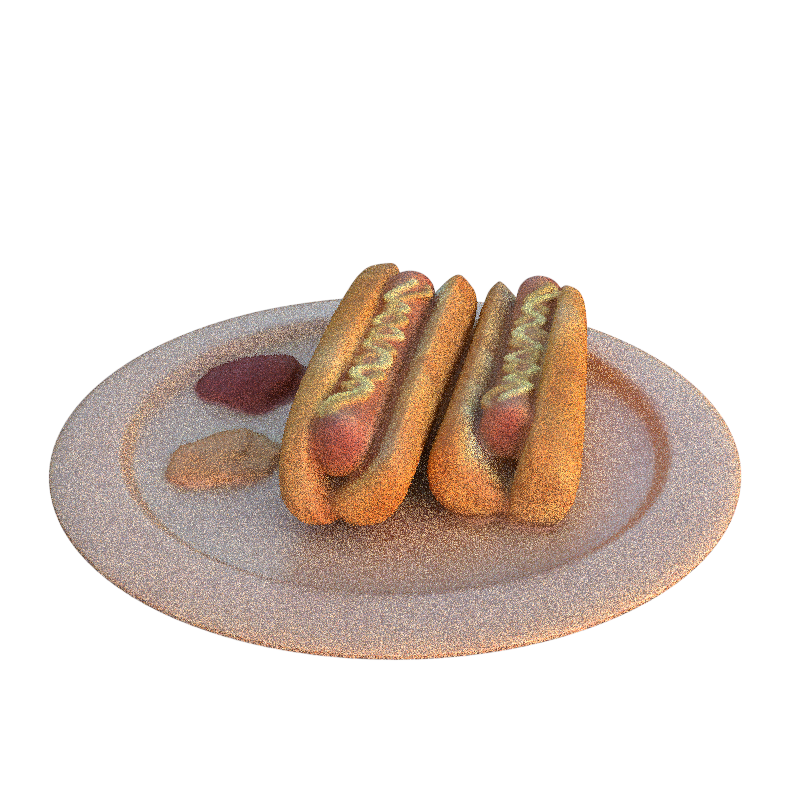}
    \end{tabular}\vspace{-5pt}
    \caption{\label{fig:variance}%
    We demonstrate the impact of both of our variance reduction strategies by showcasing a rendering using 16spp from the \textit{Hotdog} scene in the TensoIR-synthetic dataset. (Left) We use both the fast cache and vMF importance sampler, (Middle) we use only vMF importance sampler, (Right) we use neither strategy. 
    }\vspace{-5pt}
\end{figure}

\section{Experiments} \label{sec:results}
Below, we provide implementation details of our approach, and describe how we validate it on both synthetic and real inverse rendering benchmarks.

\subsection{Implementation Details}

\emph{Model Parameters.} Our NGP parameters for volumetric density closely follow those of Zip-NeRF~\cite{barron2023zipnerf}. We do proposal resampling with two additional NGPs with maximum resolutions of 1024 and 512, both of which, as well as the final density NGP, use of 2-layer 64-hidden-unit MLPs before emitting density. The material NGP as well as another NGP used for diffuse appearance in the cache (see supplement for more details) each has a maximum grid resolution of 2048 and a 2-layer 64-hidden-unit MLP. 
For the fast cache discussed in Section~\ref{sec:method-cache}, we use an NGP of maximum resolution 256 with a 4-layer 128-hidden-unit MLP to predict $S=8$ sample-point distances, and a maximum NGP of resolution 256 to predict features. For the incoming radiance importance sampler in Section~\ref{sec:method-low-variance}, we use 128 vMF lobes, NGP with a maximum resolution of 2048, and a 2-layer 64-hidden-unit MLP. The hash maps of all NGPs have size $2^{19}$.





\emph{Regularization.} We implement a smoothness regularizer $\mathcal{L}_{\textit{BRDF}}$ for BRDF parameters. In particular, we use a variant of TensoIR's smoothness regularizer, with an $L1$ rather than an $L2$ loss (see the supplement for more details). We also enforce a loss $\mathcal{L}_{\textit{consistency}}$ that encourages the specular color and diffuse color from the NeRF cache to match the specular and diffuse lobes from physically-based rendering. This \textit{indirectly} provides gradients through the physically-based model for volume rendering. We further apply the anti-aliased inter-level loss $\mathcal{L}_{\textit{interlevel}}$ from Zip-NeRF to the proposal grids, and $L1$ regularization to the parameters of all density grids with a loss $\mathcal{L}_{\textit{density}}$.

\emph{Additional Details.} We perform two-stage optimization by (1) first optimizing the NeRF cache and an initial estimate of geometry to match a scene's input color images, and (2) optimizing geometry, materials, fast cache, and importance sampler through Equation~\ref{eqn:diff-ren}. For the second stage, we use an $L2$ loss with the gradient trick~\cite{GkiouZBZL2013}, in order to ensure that our gradients for inverse rendering are unbiased. We leverage the same secondary rays that are used in rendering the physically-based model to supervise the fast cache and importance sampler. We \textit{do not} apply full stop-gradients to any part of the pipeline, so that in the second stage geometry optimization can benefit from reflection and shading cues using the more accurate physically-based model of appearance.  Altogether, we have the following:
\begin{equation}
    \mathcal{L}_{\textit{reg}} = \mathcal{L}_{\textit{normals}} + \mathcal{L}_{\textit{BRDF}} + \mathcal{L}_{\textit{consistency}} + \mathcal{L}_{\textit{interlevel}} + \mathcal{L}_{\textit{density}}
\end{equation}
\noindent All training and evaluation is carried out using a single NVIDIA A100 GPU.

\begin{figure*}[t!]
\footnotesize

\centering
\newcommand{\resultwidth}{1.05in}
\newcommand{\ficuswidth}{0.5in} 
\begin{tabular}{@{}c@{}c@{}c@{}c|c@{}c@{}c@{}}
& GT & Ours & TensoIR & GT & Ours & TensoIR \\
\rotatebox{90}{\,\,\, Albedo} & \adjincludegraphics[width=\resultwidth, trim={0.4in 2in 0.4in 4in}, clip]{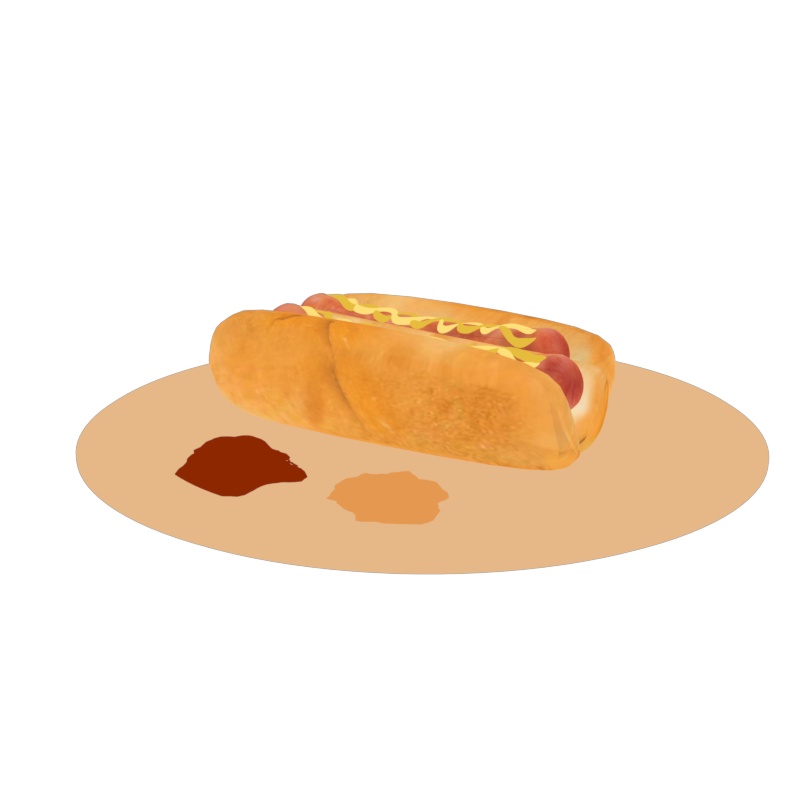} & \adjincludegraphics[width=\resultwidth, trim={0.4in 2in 0.4in 4in}, clip]{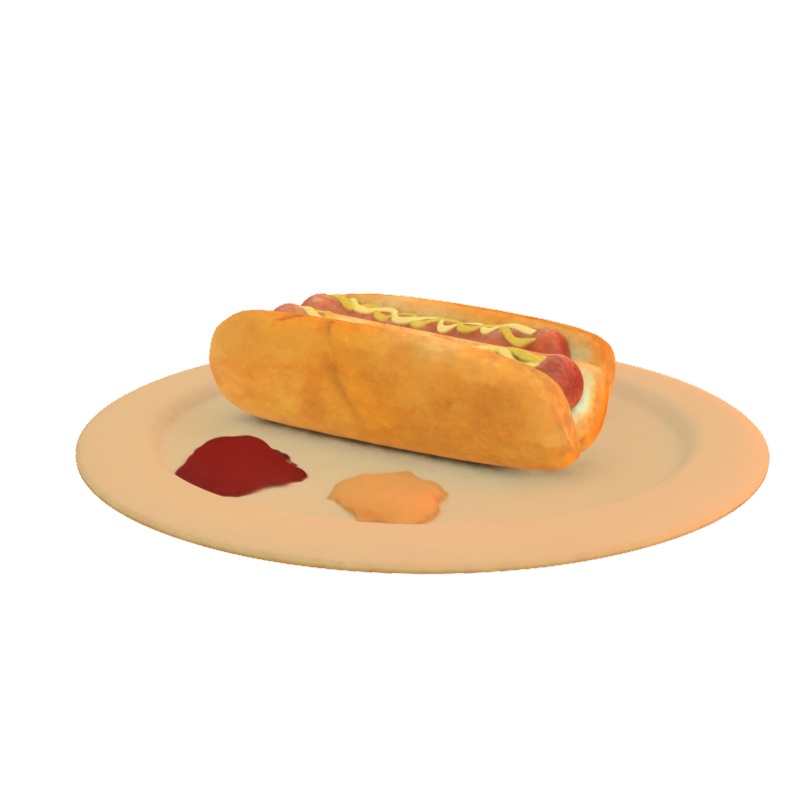} & \adjincludegraphics[width=\resultwidth, trim={0.4in 2in 0.4in 4in}, clip]{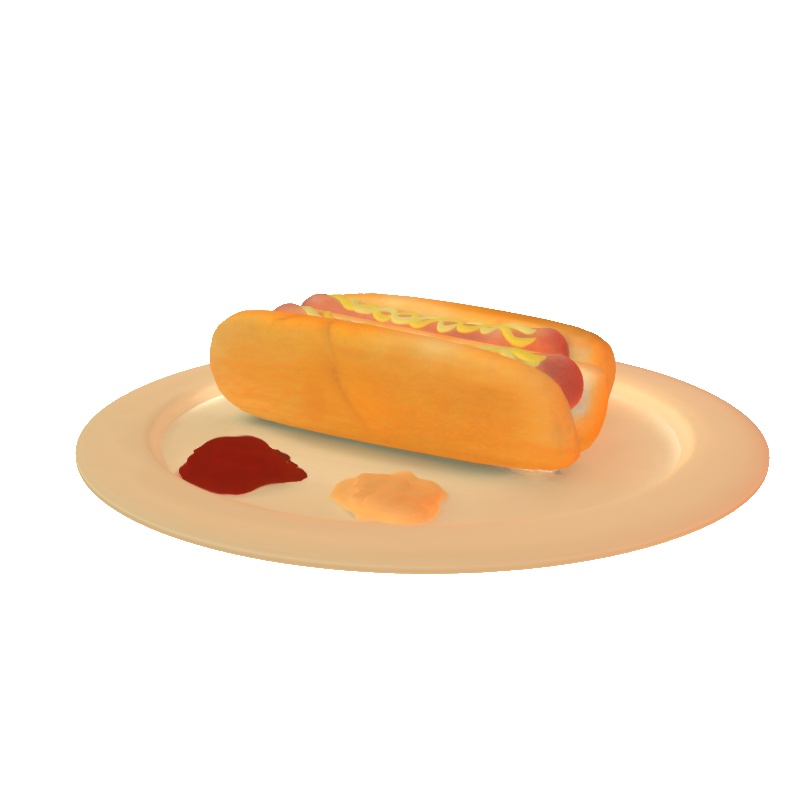} & \adjincludegraphics[width=\ficuswidth, trim={4.5in 1.75in 4.5in 6.5in}, clip]{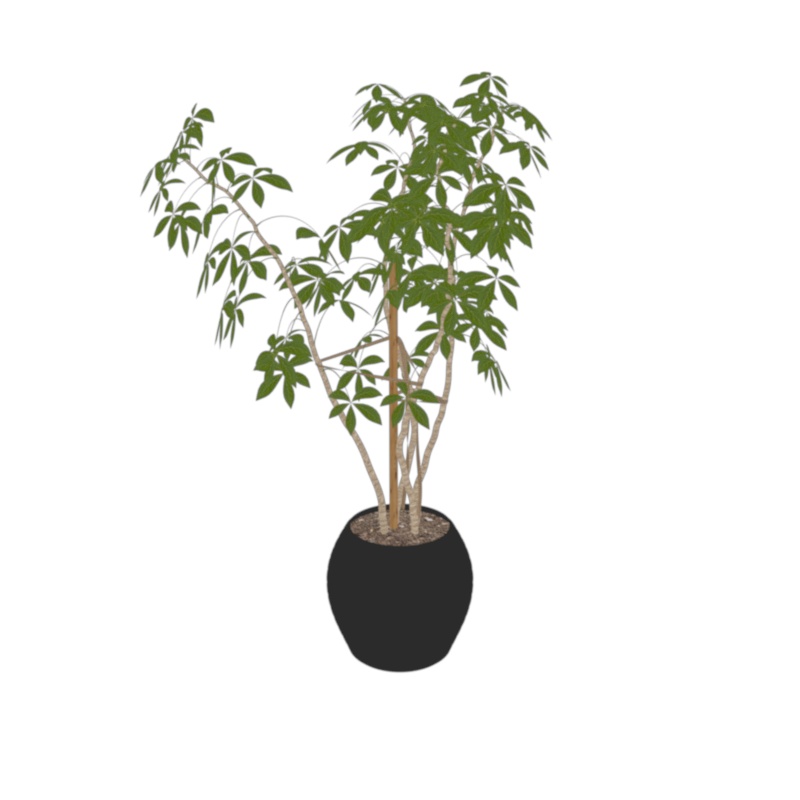} & 
\adjincludegraphics[width=\ficuswidth, trim={4.5in 1.75in 4.5in 6.5in}, clip]{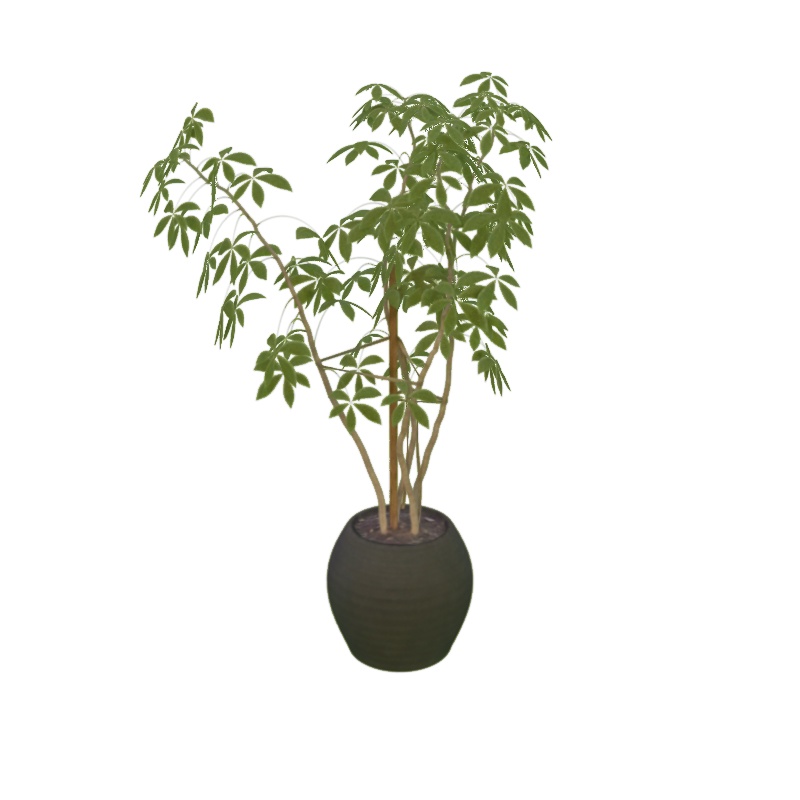} & 
\adjincludegraphics[width=\ficuswidth, trim={4.5in 1.75in 4.5in 6.5in}, clip]{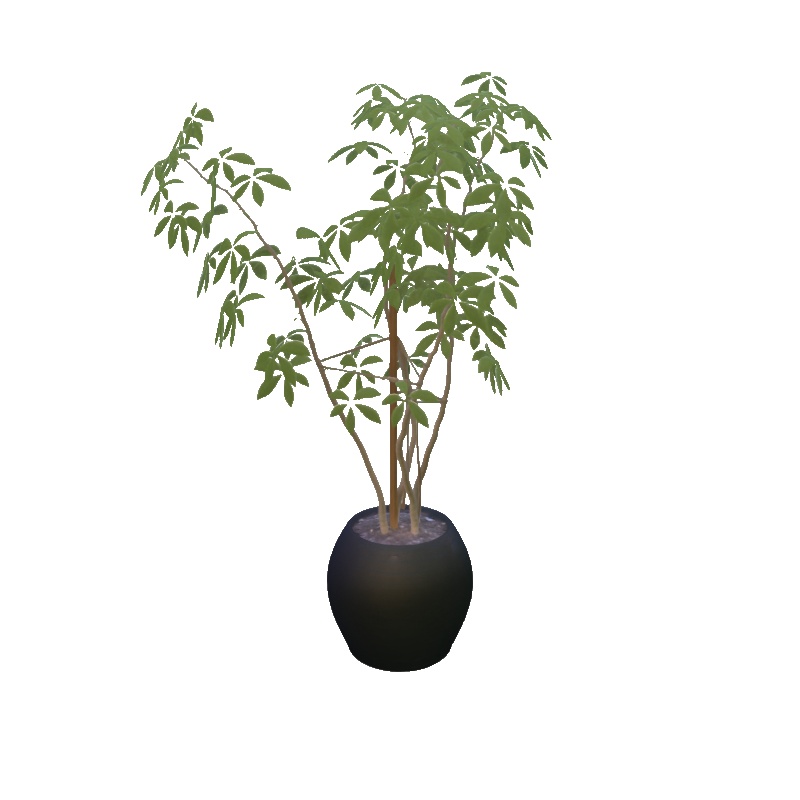} \\
\rotatebox{90}{\quad\, Relit} & \adjincludegraphics[width=\resultwidth, trim={0.4in 2in 0.4in 4in}, clip]{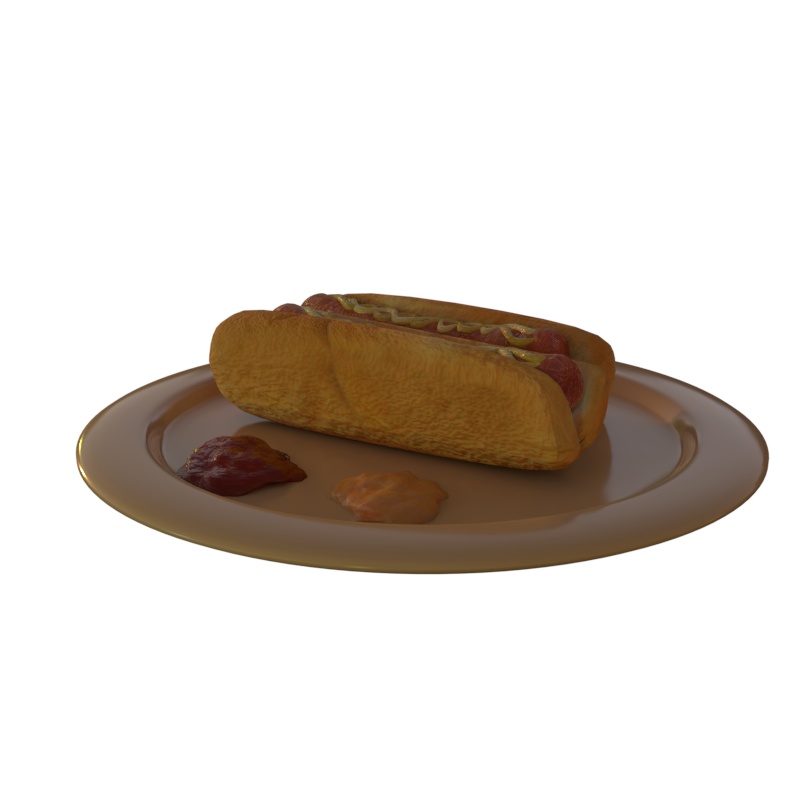} &
\adjincludegraphics[width=\resultwidth, trim={0.4in 2in 0.4in 4in}, clip]{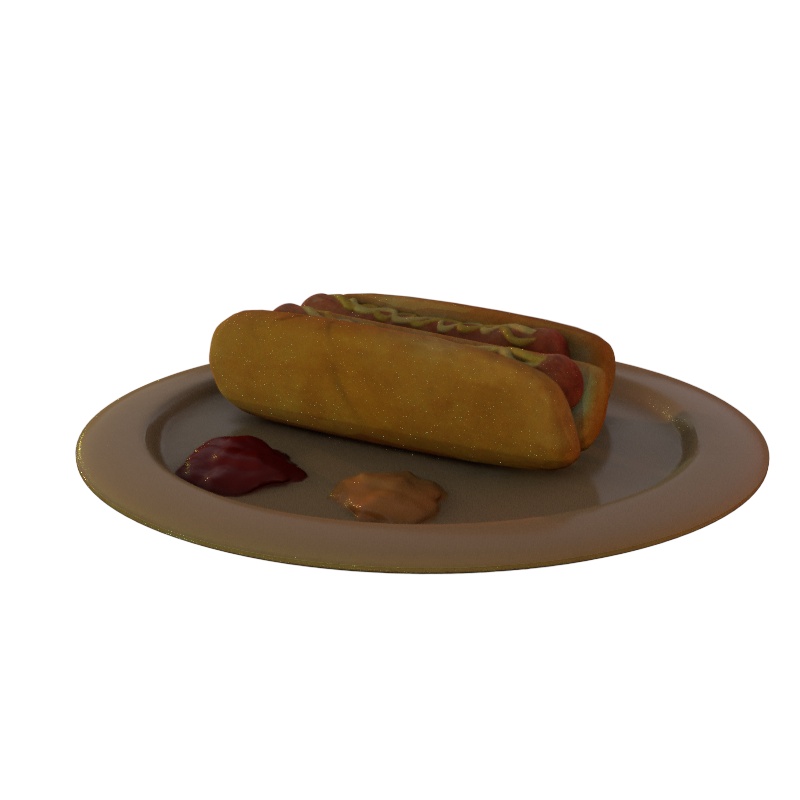} &
\adjincludegraphics[width=\resultwidth, trim={0.4in 2in 0.4in 4in}, clip]{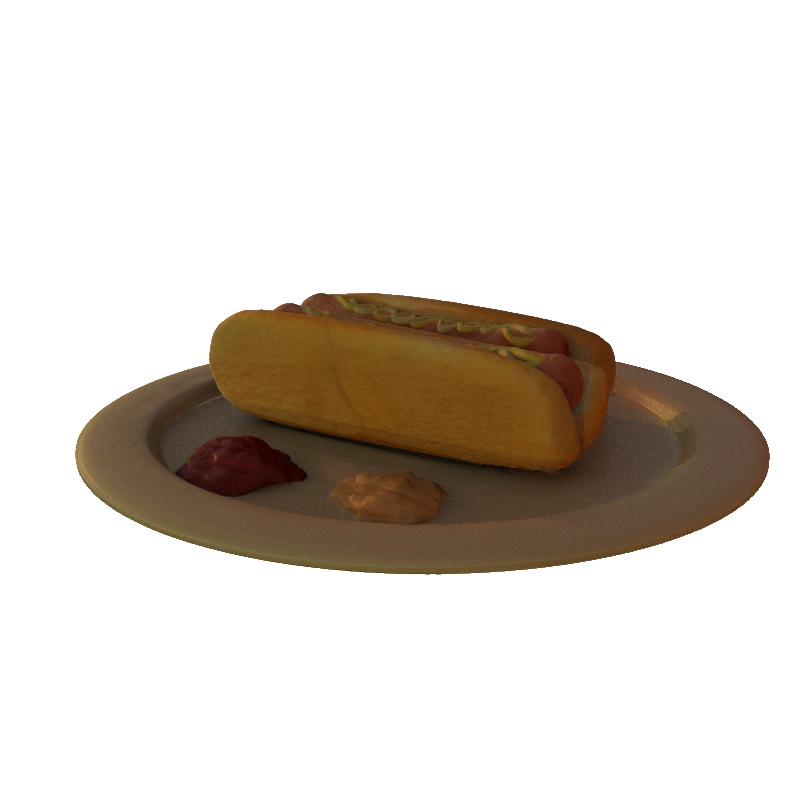} &
\adjincludegraphics[width=\ficuswidth, trim={4.5in 1.75in 4.5in 6.5in}, clip]{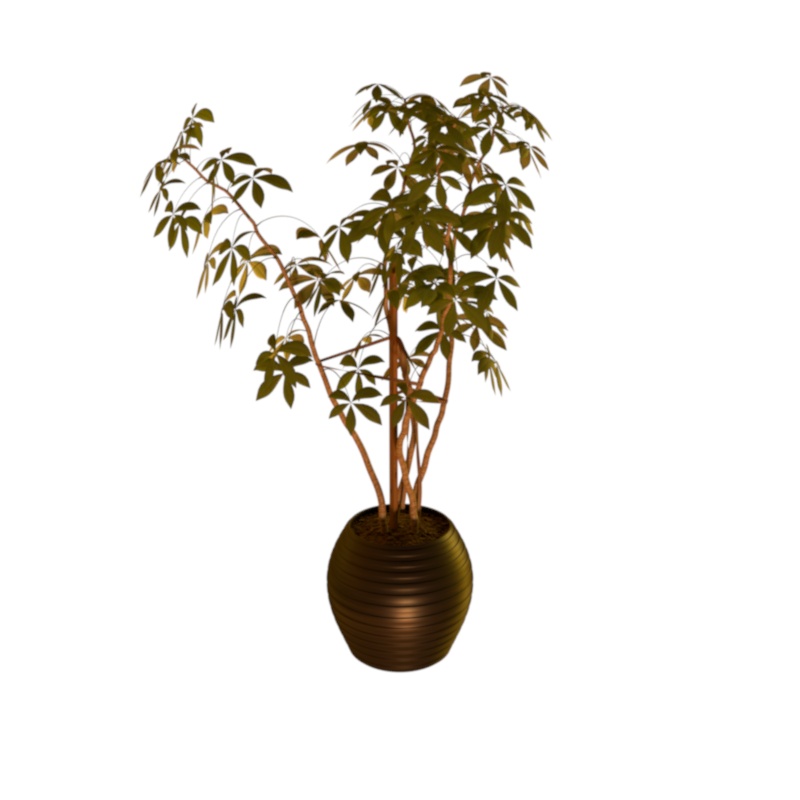} & \adjincludegraphics[width=\ficuswidth, trim={4.5in 1.75in 4.5in 6.5in}, clip]{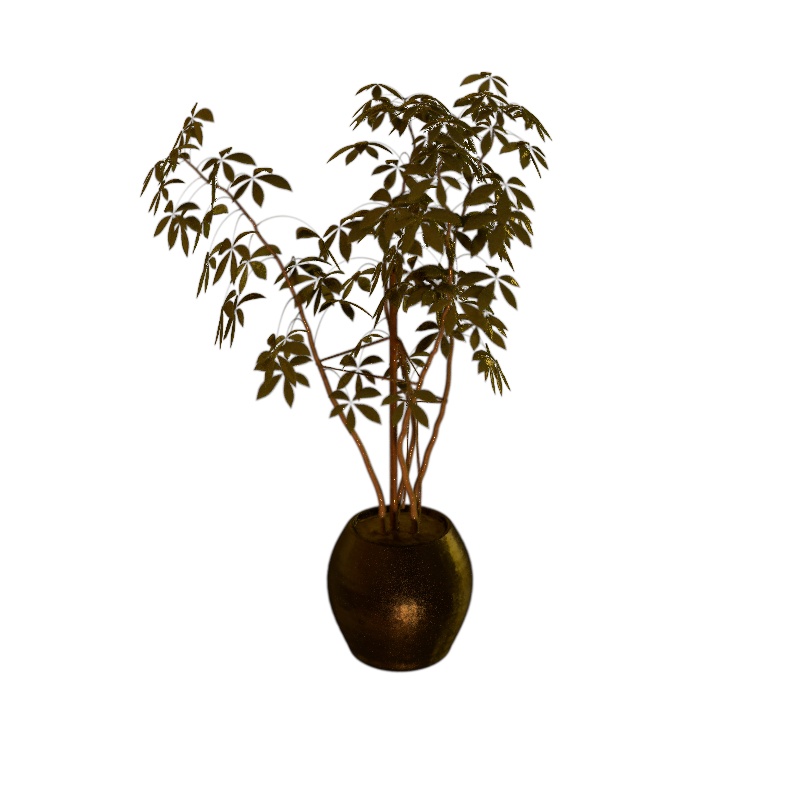} & \adjincludegraphics[width=\ficuswidth, trim={4.5in 1.75in 4.5in 6.5in}, clip]{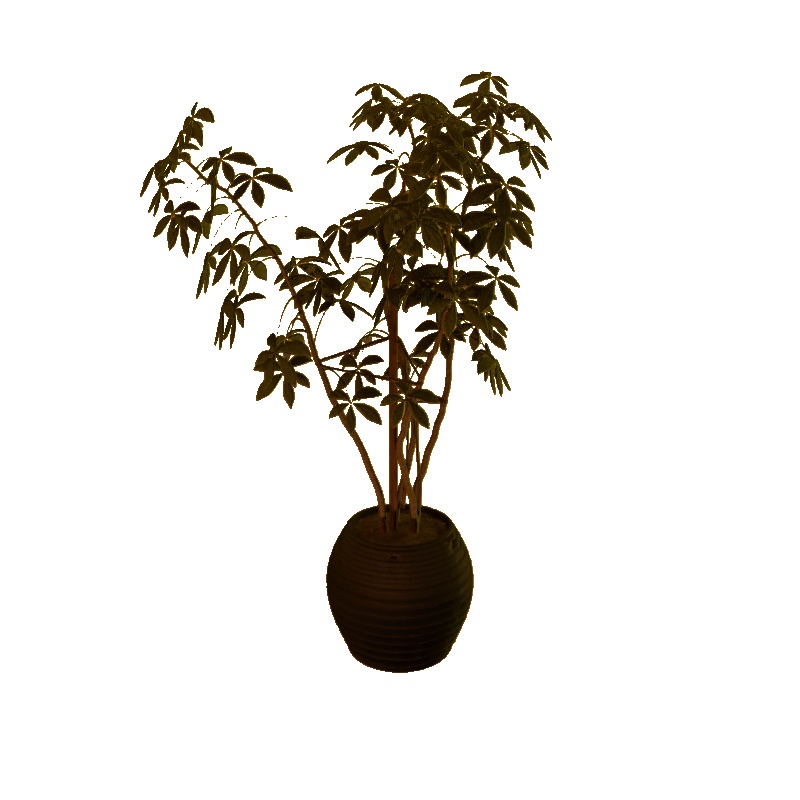}

\end{tabular}
\vspace{-0.1in}
\caption{%
A visualization of ground-truth albedo and relighting for the \textit{hotdog} (left) and \textit{ficus} (right, cropped) scenes in the Blender dataset~\cite{MildeSTBRN2020}, alongside our and TensoIR's results. Our method better decouples strong indirect illumination from the albedo (see the lip of the plate) while also reproducing more accurate highlights.
}%
\label{fig:qual}
\end{figure*}

\subsection{Results on Synthetic Data}

\begin{SCfigure}[50][t]
\caption{
Real results from Open Illumination~\cite{liu2024openillumination}. Our normals are more accurate than those from TensoIR~\cite{jin2023tensoir}, and our albedo correctly removes specularities. Because of scale and color ambiguities, albedo maps are all visualized after using Photoshop's auto-tone, auto-contrast, and auto-color features.
}
\vspace{0pt}\includegraphics[width=0.55\linewidth]{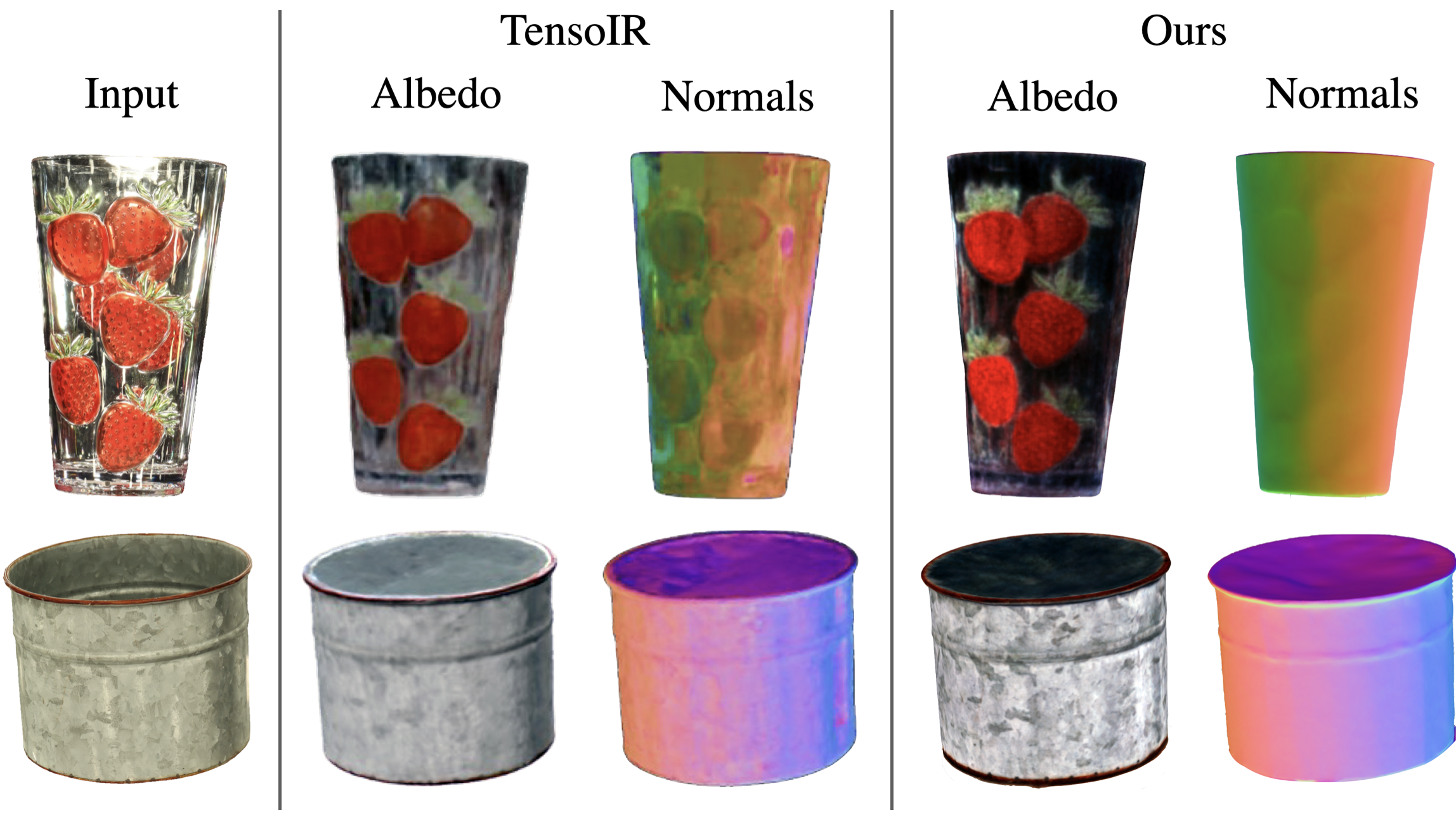}
\label{fig:openillumresults}
\end{SCfigure}

We first evaluate our approach on the TensoIR-synthetic dataset, which contains four scenes: \textit{armadillo}, \textit{ficus}, \textit{hotdog}, and \textit{lego}. Each scene consists of a set of 100 training views and 200 test views containing ground truth normals and albedo, as well as  ground truth color under several different lighting conditions. We evaluate (a) the quality of our recovered normals, (b) the quality of our recovered albedo maps, (c) novel view synthesis quality for the \textit{physically-based model}, and (d) relighting quality. Our method outperforms baselines in terms of albedo quality and relighting, while performing comparably to the best method for novel view synthesis. See Table~\ref{tab:tensoir}, Figure~\ref{fig:qual}, and the supplement for visualizations.

\begin{table*}[t]
    \centering
    \Large
    \resizebox{\linewidth}{!}{
        \begin{tabular}{@{}c c c c ccc c ccc c ccc c c@{}}
            \toprule
            \multirow{2}{*}{Method} & & 
            \multicolumn{1}{c}{Normal} & & \multicolumn{3}{c}{Albedo} & & \multicolumn{3}{c}{Novel View Synthesis} & & \multicolumn{3}{c}{Relighting}\\ \cline{3-3} \cline{5-7} \cline{9-11} \cline{13-15}
            & & MAE $\downarrow$ & & PSNR $\uparrow$ & SSIM $\uparrow$ & LPIPS $\downarrow$ & & PSNR $\uparrow$ & SSIM $\uparrow$ & LPIPS $\downarrow$ & & PSNR $\uparrow$ & SSIM $\uparrow$ & LPIPS $\downarrow$ \\ \hline
            
            NeRFactor & & 6.314 & & 25.125 & 0.940 & 0.109 & & 24.679 & 0.922 & 0.120 & & 23.383 & 0.908 & 0.131\\
            
            InvRender & & 5.074 & & 27.341 & 0.933 & 0.100 & & 27.367 & 0.934 & 0.089 & & 23.973 & 0.901 & 0.101 \\
            
            TensoIR & &  4.100 & & 29.275 & 
            0.950 & 
            0.085 & & 
            \bf 35.088 & \bf 0.976 & \bf 0.040 & & 
            28.580 & \bf 0.944 & 0.081  \\

            Ours & &  \bf 3.355 & & \bf 30.274 & 
            \bf 0.955 & 
            \bf 0.085 & & 
            34.908 & 0.965 & 0.064 & & 
            \bf 29.724 & \bf 0.944 & \bf 0.080  \\
            
            \bottomrule
        \end{tabular}
    }\vspace{1mm}
    \caption{
    \textbf{TensoIR-Synthetic Dataset~\cite{jin2023tensoir} Results.} Above, we present results of our method and 3 baselines on the TensoIR-synthetic dataset. We outperform all methods in terms of normal MAE, albedo quality, and relighting quality, while performing comparably to TensoIR in terms of novel view synthesis quality.
    }
    \label{tab:tensoir}
\end{table*}

\begin{SCtable}[][t]
\centering
\resizebox{0.6\linewidth}{!}{
\begin{tabular}{c@{\quad}c@{\quad}c@{\quad} c c@{\quad}c@{\quad}}
    \toprule
    \multirow{2}{*}{Method} &
     \multicolumn{2}{c}{PSNR$\uparrow$  (\textit{diffuse})} & & \multicolumn{2}{c}{PSNR$\uparrow$ (\textit{specular})}\\ 
    \cline{2-3} \cline{5-6}
     & NVS & Relighting  & & NVS & Relighting \\ \hline

    TensoIR  & \bf 32.043 & \bf 30.932 & & 27.115 & 26.955  \\
    Ours     & 31.781 & 29.679 & & \bf 27.180 & \bf 27.810 \\
    
    \bottomrule
\end{tabular}
}
\caption{
A comparison to TensoIR~\cite{jin2023tensoir} on the Open Illumination dataset~\cite{liu2024openillumination} split into \textit{diffuse} and \textit{specular} scenes.
}
\label{tab:open}
\end{SCtable}
\begin{table*}[t]
\centering
\caption{
\textbf{Ablations} We progressively ablate (a) our control variate scheme~ (Equation~\ref{eqn:fast-cache}), (b) our vMF importance sampler~(Section \ref{sec:method-low-variance}), and (c) our estimator~(Section \ref{eqn:fast-volume-render}), replacing it with median depth sampling.
}
\vspace{-0.1in}
\begin{tabular}{l c@{\,\,\,\,}c@{\,\,\,\,}c}
    \toprule
    Method & NVS PSNR$\uparrow$ & Albedo PSNR$\uparrow$ & MAE$\downarrow$ \\ \hline

    \phantom{(a)} Ours & \bf 34.915 & \bf 30.345 & 3.355  \\
    (a) Ours, no variate & 34.629 & 30.329 & \bf 3.352  \\
    (b) Ours, no variate, no sample & 34.653 & 30.237 & \bf 3.352  \\
    (c) Ours, no variate, no sample, no estimator & 34.249 & 29.838 & 3.491\\
    \bottomrule
\end{tabular}
\vspace{-0.1in}
\label{tab:ablation}
\end{table*}

\subsection{Results on Real Data}

We further evaluate on 10 real scenes from the Open Illumination dataset. Each scene has 38 training views and 10 test views under 13 different lighting conditions. We split this dataset into two parts: (a) one that contains mostly diffuse objects, and (b) one that contains mostly glossy/specular objects. We evaluate novel view synthesis quality for our physically-based model, as well as relighting quality. We tend to outperform TensoIR on scenes with specular materials (visualized in Figure~\ref{fig:openillumresults}), while TensoIR performs better for diffuse materials.

\subsection{Ablations}

We ablate various design decisions and parts of our pipeline on the TensoIR synthetic dataset. In particular, we show that all of the following improve result quality: (a) volume rendering vs. median depth sampling, (b)  vMF-importance sampling, and (c) our fast cache as a control variate for incoming illumination.

\section{Conclusion}

In this paper, we have outlined a system for volumetric inverse rendering of geometry, materials, and lighting, where we reduce bias in two ways: (a) by approximating the volume rendering integral (Equation~\ref{eqn:volume-rendering}), and (b) by using Monte Carlo sampling to evaluate the rendering integral (Equation~\ref{eqn:rendering-estimator}). In particular, we accelerate volume rendering while providing correct \emph{expected} gradients for optimization, and we accelerate the evaluation of the rendering equation by combining the high-quality NeRF-based radiance cache with an occlusion-aware importance sampler and a fast-cache-based control variate. This is in contrast to existing approaches which either use a single fixed intersected surface per ray, or use low-capacity and blurry representations of incoming radiance; both of which introduce bias. We show that our system achieves better quality material reconstruction and relighting than baselines on standard benchmarks containing both real and synthetic data, and justify our design decisions with ablations. 

\paragraph{Limitations \& Future Work:} Despite our method's reduced bias, the quadrature approximation to the volume rendering integral (Equation~\ref{eqn:quadrature}) is still a source of bias. Our proposed approach can potentially be combined with unbiased volume rendering methods such as differential ratio tracking [Nimier-David et al. 2022].  

Our fast cache sometimes struggles to capture particularly thin structural details in near-field incoming illumination. While our control variate scheme prevents this from introducing bias into the optimization process, it does increase gradient variance. In principle, any implementation of a fast cache may be used, such as a high-quality baked representation of a neural radiance field~\cite{HedmaSMBD2021, yariv2023bakedsdf, reiser2023merf}. We leave investigation of alternative fast cache architectures up to future work.

Note that we do not claim that our method is more efficient than existing approaches; rather, that we provide efficient tools to accelerate radiance cache queries and reduce rendering variance, allowing our less-biased \& more physically accurate system to run tractably on commodity GPUs. Our method could potentially be combined with other tools (e.g. denoisers) to further reduce secondary ray sample counts, accelerate inference, and increase convergence speed. 

\clearpage

\paragraph{Acknowledgements}
This work was carried out while Benjamin was an intern at Google Research. Authors thank Rick Szeliski, Aleksander Holynski, and Janne Kontkanen for fruitful discussions, as well as Isabella Liu for help with TensoIR comparisons. Benjamin Attal is supported by a Meta Research PhD Fellowship. Matthew O'Toole acknowledges support from NSF CAREER 2238485 and a Google gift.

%
%
\bibliographystyle{splncs04}
\bibliography{egbib}

\clearpage

\appendix
\section{Overview}

In Section~\ref{sec:add-implementation}, we provide additional implementation details regarding the cost of the fast cache, the cost of the NeRF radiance cache, and the cost and implementation of the physically-based model. We also provide additional details about training, evaluation, and the losses that we use. In Section~\ref{sec:proofs} we prove the unbiasedness of our control variate scheme and our estimator for volume rendering quadrature. In Section~\ref{sec:add-ablate} we clarify some details regarding our ablations, and provide additional ablations for different values of $\mathrm{K}$ for our volume rendering quadrature estimator. Finally, in Section~\ref{sec:add-results}, we provide additional result figures and per-scene quantitative metrics. Please see our \href{https://benattal.github.io/flash-cache/}{supplemental website} for more results and videos.

\section{Additional Implementation Details}
\label{sec:add-implementation}

\subsection{Fast Cache}
\label{sec:add-implementation:1}
Rendering a single ray with our fast cache requires:
\begin{enumerate}
    \item Querying the distance  NGP grid once.
    \item Querying the feature NGP grid 8 times.
    \item Querying the deferred shading MLP once.
    \item Querying the environment color once.
\end{enumerate}
\noindent We additionally predict the sum of rendering weights for an incoming ray, which is supervised to match the sum of the rendering weights from the NeRF cache. This prevents us from having to query the NeRF cache before blending the fast cache color with the environment color. The full cost of casting a secondary ray (steps 1-4 above) is 0.128~sec for the fast cache (per million rays).

\subsection{NeRF Radiance Cache}
\label{sec:add-implementation:2}
The cost of rendering a ray from the NeRF radiance cache requires:
\begin{enumerate}
    \item Querying the first proposal NGP 64 times.
    \item Querying the second proposal NGP 64 times.
    \item Querying the density NGP 32 times.
    \item Querying diffuse color $\mathbf{c}_{d}$ and specular color $\mathbf{c}_{s}$ at each of the final 32 sample points, where the specular color is produced using the same architecture as the fast cache.
    \item Querying environment color once.
\end{enumerate}

The full cost of casting a secondary ray (steps 1-5 above) is  1.574~sec for the NeRF cache (per million rays).

\subsection{Physically-Based Model}
\label{sec:add-implementation:3}
For ablations that \textit{do not} use the fast cache, we use 16 samples from the NeRF cache. For models and ablations that use the fast cache, when rendering from the physically-based model, we use 64 samples to estimate incoming $\hat{L}_\mathrm{o}^{\mathit{fast}}$ from the fast cache, and 16 samples for both the fast cache and NeRF cache to estimate $\mathrm{\Delta}\hat{L}_\mathrm{o}$.

\subsection{Training Details}
\label{sec:add-implementation:4}

\paragraph{Cache training:} In the first stage of training, we optimize the Zip-NeRF-based radiance cache with a batch size of $16384$ rays for $100{,}000$ iterations using the Charbonnier loss as in~\cite{barron2023zipnerf}. We use a learning rate schedule that warms up from $0$ to $0.01$ over the first $10{,}000$ iterations, and then decreases linearly to $0.001$ for the remaining $90{,}000$ iterations. Training for $100{,}000$ iterations takes $5.5$ hours on a single A100.

\paragraph{Joint training:} In the second stage of training, we optimize the physically-based model, environment map, fast cache, NeRF-cache, and occlusion-aware importance sampler. We use a batch size of $1024$ rays for $40{,}000$ iterations for the TensoIR-synthetic dataset and $100{,}000$ iterations for the Open Illumination dataset, with $64$ secondary rays for the fast cache, and $16$ rays for the NeRF-cache. We use a learning rate of $1.5625  \times 10^{-4}$ for the fast cache and occlusion aware importance sampler. For everything else, we use a learning rate of $3.125 \times 10^{-5}$. Training for $100{,}000$ iterations takes 6.5 hours on a single A100.

\paragraph{Photometric loss:} For the physically-based model, instead of the Charbonnier loss we use a variant of the RawNeRF~\cite{mildenhall2022nerf} loss. For a given ray $(\xorigin, \dirout)$, predicted color $L_\mathrm{i}(\xorigin, \dirout ; \mathrm\Phi)$ and ground truth $I(\xorigin, \dirout)$, we \textit{would like to} minimize:
\begin{align}
   \mathcal{L}_{\textit{photometric}} = \frac{\lft\lVert I(\xorigin, \dirout) - \mathbb{E}\lft[\hat{L}_\mathrm{i}(\xorigin, \dirout ; \mathrm\Phi)\rgt] \rgt\rVert ^ 2}{\operatorname{stopgrad}\lft(L_\mathrm{i}^{\textit{cache}}(\xorigin, \dirout ; \mathrm\Phi)\rgt)}
\end{align}

\noindent where $\mathbb{E}\lft[\hat{L}_\mathrm{i}(\xorigin, \dirout ; \mathrm\Phi)\rgt]$ is the expected value of the estimator for radiance $\hat{L}_\mathrm{i}$. However, we cannot minimize this loss directly, since we do not have access to an analytic expression for the expected value. Instead, we apply the gradient trick~\cite{GkiouZBZL2013}, which gives correct gradients through this loss in expectation for the physically-based model. More concretely, we minimize:
\begin{align}
   \mathcal{L}_{\textit{photometric}} = \frac{2 \lft( I(\xorigin, \dirout) - \hat{L}_\mathrm{i}(\xorigin, \dirout ; \mathrm\Phi) \rgt) \operatorname{stopgrad}\lft( I(\xorigin, \dirout) - \hat{L}_\mathrm{i}(\xorigin, \dirout ; \mathrm\Phi)\rgt)}{ \operatorname{stopgrad}(\hat{L}_\mathrm{i}^{\textit{cache}}(\xorigin, \dirout ; \mathrm\Phi))}
   \label{eqn:rawnerf}
\end{align}
where the $\operatorname{stopgrad}(\cdot)$ operator treats its argument as a constant in the compute graph, and the first and second differences in the numerator are estimated using independent secondary samples.

\subsection{Evaluation Details}
During evaluation we use $64$ secondary rays, but render the predicted color image 32 times and average the results in order to reduce Monte Carlo noise. We noticed that TensoIR renderings still contain some noise, and acknowledge that different sample counts and different importance sampling schemes could impact relative PSNR, but note that we outperform TensoIR in terms of albedo metrics, which is not affected by noise, and relighting by a fairly large margin.

\subsection{BRDF Model}
As we discuss in the main text, we use the Disney-GGX BRDF~\cite{burley2012physically}, which is comprised of three terms: albedo $\mathbf{a}(\xpos)$, metalness $m(\xpos)$, and roughness $r(\xpos)$. This BRDF can be written as:
\begin{gather}
    f(\dirin, \dirout, \xpos) = f_{\textit{diffuse}}(\xpos) + f_{\textit{specular}}(\dirin, \dirout, \xpos) \\
    f_{\textit{diffuse}}(\xpos) = \frac{(1 - m(\xpos)) \mathbf{a}(\xpos)}{\pi} \\
    f_{\textit{specular}}(\dirin, \dirout, \xpos) = \frac{DFG}{4 (\normal \cdot \dirin) (\normal \cdot \dirout)}
\end{gather}

\noindent We refer readers to Burley~\cite{burley2012physically} and Liu \etal~\cite{liu2023nero} for definitions of $(D, F, G)$ --- the normal distribution function (NDF), Fresnel, and geometry terms. We use the Trowbridge-Reitz distribution function~\cite{pharr2023physically} for the NDF $D$.

\subsection{Importance Sampling}
For both the fast and NeRF caches, we split secondary samples evenly between the diffuse color (due to the diffuse BRDF component $f_{\textit{diffuse}}$) and the specular color (due to the specular BRDF component $f_{\textit{specular}}$). For the diffuse color, we leverage multiple importance sampling~\cite{pharr2023physically} with half of the samples coming from the occlusion-aware importance sampler and half of the samples coming from cosine-weighted hemisphere sampling. For the specular color, we importance sample according to the distribution function $D$~\cite{pharr2023physically}.

\subsection{Normal Loss}
As discussed in the paper, we emit predicted normals from the density NGP. Similar to Ref-NeRF~\cite{VerbiHMZBS2022} and TensoIR~\cite{jin2023tensoir}, we constrain our predicted normals to match the negative gradient of the density field with an L2 loss:
\begin{align}
    \mathcal{L}_{\textit{normals}} = C_{\textit{normals}} \sum_{k} w_k \lft\lVert \normal_k^{\textit{pred}} - \normal_k^{\textit{derived}} \rgt\rVert ^ 2
\end{align}

\noindent where $w_k$ are the render weights for a given ray, and
\begin{align}
    \normal_k^{\textit{derived}} = -\frac{\nabla \sigma(\xpos_k)}{\lVert \nabla \sigma(\xpos_k)\rVert}
\end{align}

\noindent The loss weight $C_{\textit{normals}}$ varies per-dataset. For the Open Illumination dataset, we use $C_{\textit{normals}} = 1.0$. For the TensoIR-synthetic dataset, we linearly interpolate $C_{\textit{normals}}$ from $0.0001$ to $1.0$ from iteration $20{,}000$ to iteration $40{,}000$.

\subsection{Cache Consistency Loss}
To allow the physically-based model to constrain the appearance of the NeRF-cache, we supervise the diffuse and specular colors from the cache $\mathbf{c}_{d}^{\textit{cache}}(\xpos)$ and $\mathbf{c}_{s}^{\textit{cache}}(\xpos, \dirin)$ to match the diffuse and specular colors from the physically-based model $\mathbf{c}_{d}^{\textit{phys}}(\xpos)$ and $\mathbf{c}_{s}^{\textit{phys}}(\xpos, \dirin)$. We further predict an additional output from the cache $\mathbf{c}_{\textit{irradiance}}(\xpos, \dirout)$, which is supervised to match the irradiance from the physically-based model (computed by setting the BRDF to be perfectly Lambertian with $\mathbf{a}(\xpos) = \mathbf{1}$). For all of the above, we use the same Raw-NeRF loss as in Equation~\ref{eqn:rawnerf}.

\subsection{Smoothness Loss}
To enforce BRDF smoothness we use a variant of TensoIR's smoothness loss:
\begin{gather}
    \mathcal{L}_{\textit{BRDF}} = C_{\textit{BRDF}} \sum_{k} w_k \lft| \frac{\mathbf{\beta}(\xpos_k) - \mathbf{\beta}(\xpos_k + \xi)}{\operatorname{max}\lft(\mathbf{\beta}(\xpos_k), \mathbf{\beta}(\xpos_k + \xi)\rgt)} \rgt| \lambda(\xpos, \xpos + \xi) \\
    \lambda(\xpos, \xpos + \xi) = \lft| \mathbf{a}_{\textit{pseudo}}(\xpos_k) - \mathbf{a}_{\textit{pseudo}}(\xpos_k + \xi) \rgt| \quad\quad\quad \xi \sim \mathcal{N}(\mathbf{0}, \epsilon \mathrm{I})
\end{gather}
Where $\mathbf{a}_{\textit{pseudo}}(\xpos_k)$ is the ``pseudo-albedo'' at point $\xpos_k$:
\begin{align}
    \mathbf{a}_{\textit{pseudo}}(\xpos_k) = \frac{\mathbf{c}(\xpos, \dirout)}{\mathbf{c}_{\textit{irradiance}}(\xpos, \dirout)}
\end{align}

\noindent For the TensoIR dataset, we set $\epsilon = 0.01$ with  $C_{\textit{BRDF}} = 0.05$, and for other datasets we set $\epsilon = 0.005$ with $C_{\textit{BRDF}} = 0.001$.

\section{Proofs}
\label{sec:proofs}

\subsection{Control Variates}
Here we show that Equation~\ref{eqn:fast-cache} is an unbiased estimator of the rendering equation (provided that incoming illumination from the NeRF cache is correct):
\begin{align*}
    \mathbb{E}\lft[\hat{L}_\mathrm{o}\rgt] =& \,\,\mathbb{E}\lft[\hat{L}_\mathrm{o}^{\mathit{fast}}\rgt] + \mathbb{E}\lft[\mathrm{\Delta}\hat{L}_\mathrm{o}\rgt] &\text{(Eq. 9)} \\
    =& \int_{\mathrm{\Omega}} f\lft(\xposof{t}, \dirin, \dirout\rgt) \hat{L}_\mathrm{i}^{\textit{NeRF}}\lft(\xpos, \dirin\rgt) \lft(\mathbf{n} \cdot \dirin\rgt)\, d \dirin & \text{(unbiasedness of Eq. 5)}\\
    & + \int_{\mathrm{\Omega}} f\lft(\xposof{t}, \dirin, \dirout\rgt) \lft(\hat{L}_\mathrm{i}^{\textit{NeRF}} - \hat{L}_\mathrm{i}^{\textit{fast}}\rgt)\lft(\xpos, \dirin\rgt) \lft(\mathbf{n} \cdot \dirin\rgt)\, d \dirin \span \\
    =& \int_{\mathrm{\Omega}} f\lft(\xposof{t}, \dirin, \dirout\rgt) \hat{L}_\mathrm{i}^{\textit{NeRF}}\lft(\xpos, \dirin\rgt) \lft(\mathbf{n} \cdot \dirin\rgt)\, d \dirin & \square
\end{align*}
\subsection{Volume Rendering}
Here we show that ~\ref{eqn:fast-volume-render} is an unbiased estimator for volume rendering quadrature (\eg its expected value is equal to Equation~\ref{eqn:quadrature}):

\begin{align*}
    \mathbb{E}\lft[\hat{L}_\mathrm{i}(\xorigin, \dirout)\rgt] &= \mathbb{E}\lft[\frac{1}{\kcount} \sum_{k=1}^{\kcount} L_\mathrm{o}\lft(\xposof{t_{j_k}}, \dirout\rgt) \rgt] & \text{(Eq. 11)} \\
    &= \frac{1}{\kcount} \sum_{k=1}^{\kcount} \mathbb{E}[L_\mathrm{o}\lft(\xposof{t_{j_k}}, \dirout\rgt)] & ({j_k} \sim \operatorname{Cat}(w_1, \ldots, w_\ncount) ) 
 \\
    &= \frac{1}{\kcount} \sum_{k=1}^{\kcount} \sum_{j=1}^{\ncount} w_j L_\mathrm{o}\lft(\xposof{t_j}, \dirout\rgt) & \\
    &= \sum_{j=1}^{\ncount} w_j L_\mathrm{o}\lft(\xposof{t_j}, \dirout\rgt) & \square
\end{align*}
where $(w_1, \ldots, w_\ncount)$ are render weights for the ray $(\xorigin, \dirout)$.

\section{Additional Ablations}
\label{sec:add-ablate}

All ablations are evaluated in the single-light setting of the TensoIR-synthetic dataset.




\subsection{Secondary Sample Ablations}
We report aggregate novel view synthesis PSNR, albedo PSNR, and normal MAE for three additional ablations in Table~\ref{tab:samples} as we vary the value of $\kcount$ in Equation~\ref{eqn:fast-volume-render}. In practice, we find that $\kcount = 1$ provides the best results.

\begin{table*}[h]
\centering
\caption{
\textbf{Sample Number Ablations}
}
\vspace{-0.1in}
\begin{tabular}{l c@{\,\,\,\,}c@{\,\,\,\,}c}
    \toprule
    Method & NVS PSNR$\uparrow$ & Albedo PSNR$\uparrow$ & MAE$\downarrow$ \\ \midrule

    (a) $\kcount = 1$ (Ours) & \bf 34.915 & \bf 30.345 & 3.355  \\
    (b) $\kcount = 2$ & 34.913 & 30.208 & \bf 3.354  \\
    (c) $\kcount = 4$ & 34.850 & 30.275 & 3.356  \\
    (d) $\kcount = 8$ & 34.778 & 30.059 & 3.355  \\
    \bottomrule
\end{tabular}
\label{tab:samples}
\end{table*}

We expect that K = 1 will do worse on complex datasets with “more volumetric” geometry or partial transparencies, although we note that for any K, Equation~\ref{eqn:fast-volume-render} is still an unbiased estimator for volume rendering quadrature.

\section{Additional Results}
\label{sec:add-results}

\subsection{Per-Scene Results for TensoIR}
We provide per-scene results for both the TensoIR-Synthetic dataset in Table~\ref{tab:tensoir}. We additionally provide results for the multi-light TensoIR setting.

\begin{table*}[h]
    \caption{
    \textbf{Per-Scene TensoIR-Synthetic Dataset~\cite{jin2023tensoir} Results. 
    }
    }
    \vspace{-0.1in}
    \centering
    \resizebox{\linewidth}{!}{
        \begin{tabular}{c @{\,\,\,}| c | c | cc@{}c | cc@{}c | cc@{}c }
            \toprule
            & \multirow{2}{*}{Method} & 
            Normal & \multicolumn{3}{c|}{Albedo} & \multicolumn{3}{c|}{Novel View Synthesis} & \multicolumn{3}{c}{Relighting}\\
            & & MAE$\downarrow$ & PSNR$\uparrow$ & SSIM$\uparrow$ & LPIPS$\downarrow$ & PSNR$\uparrow$ & SSIM$\uparrow$ & LPIPS$\downarrow$ & PSNR$\uparrow$ & SSIM$\uparrow$ & LPIPS$\downarrow$ \\ \midrule
            
            \parbox[t]{2mm}{\multirow{6}{*}{\rotatebox[origin=c]{90}{\textit{Armadillo}}}}
 & NeRFactor & 3.467 &
            28.001 & 0.946 & 0.096 &
            26.479 & 0.947 & 0.095 &
            26.887 & 0.944 &  0.102 \\
            
            & InvRender & 1.723 &
            35.573 &  0.959 & 0.076 &
            31.116 & 0.968 & 0.057 &
            27.814 & 0.949 & 0.069 \\
            
            & TensoIR &  1.950 &
            34.360 & 0.989 & 0.059 & 
            39.050 & 0.986 & 0.039  & 
            34.504 & 0.975 & 0.045  \\

            & Ours &  1.564 &
            34.832 & 0.960 & 0.082 & 
            39.313 & 0.977 & 0.043 & 
            34.809 & 0.959 & 0.058  \\

            & TensoIR multi-light &  1.550 &
            34.270 & 0.989 & 0.057 & 
            38.230 & 0.984 & 0.043  & 
            34.941 & 0.977 & 0.043  \\

            & Ours multi-light &  1.427 &
            35.921 & 0.961 & 0.079 & 
            39.097 & 0.978 & 0.042 & 
            35.937 & 0.965 & 0.052  \\ \midrule

            \parbox[t]{2mm}{\multirow{6}{*}{\rotatebox[origin=c]{90}{\textit{Ficus}}}} & NeRFactor & 6.442 &
            22.402 & 0.928 & 0.085 &
            21.664 & 0.919 & 0.095 &
            20.684 & 0.907 & 0.107 \\
            
            & InvRender & 4.884 &
            25.335 & 0.942 & 0.072 &
            22.131 & 0.934 & 0.057 &
            20.330 & 0.895 & 0.073 \\
            
            & TensoIR &  4.420 &
             27.130 &  0.964 & 0.044 & 
            29.780 &  0.973 &  0.041  & 
             24.296 & 0.947 &  0.068  \\

            & Ours &  2.709 &
            28.337 & 0.972 & 0.048 & 
            30.380 & 0.976 & 0.036 & 
            26.286 & 0.960 & 0.052  \\

            & TensoIR multi-light &  4.060 &
            26.220 & 0.952 &  0.054 & 
             28.640 & 0.967 & 0.050  & 
             24.622 & 0.949 &  0.068  \\

            & Ours multi-light &  2.689 &
            28.137 & 0.971 & 0.046 & 
            30.175 & 0.976 & 0.037 & 
            26.730 & 0.964 & 0.049  \\ \midrule

            \parbox[t]{2mm}{\multirow{6}{*}{\rotatebox[origin=c]{90}{\textit{Hotdog}}}} & NeRFactor &  5.579 &
            24.654 &  0.950 &  0.142 &
            24.498 & 0.940 &  0.141 &
            22.713 & 0.914 &  0.159 \\
            
            & InvRender & 3.708 &
            27.028 &   0.950 & 0.094 &
            31.832 & 0.952 & 0.089 &
            27.630 & 0.928 & 0.089 \\
            
            & TensoIR &   4.050 &
            30.370 &  0.947 & 0.093 & 
            36.820 &  0.976 &  0.045  & 
            27.927 & 0.933 & 0.115  \\

            & Ours &  2.882 &
           30.832 & 0.966 & 0.073 & 
            36.966 & 0.961 & 0.095 & 
            29.241 & 0.941 & 0.104  \\

            & TensoIR multi-light &   3.220 &
            31.240 & 0.958 &  0.080 & 
             35.670 & 0.973 &  0.048  & 
              28.952 & 0.939 &   0.110  \\

            & Ours multi-light &  2.741 &
            31.180 & 0.968 & 0.077 & 
            36.036 & 0.958 & 0.097 & 
            29.050 & 0.942 & 0.103  \\ \midrule 
            
            \parbox[t]{2mm}{\multirow{6}{*}{\rotatebox[origin=c]{90}{\textit{Lego}}}} & NeRFactor &  9.767 &
            25.444 & 0.937 & 0.112 &
            26.076 &  0.881 &  0.151 &
            23.246 & 0.865 &   0.156 \\
            
            & InvRender &  9.980 &
             21.435 &   0.882 & 0.160 &
            24.391 & 0.883 & 0.151 &
             20.117 & 0.832 & 0.171 \\
            
            & TensoIR &  5.980 &
            25.240 &  0.900 &  0.145 & 
             34.700 & 0.968 & 0.037  & 
             27.596 &  0.922 &  0.095  \\

            & Ours &  6.265 &
            27.097 & 0.922 & 0.137 & 
            32.973 & 0.945 & 0.081 & 
            28.560 & 0.917 & 0.105  \\

            & TensoIR multi-light &  5.370 &
            25.560 & 0.905 &  0.146 & 
             34.350 &  0.967 & 0.038  & 
              27.517 & 0.922 &  0.091 \\

            & Ours multi-light &  5.713 &
            27.735 & 0.917 & 0.143 & 
            31.942 & 0.942 & 0.083 & 
            28.286 & 0.918 & 0.102  \\
            
            \bottomrule
        \end{tabular}
    }
    \label{tab:tensoir}
\end{table*}

\subsection{Per-Scene Results for Open Illumination}
We provide per-scene results for both the Open Illumination dataset in Table~\ref{tab:open}. We label each scene as \textit{diffuse} or \textit{specular}. The dataset provides both single-light data (with the object illuminated under a single lighting condition) and multi-light data (with the object illuminated under many different lighting conditions). We perform evaluation in the single-light setting for novel view synthesis, and in the multi-light setting for relighting, as no relighting metrics are provided in the original Open Illumination paper for the single-light setting. 

\begin{table*}[h]
\centering
\caption{
\textbf{Per-Scene Open Illumination dataset~\cite{liu2024openillumination} Results.}
}
\begin{tabular}{c@{\,\,\,}|c@{\quad}c@{\quad}c@{\quad}c@{\quad}c}
    \toprule
    & \multirow{2}{*}{Scene} & \multirow{2}{*}{Method} &
     \multicolumn{2}{c}{PSNR $\uparrow$} &\\ 
    && & NVS & Relit.  \\ \midrule 
    \parbox[t]{2mm}{\multirow{6}{*}{\rotatebox[origin=c]{90}{Diffuse\quad\quad\quad\quad\quad\quad\quad}}}  
    & \multirow{2}{*}{\textit{Egg}}  & TensoIR  & \bf 34.88 &  \bf 31.99 &  \\
        && Ours     & 34.48 & 30.76 \\ \cmidrule{2-5}

    & \multirow{2}{*}{\textit{Stone}}  & TensoIR  & 29.96 & \bf 31.07 \\
        && Ours     & \bf 30.52 & 30.55 \\  \cmidrule{2-5}
        
    & \multirow{2}{*}{\textit{Pumpkin}}  & TensoIR  &  \bf 28.20 & \bf 27.16 &  \\
        && Ours     & 27.64 & 26.58 & \\ \cmidrule{2-5}

    & \multirow{2}{*}{\textit{Hat}}  & TensoIR  &  \bf 31.96 & \bf 32.38 &  \\
        && Ours     & 31.02 & 30.41 \\ \cmidrule{2-5}
    & \multirow{2}{*}{\textit{Sponge}}  & TensoIR  & \bf 32.49 &  \bf 30.86 &  \\
        && Ours     & 32.14 & 28.87 \\ \cmidrule{2-5}

    & \multirow{2}{*}{\textit{Banana}}  & TensoIR  & 34.77 & \bf 32.13  \\
        && Ours     & \bf 34.89 & 30.90 \\ \midrule 

    \parbox[t]{2mm}{\multirow{4}{*}{\rotatebox[origin=c]{90}{Specular\quad\quad\quad\,\,\,}}}  
 & \multirow{2}{*}{\textit{Bird}}  & TensoIR  &  \bf 30.21 & 30.16  \\
        && Ours     & 29.92 & \bf 30.19 \\ \cmidrule{2-5}

    & \multirow{2}{*}{\textit{Box}}  & TensoIR  & \bf 26.80 &  27.57  \\
        && Ours     & 26.40 & \bf 28.93 \\ \cmidrule{2-5}

    & \multirow{2}{*}{\textit{Cup}}  & TensoIR  & \bf 22.13 & 22.96  \\
        & & Ours     & 21.84 & \bf 23.35 \\ \cmidrule{2-5}

    & \multirow{2}{*}{\textit{Bucket}}  & TensoIR  & 29.32 & 27.13  \\
        && Ours     & \bf 30.55 & \bf 28.77 \\
    \bottomrule
\end{tabular}
\label{tab:open}
\end{table*}

\subsection{Qualitative Results}

\begin{figure*}[t!]
\footnotesize

\centering
\newcommand{\resultwidth}{0.95in}
\begin{tabular}{@{}c@{}c@{}c@{}c@{}c@{}c}
& Normal & Albedo & Roughness & Relighting 1 & Relighting 2 \\

\rotatebox{90}{\,\,\, Single Light} & \adjincludegraphics[width=\resultwidth, trim={0.0in 0.0in 0.0in 0.0in}, clip]{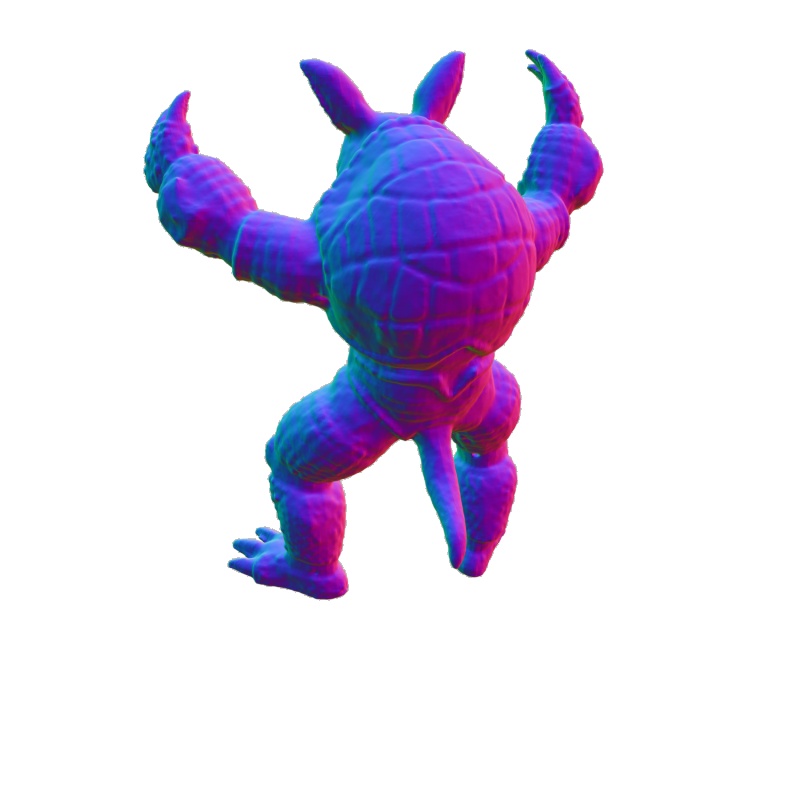} & \adjincludegraphics[width=\resultwidth, trim={0.0in 0.0in 0.0in 0.0in}, clip]{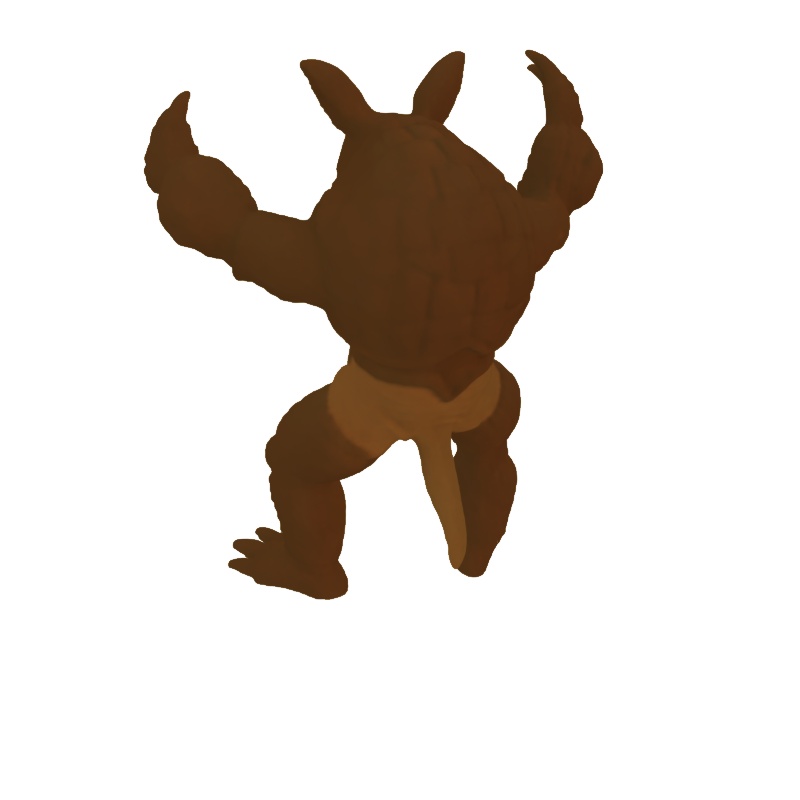} & \adjincludegraphics[width=\resultwidth, trim={0.0in 0.0in 0.0in 0.0in}, clip]{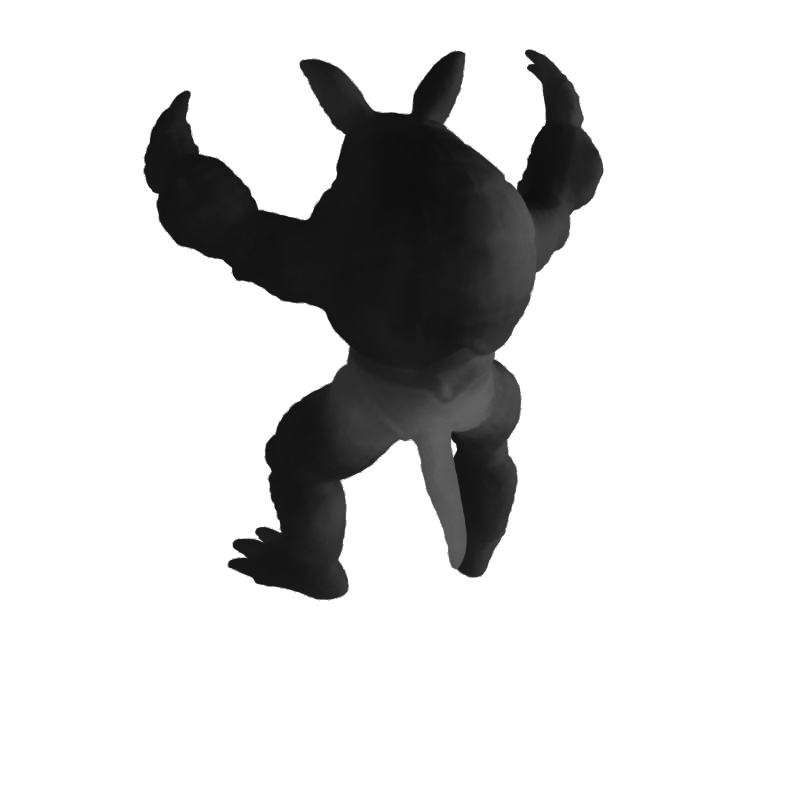} & \adjincludegraphics[width=\resultwidth, trim={0.0in 0.0in 0.0in 0.0in}, clip]{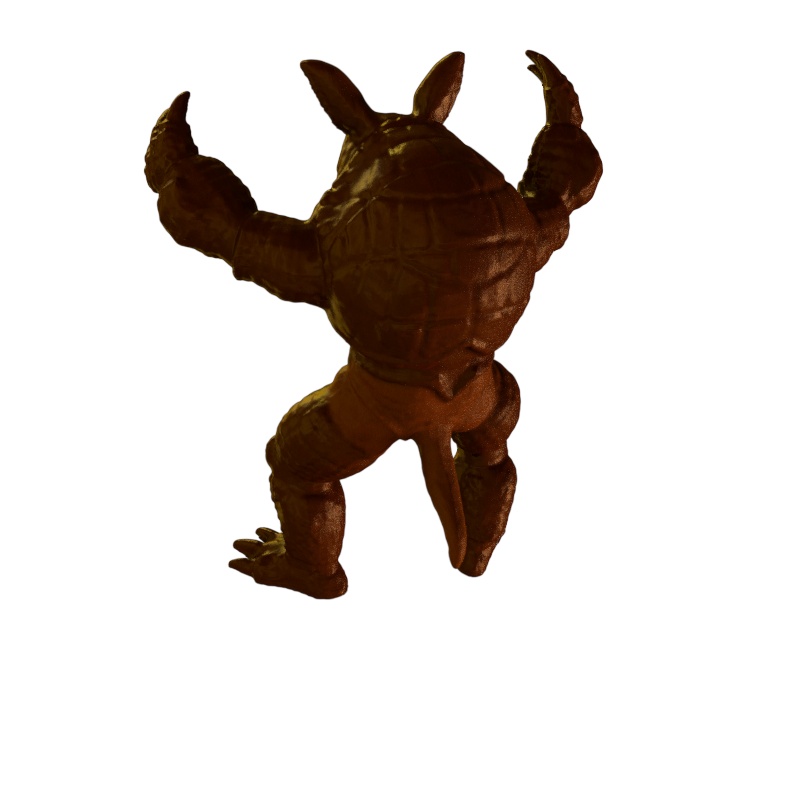} & \adjincludegraphics[width=\resultwidth, trim={0.0in 0.0in 0.0in 0.0in}, clip]{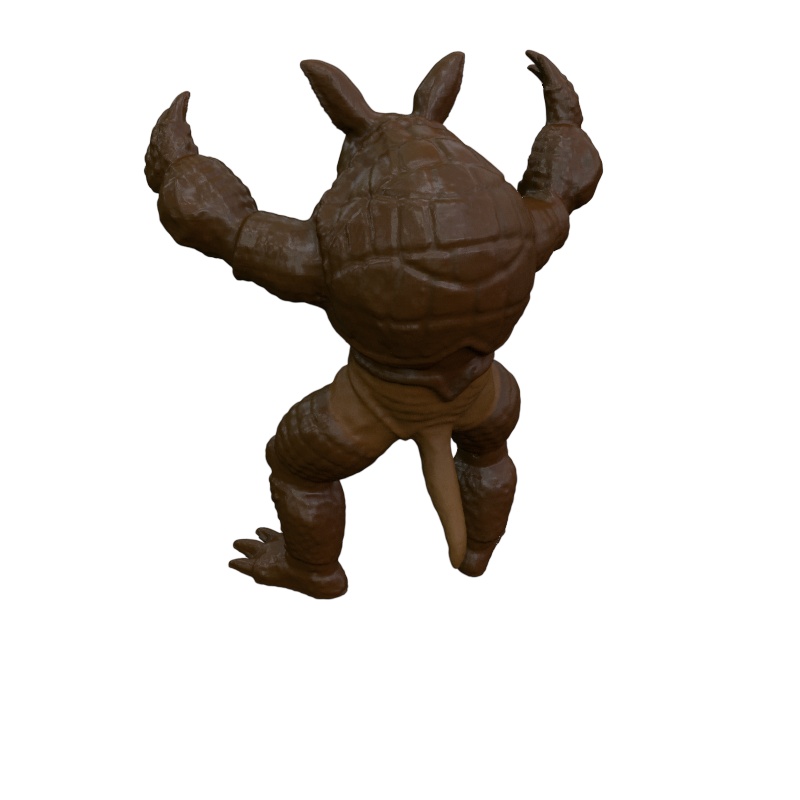} \\

\rotatebox{90}{\,\,\, Multi Light} & \adjincludegraphics[width=\resultwidth, trim={0.0in 0.0in 0.0in 0.0in}, clip]{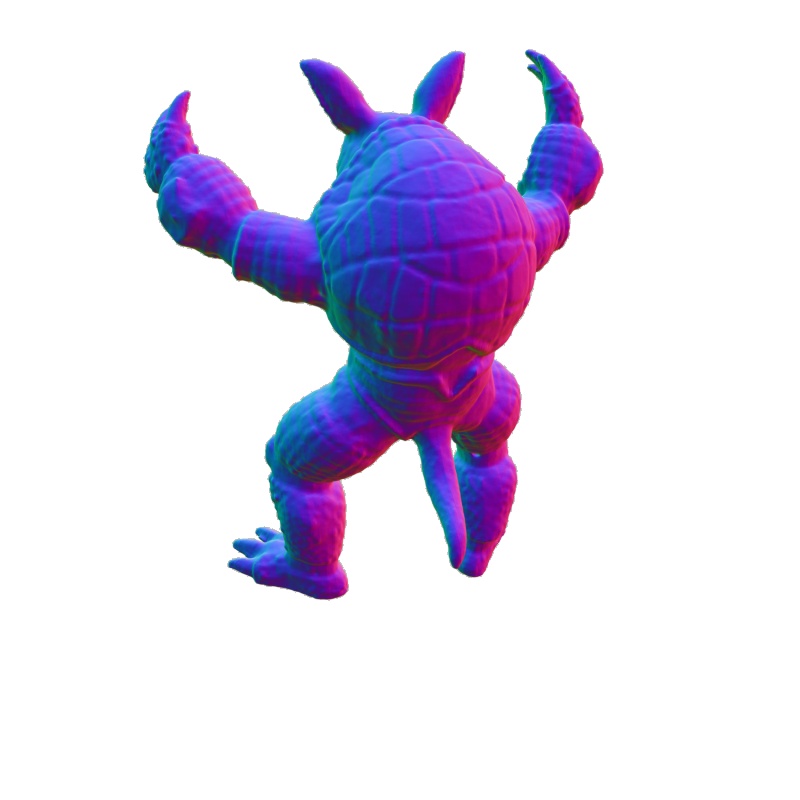} & \adjincludegraphics[width=\resultwidth, trim={0.0in 0.0in 0.0in 0.0in}, clip]{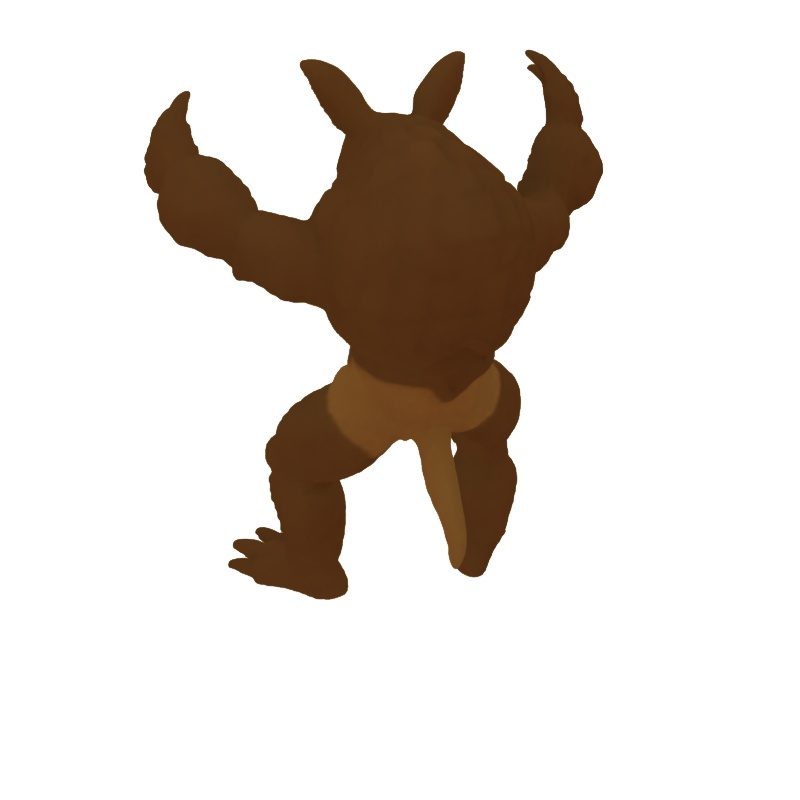} & \adjincludegraphics[width=\resultwidth, trim={0.0in 0.0in 0.0in 0.0in}, clip]{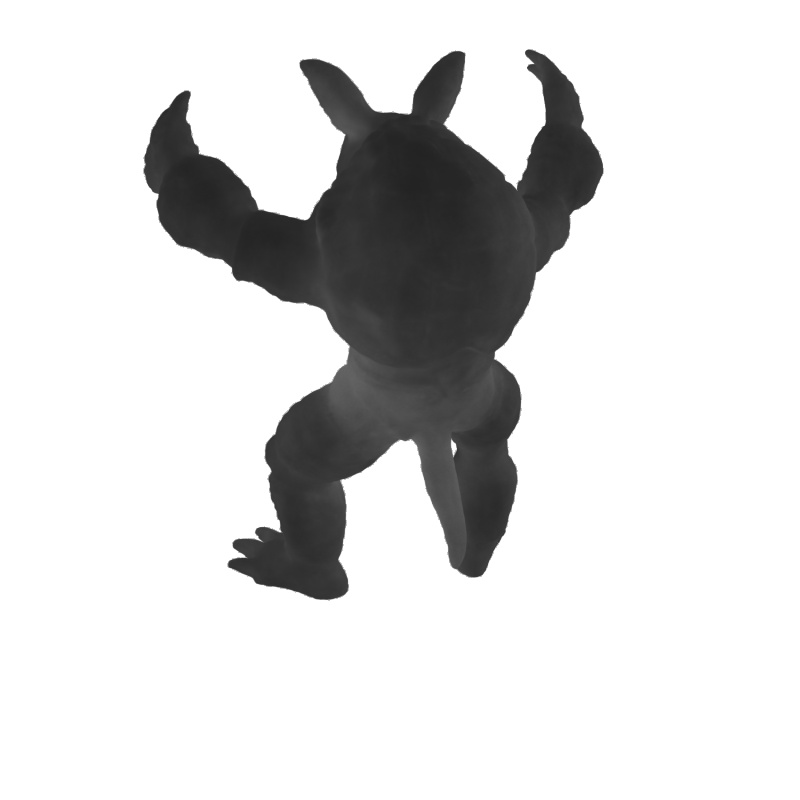} & \adjincludegraphics[width=\resultwidth, trim={0.0in 0.0in 0.0in 0.0in}, clip]{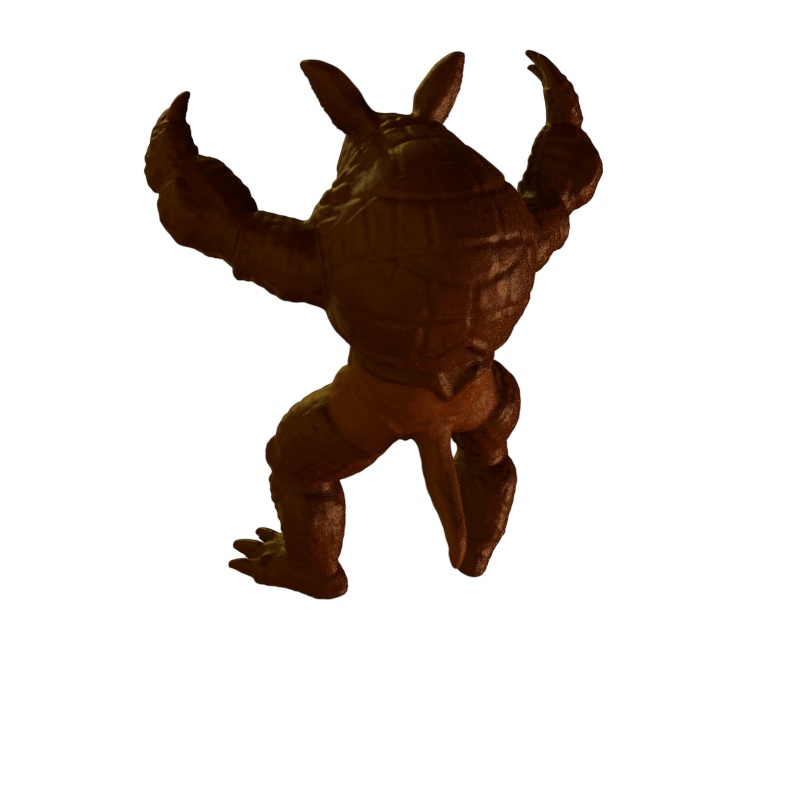} & \adjincludegraphics[width=\resultwidth, trim={0.0in 0.0in 0.0in 0.0in}, clip]{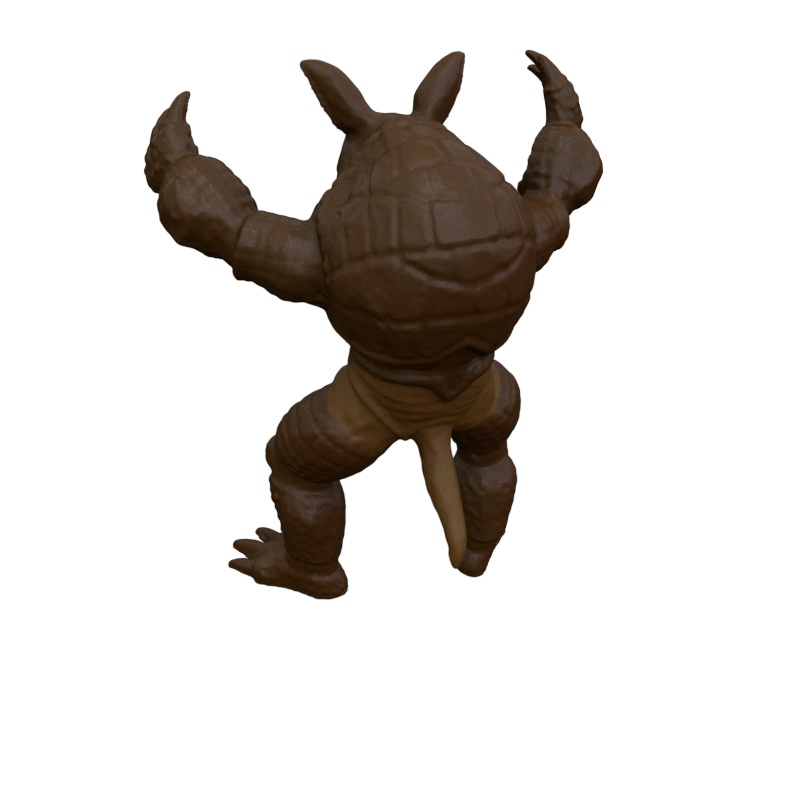} \\

\rotatebox{90}{\,\,\,\,\,\,\,\,\,\, GT} & \adjincludegraphics[width=\resultwidth, trim={0.0in 0.0in 0.0in 0.0in}, clip]{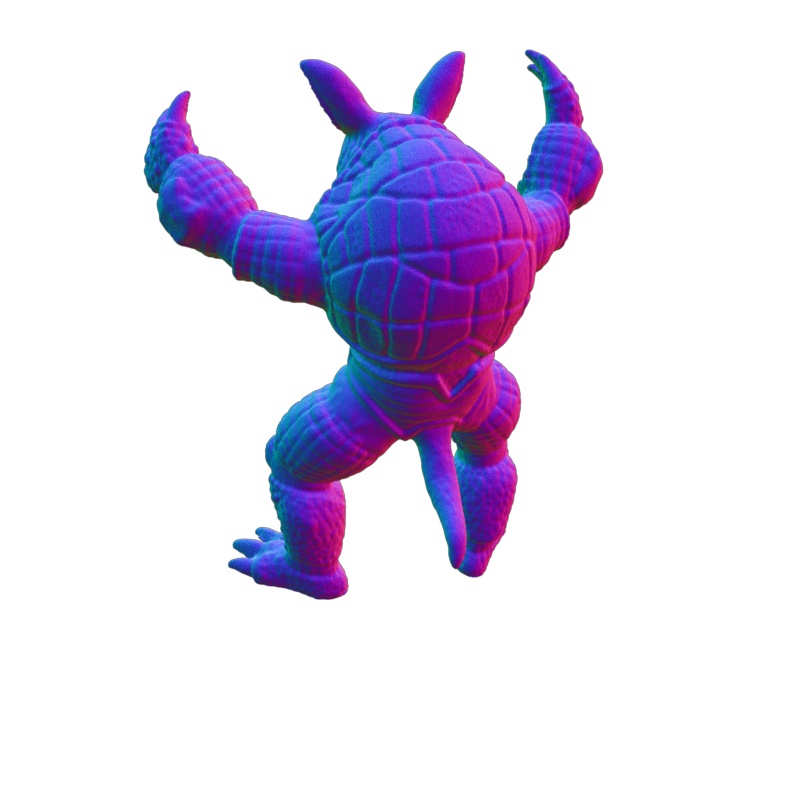} & \adjincludegraphics[width=\resultwidth, trim={0.0in 0.0in 0.0in 0.0in}, clip]{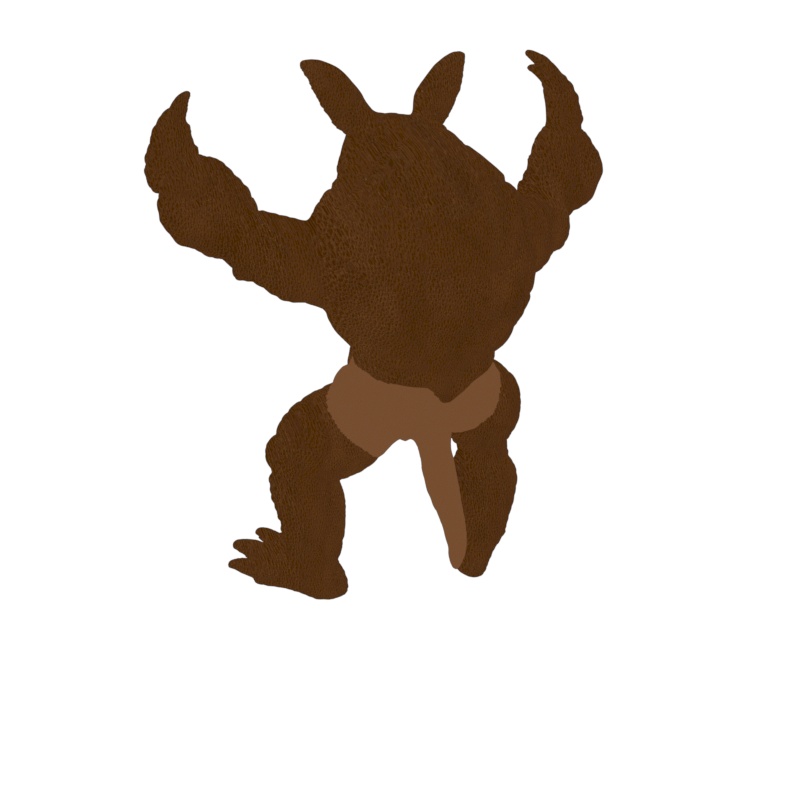} & & \adjincludegraphics[width=\resultwidth, trim={0.0in 0.0in 0.0in 0.0in}, clip]{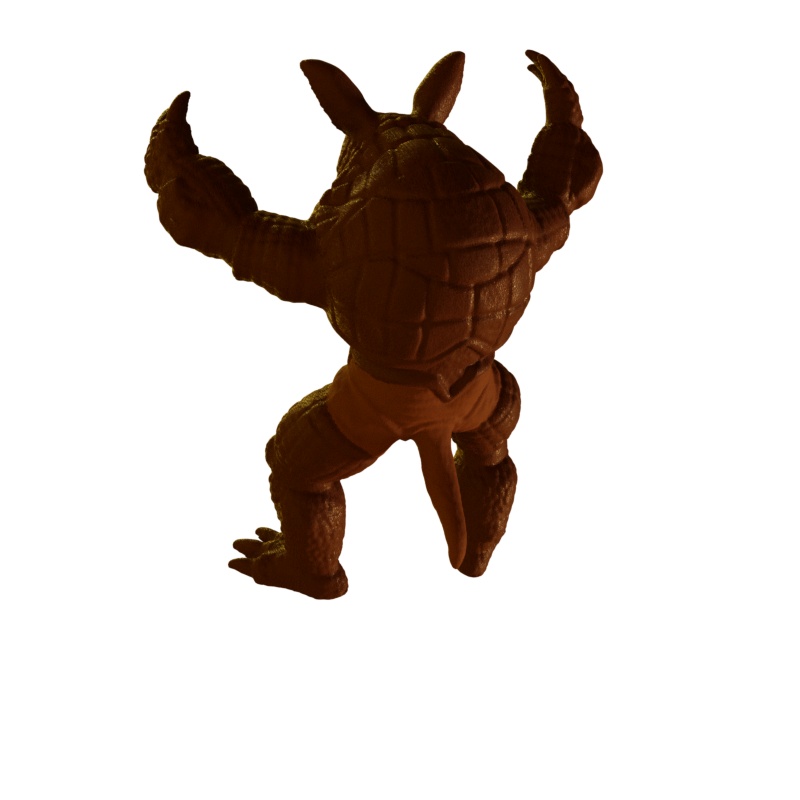} & \adjincludegraphics[width=\resultwidth, trim={0.0in 0.0in 0.0in 0.0in}, clip]{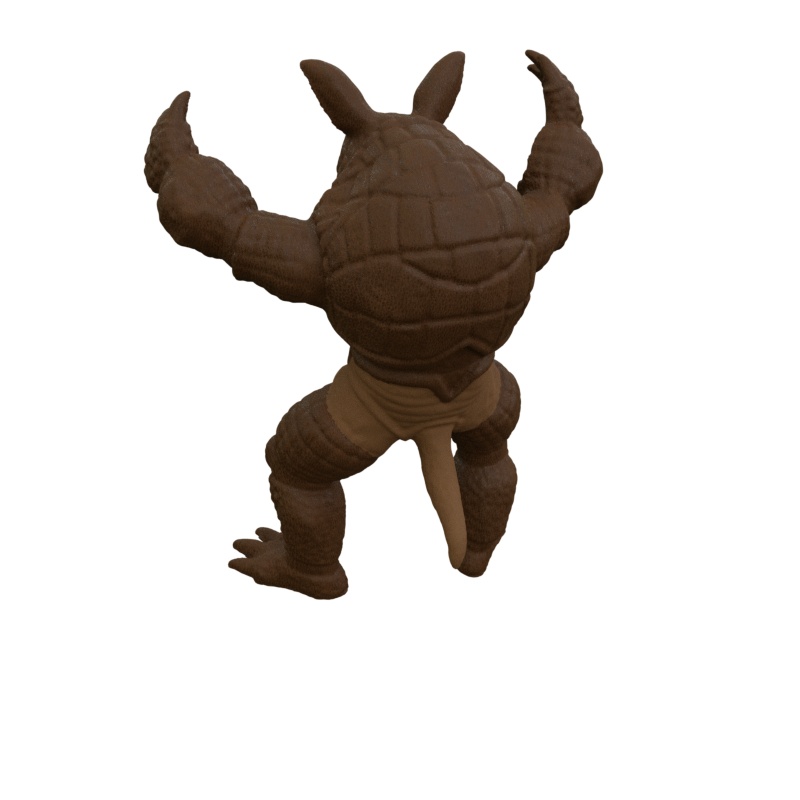} \\

\rotatebox{90}{\,\,\, Single Light} & \adjincludegraphics[width=\resultwidth, trim={0.0in 0.0in 0.0in 0.0in}, clip]{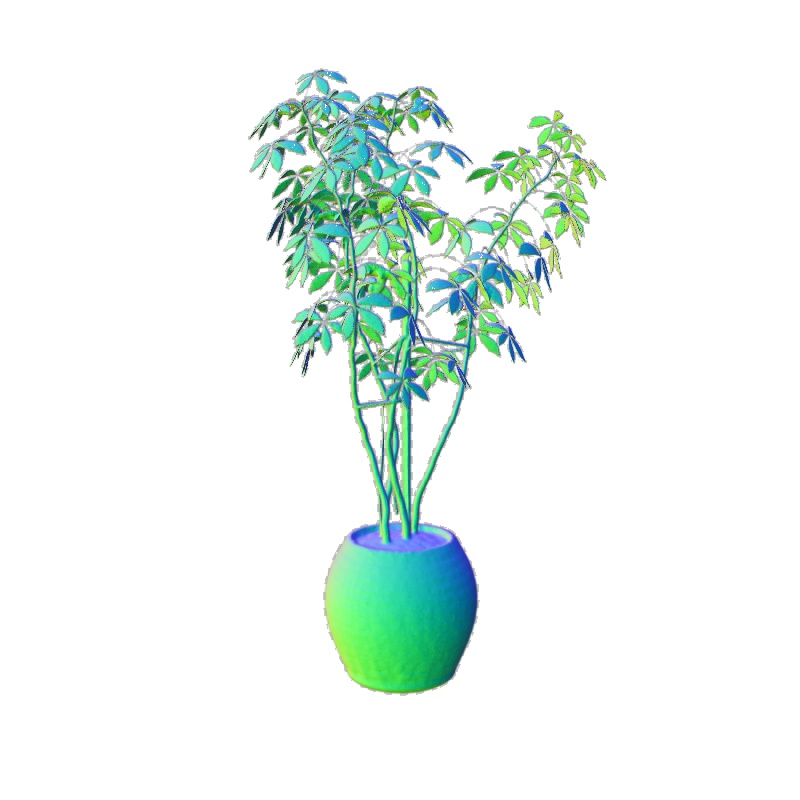} & \adjincludegraphics[width=\resultwidth, trim={0.0in 0.0in 0.0in 0.0in}, clip]{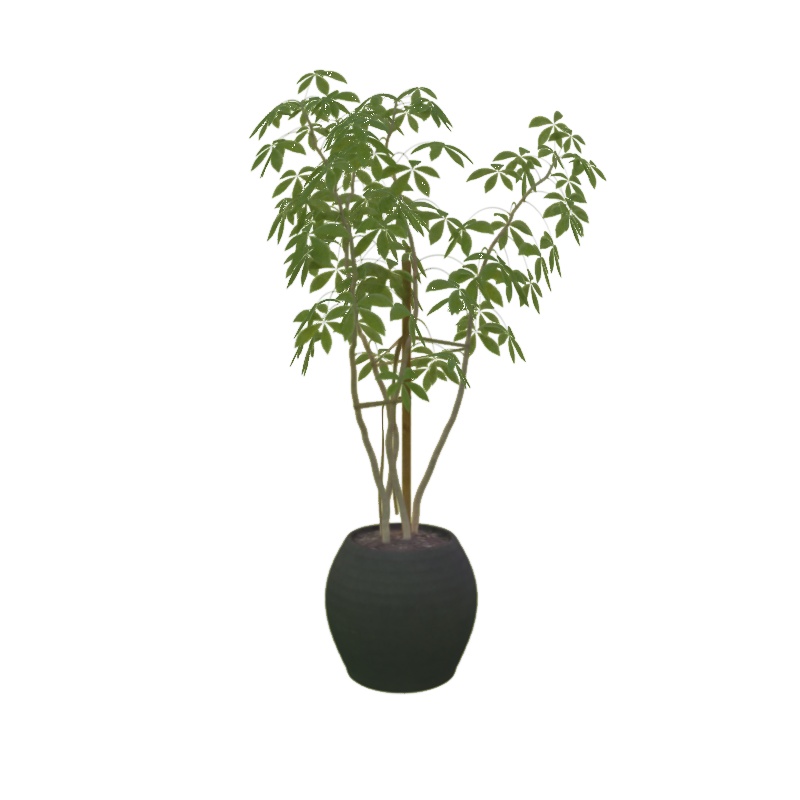} & \adjincludegraphics[width=\resultwidth, trim={0.0in 0.0in 0.0in 0.0in}, clip]{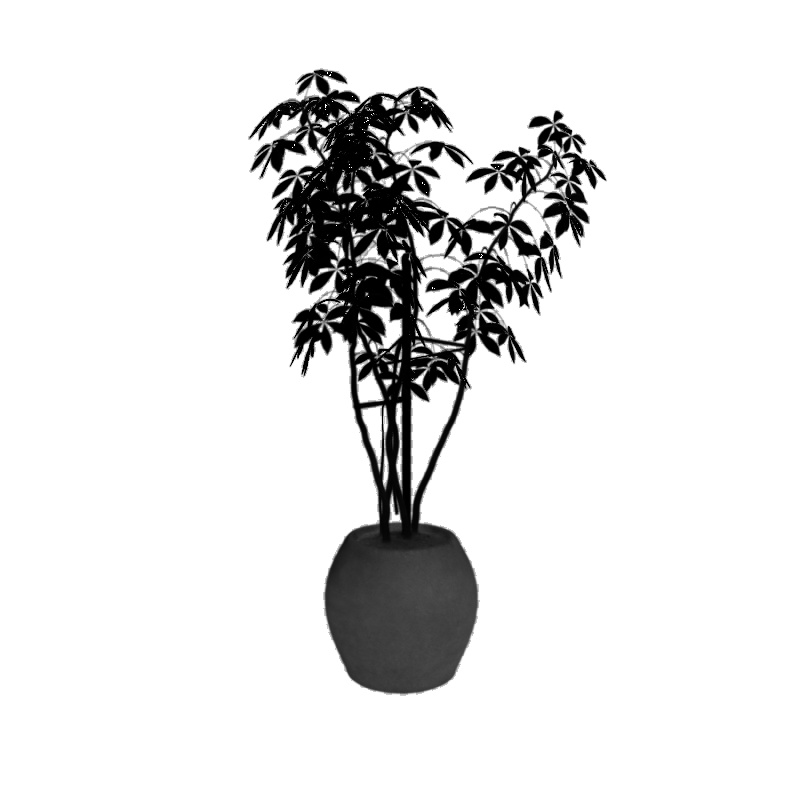} & \adjincludegraphics[width=\resultwidth, trim={0.0in 0.0in 0.0in 0.0in}, clip]{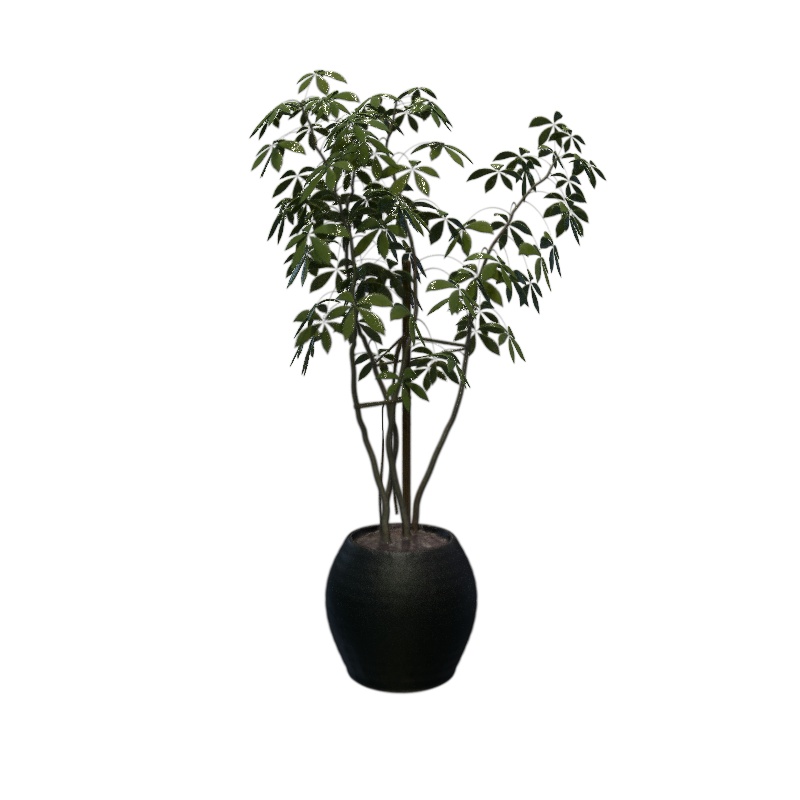} & \adjincludegraphics[width=\resultwidth, trim={0.0in 0.0in 0.0in 0.0in}, clip]{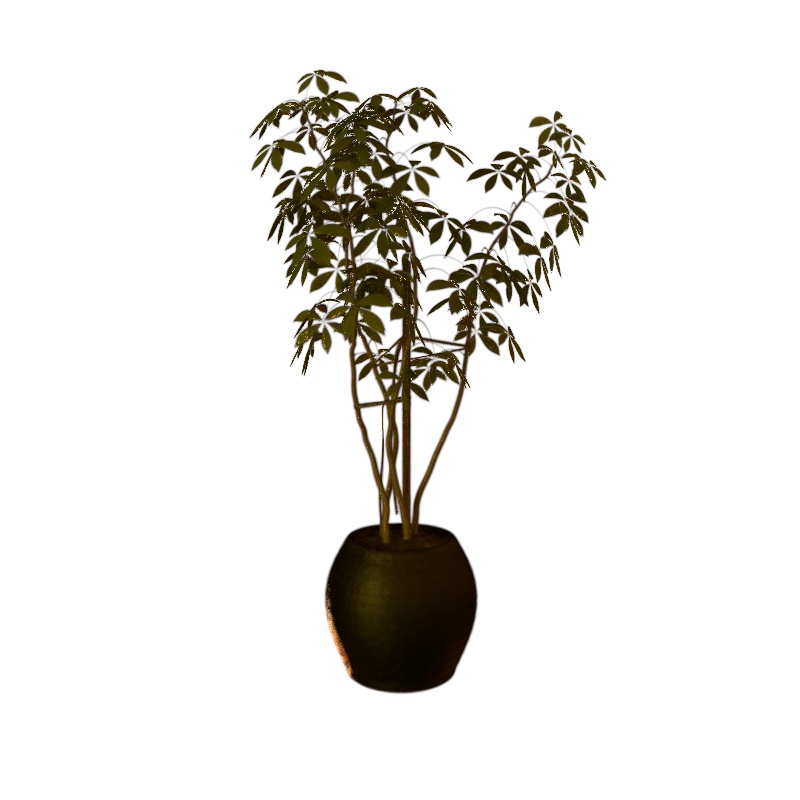} \\

\rotatebox{90}{\,\,\, Multi Light} & \adjincludegraphics[width=\resultwidth, trim={0.0in 0.0in 0.0in 0.0in}, clip]{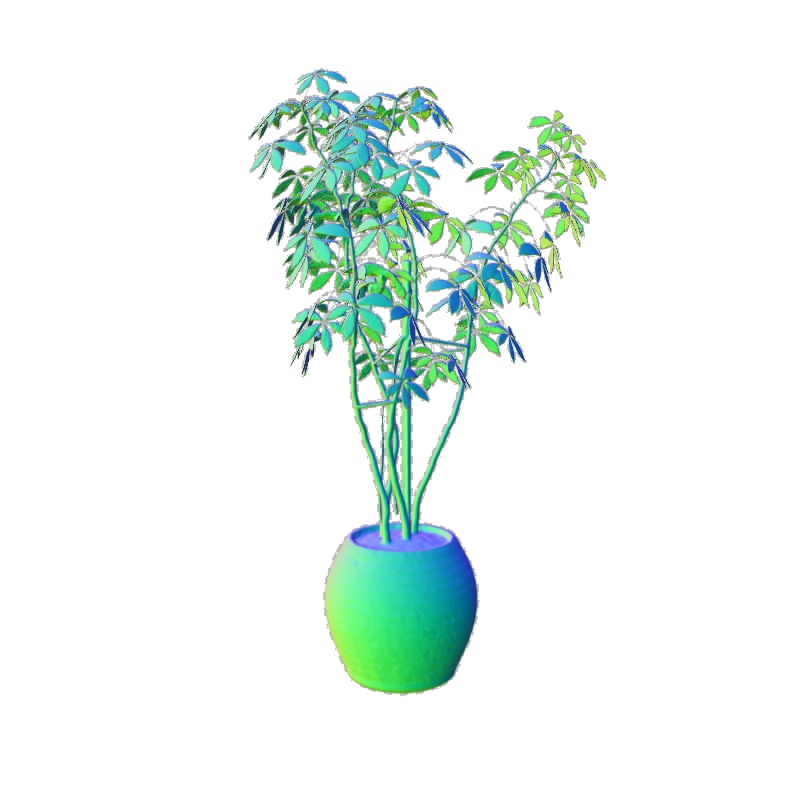} & \adjincludegraphics[width=\resultwidth, trim={0.0in 0.0in 0.0in 0.0in}, clip]{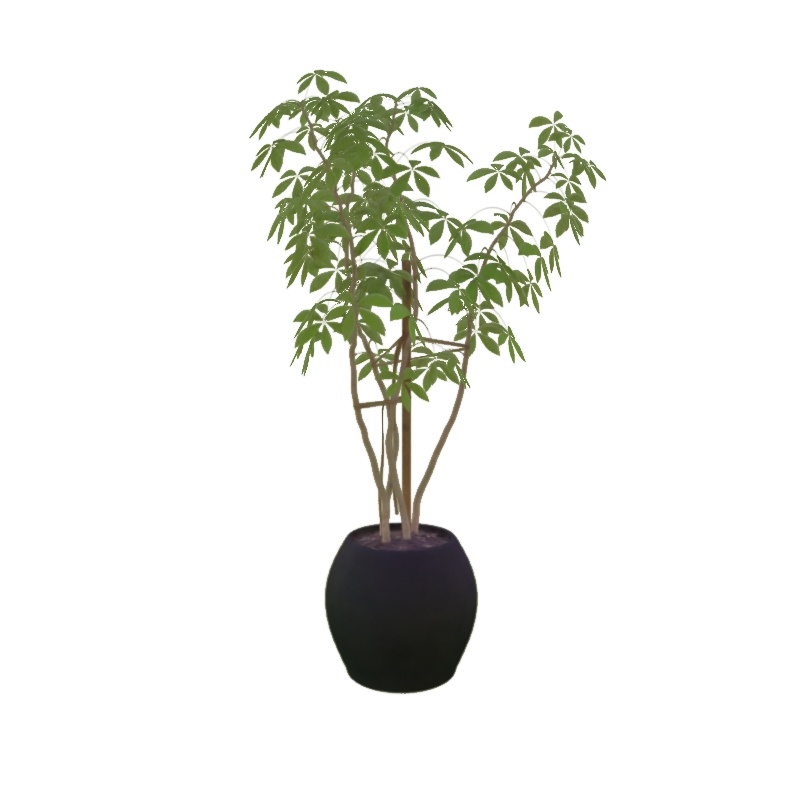} & \adjincludegraphics[width=\resultwidth, trim={0.0in 0.0in 0.0in 0.0in}, clip]{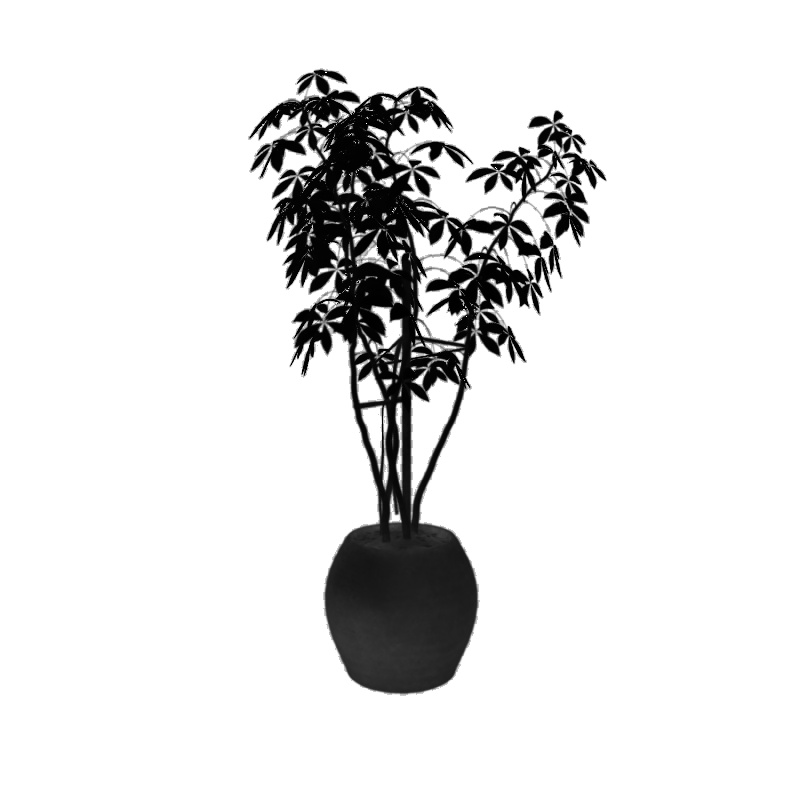} & \adjincludegraphics[width=\resultwidth, trim={0.0in 0.0in 0.0in 0.0in}, clip]{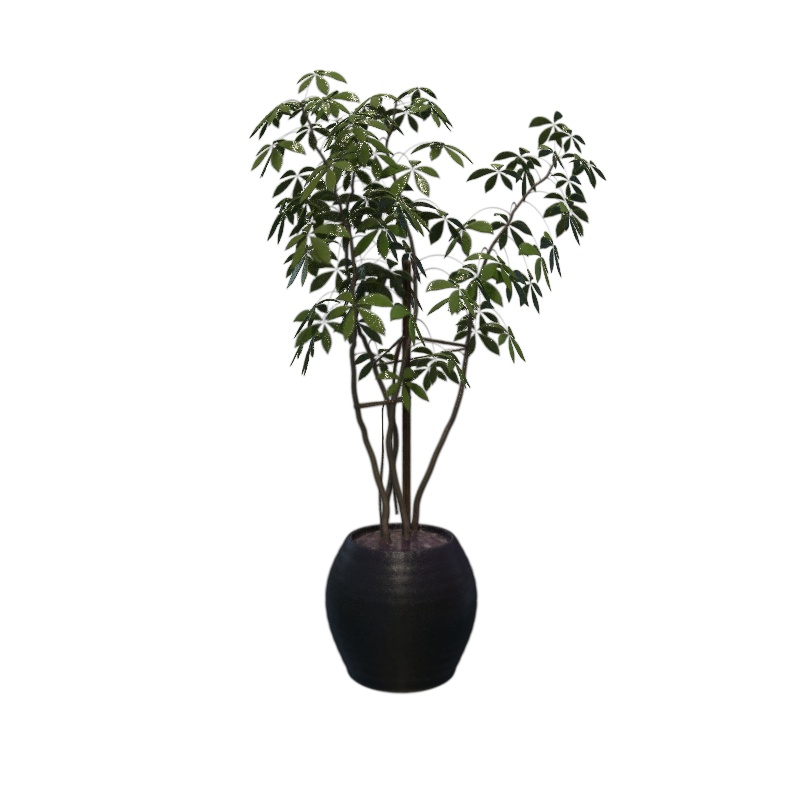} & \adjincludegraphics[width=\resultwidth, trim={0.0in 0.0in 0.0in 0.0in}, clip]{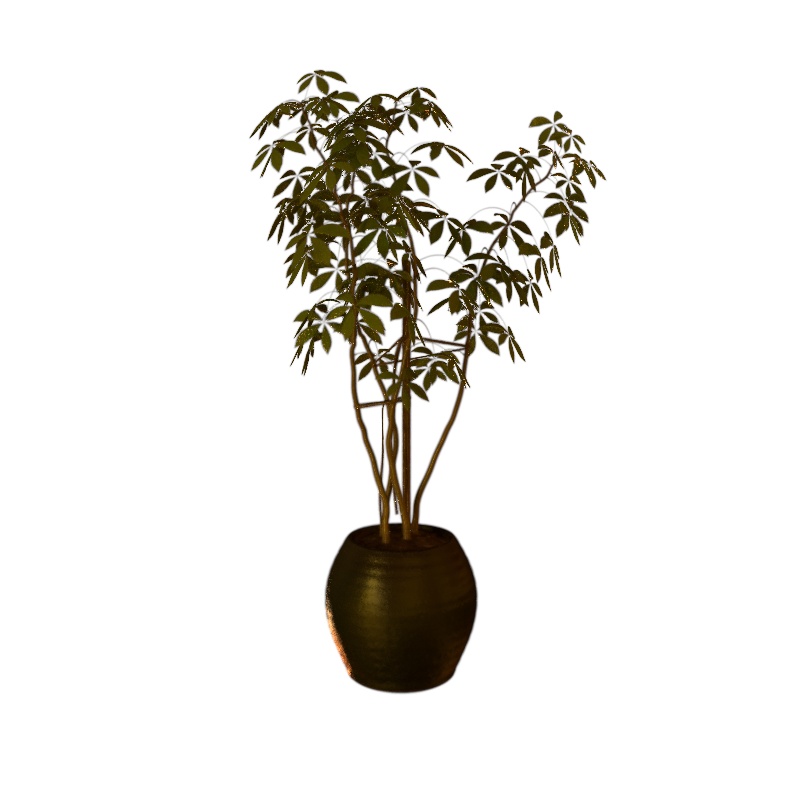} \\

\rotatebox{90}{\,\,\,\,\,\,\,\,\,\, GT} & \adjincludegraphics[width=\resultwidth, trim={0.0in 0.0in 0.0in 0.0in}, clip]{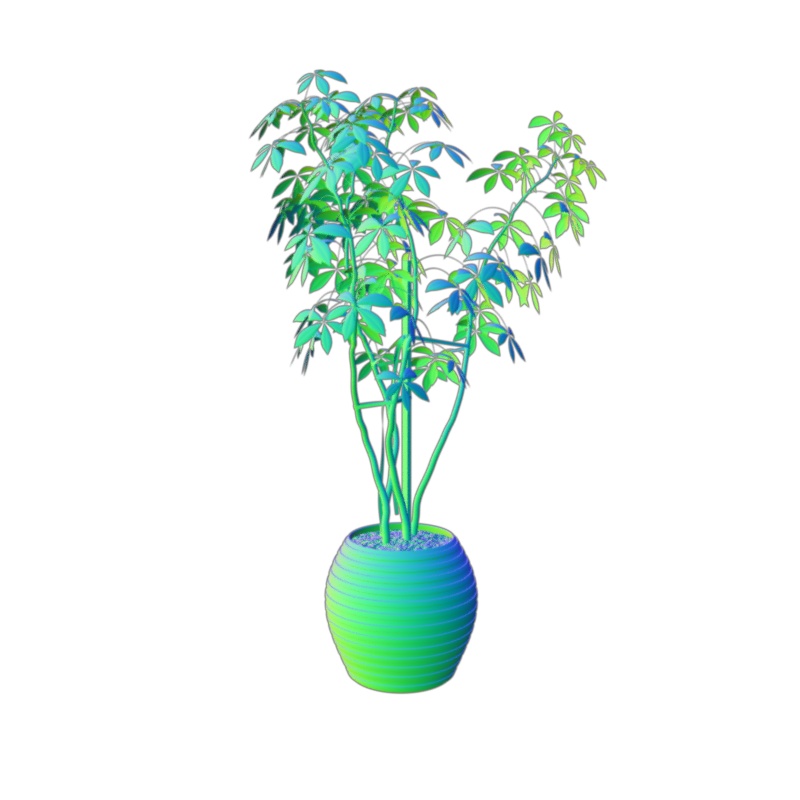} & \adjincludegraphics[width=\resultwidth, trim={0.0in 0.0in 0.0in 0.0in}, clip]{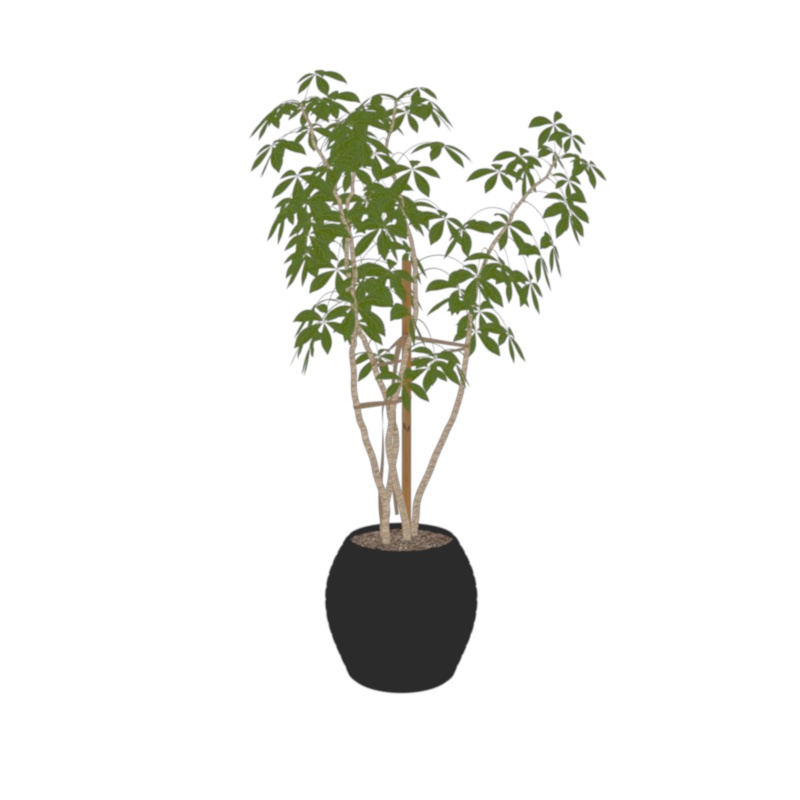} & & \adjincludegraphics[width=\resultwidth, trim={0.0in 0.0in 0.0in 0.0in}, clip]{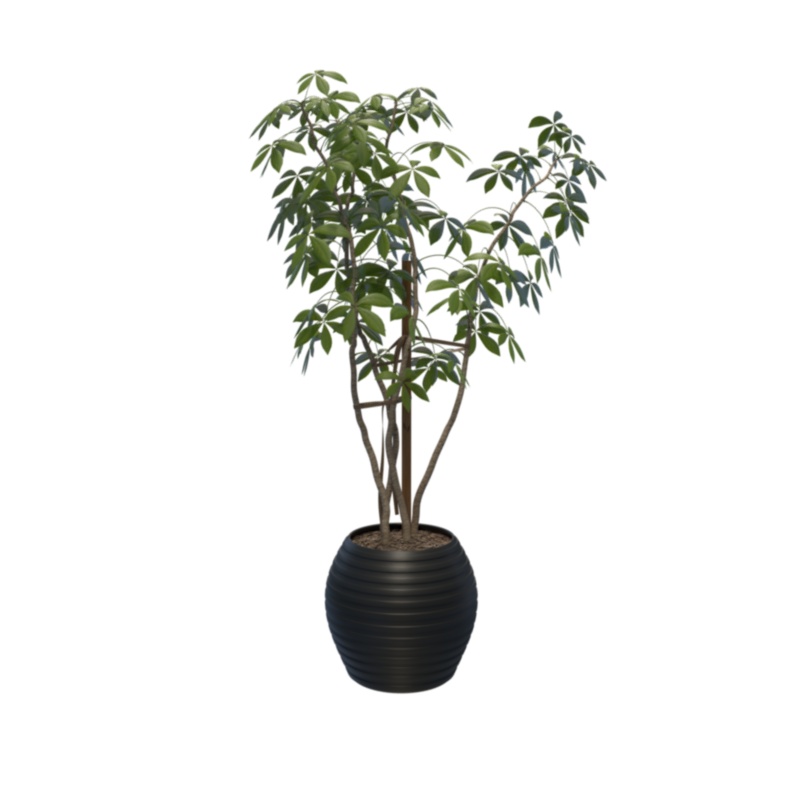} & \adjincludegraphics[width=\resultwidth, trim={0.0in 0.0in 0.0in 0.0in}, clip]{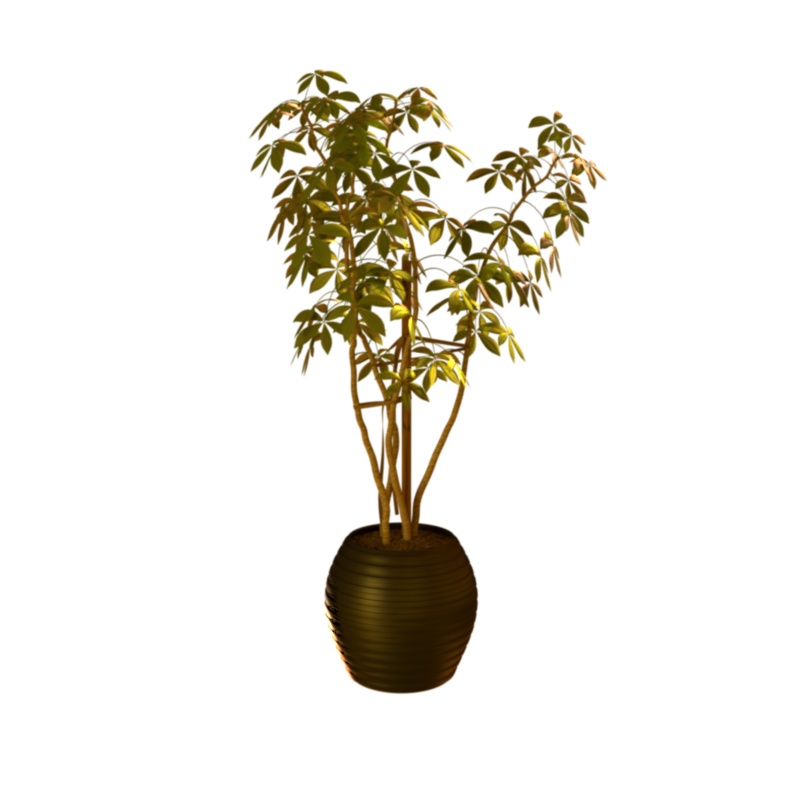}

\end{tabular}
\vspace{-0.1in}
\caption{%
\textbf{Additional TensoIR-Synthetic Results~\cite{jin2023tensoir} Results.}
}%
\label{fig:qual-tensoir}
\end{figure*}

\begin{figure*}[t!]
\footnotesize
\centering
\newcommand{\resultwidth}{0.92in}
\begin{tabular}{@{}c@{}c@{}c@{}c@{}c@{}c}
& Normal & Albedo & Roughness & Relighting 1 & Relighting 2 \\

\rotatebox{90}{\,\,\,\,\,\,\,\,\,\,\,\,\,\,\,\,\,\,\,\, Bird} \hspace{1mm} & \adjincludegraphics[width=\resultwidth, trim={0.0in 0.0in 0.0in 0.0in}, clip]{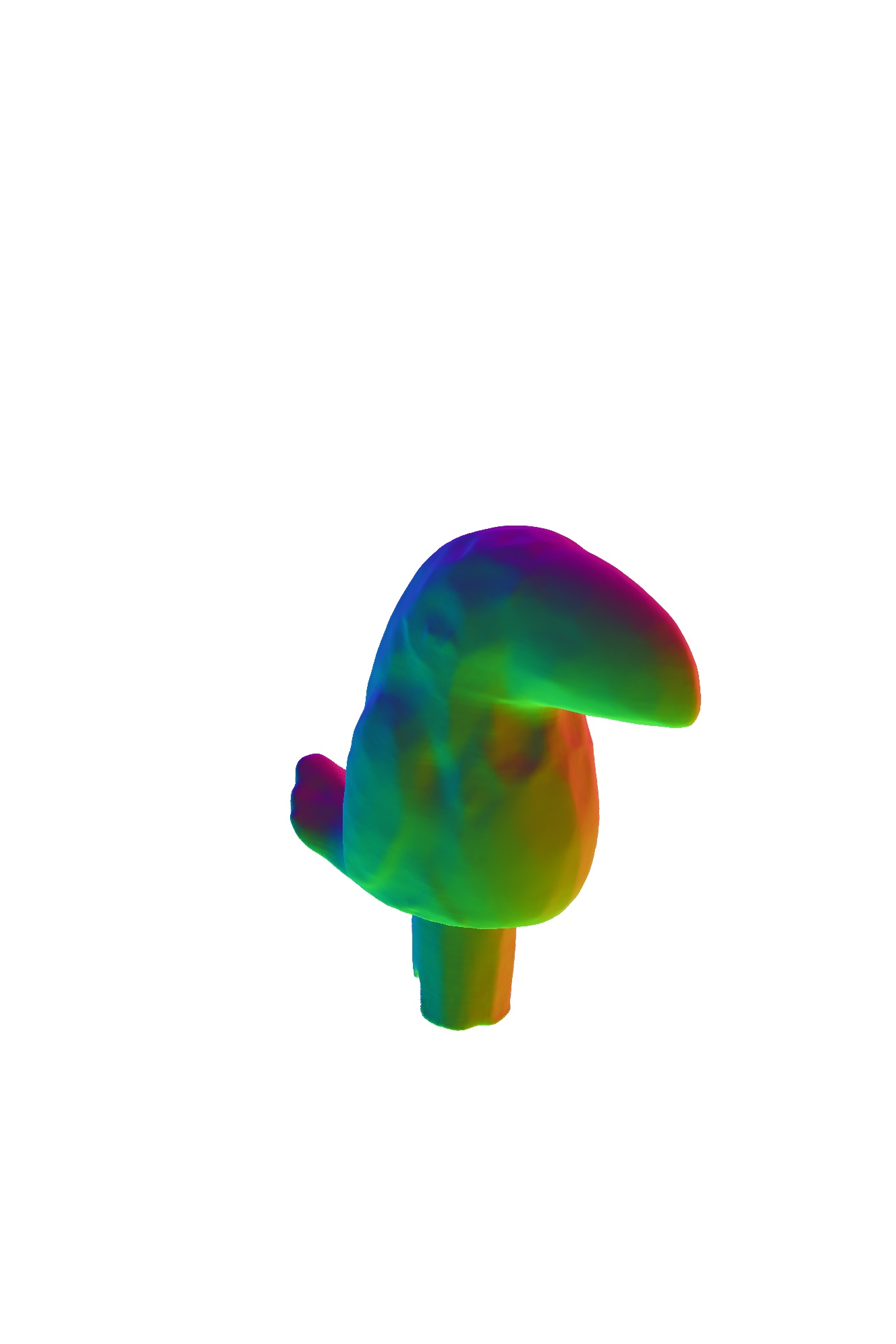} & \adjincludegraphics[width=\resultwidth, trim={0.0in 0.0in 0.0in 0.0in}, clip]{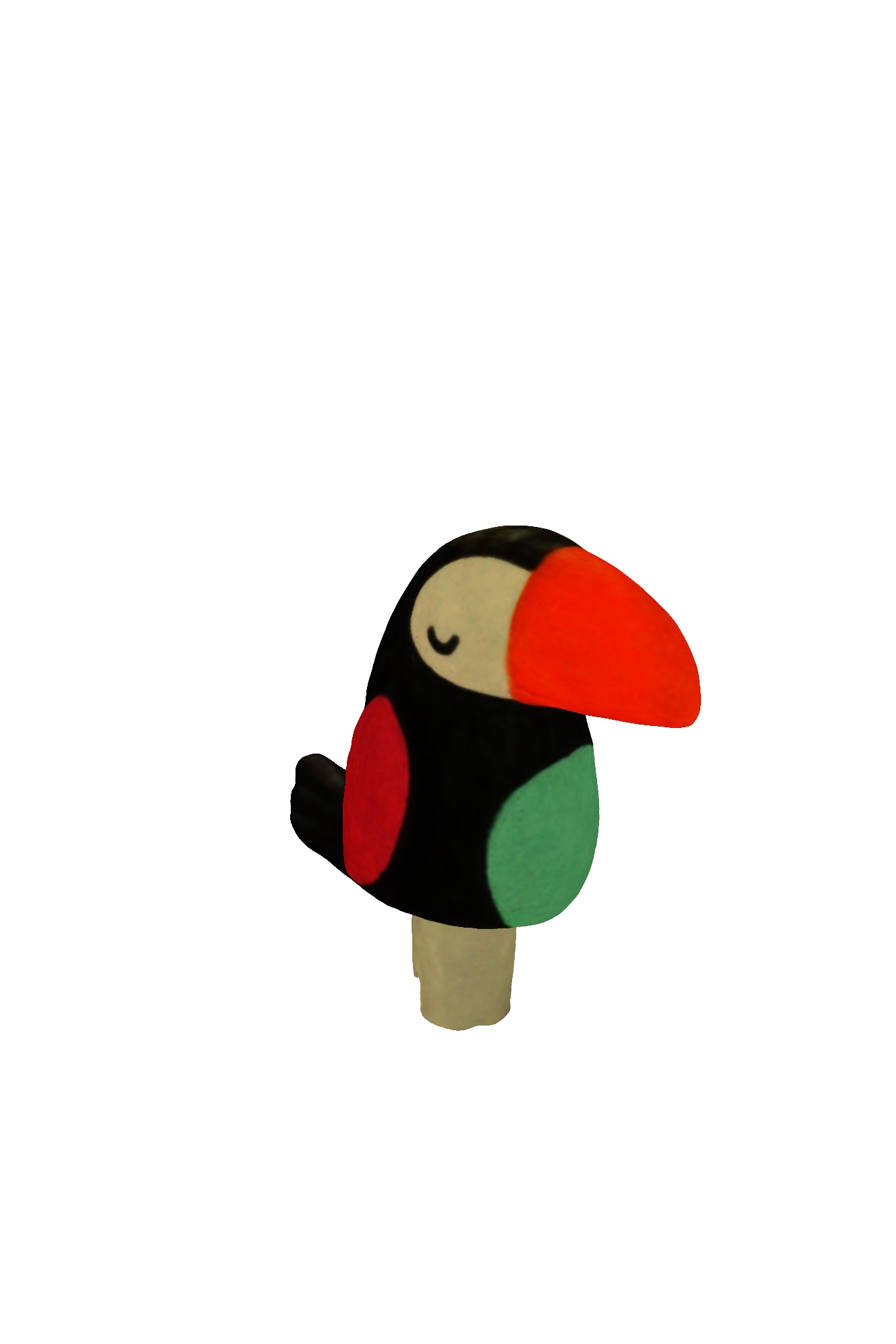} & \adjincludegraphics[width=\resultwidth, trim={0.0in 0.0in 0.0in 0.0in}, clip]{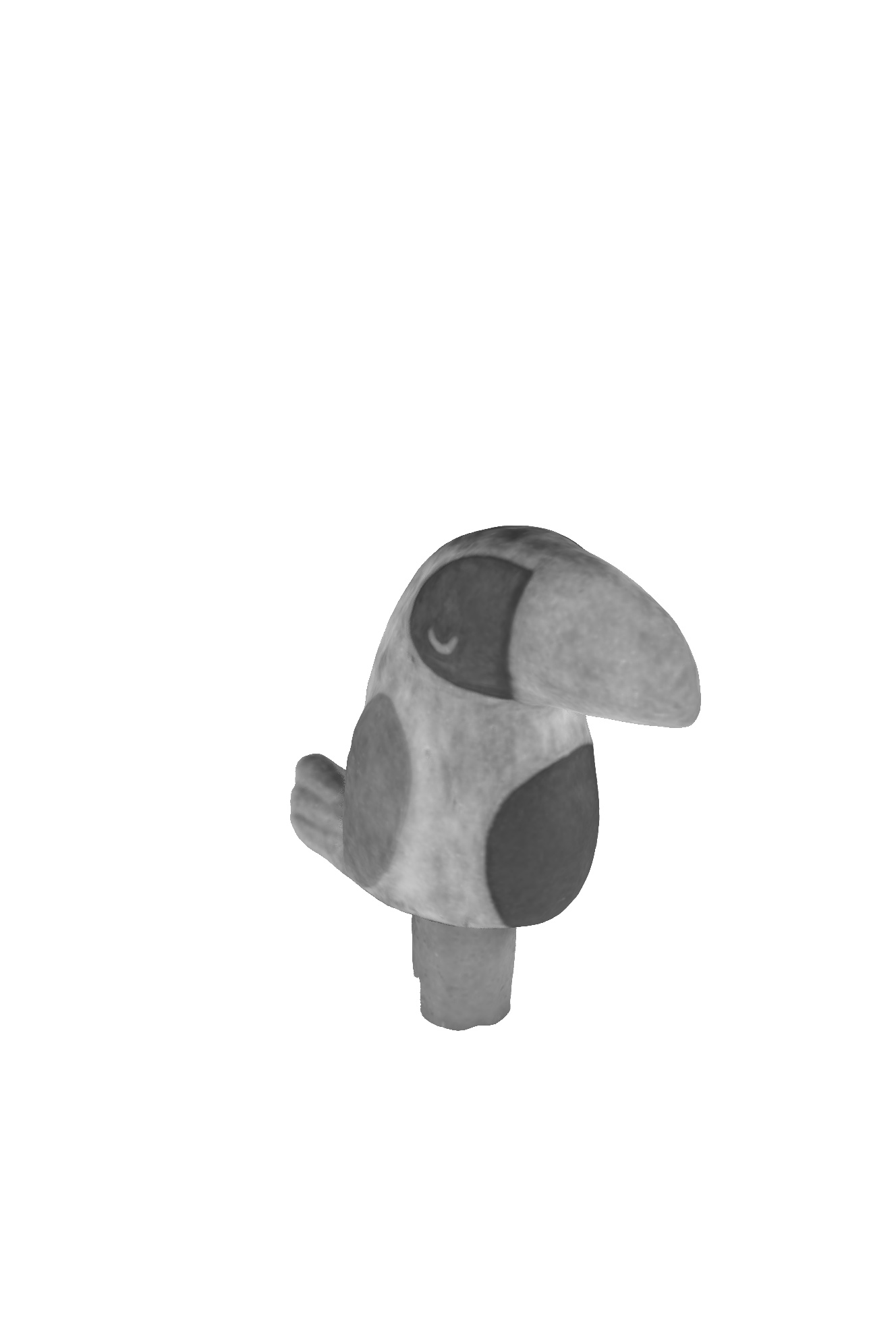} & \adjincludegraphics[width=\resultwidth, trim={0.0in 0.0in 0.0in 0.0in}, clip]{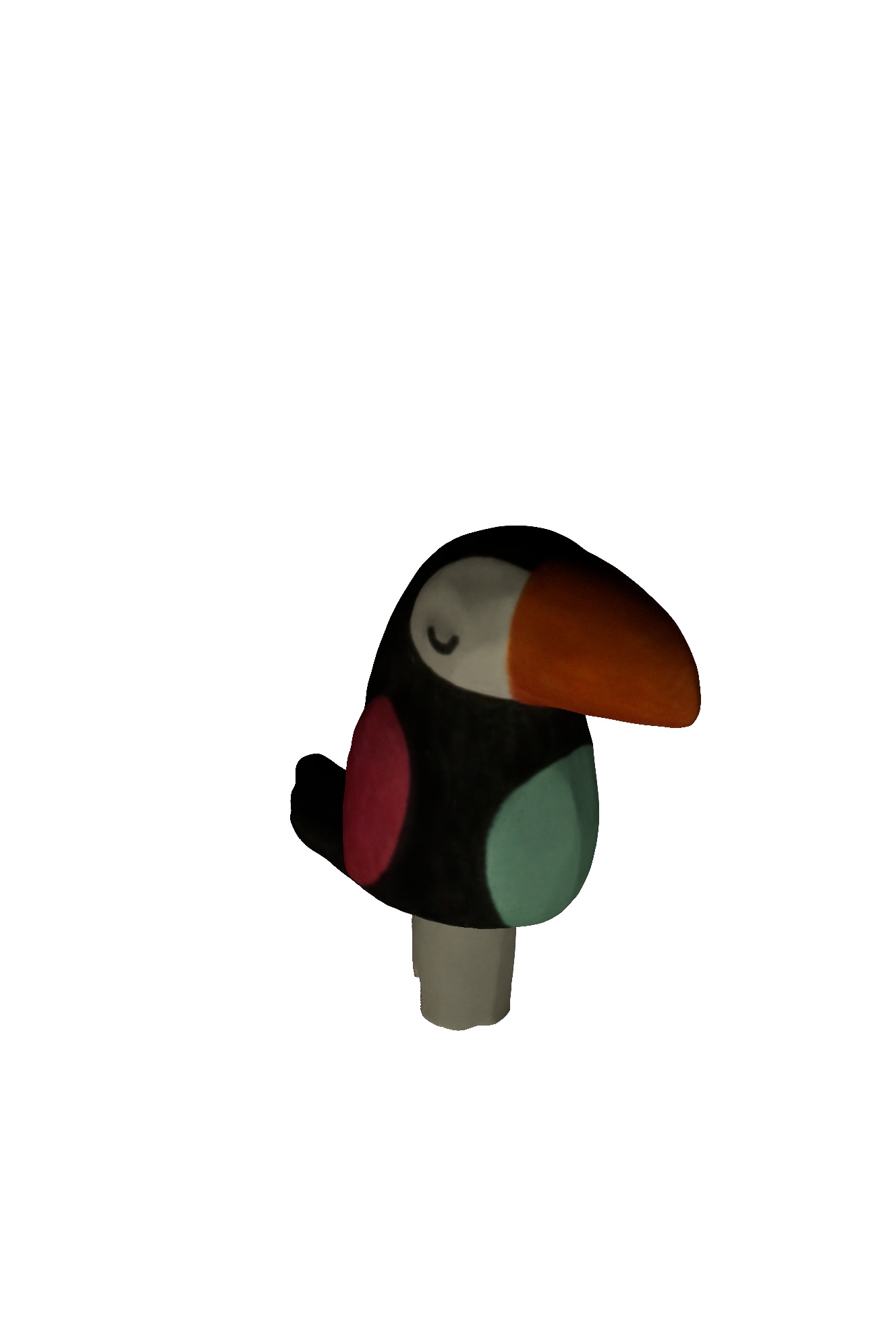} & \adjincludegraphics[width=\resultwidth, trim={0.0in 0.0in 0.0in 0.0in}, clip]{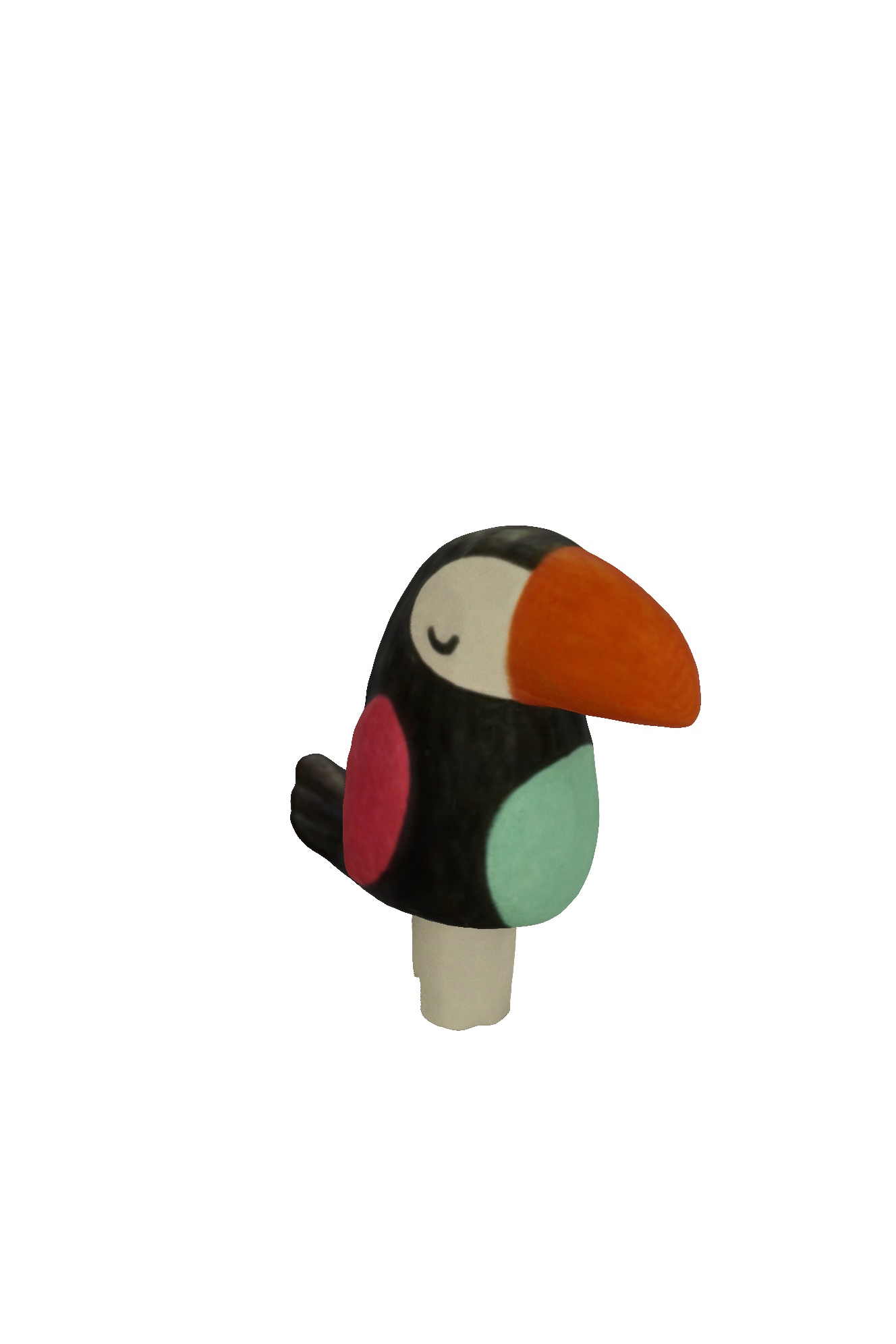} \\

\rotatebox{90}{\,\,\,\,\,\,\,\,\,\,\,\,\,\,\,\,\,\,\,\, Box} \hspace{1mm} & \adjincludegraphics[width=\resultwidth, trim={0.0in 0.0in 0.0in 0.0in}, clip]{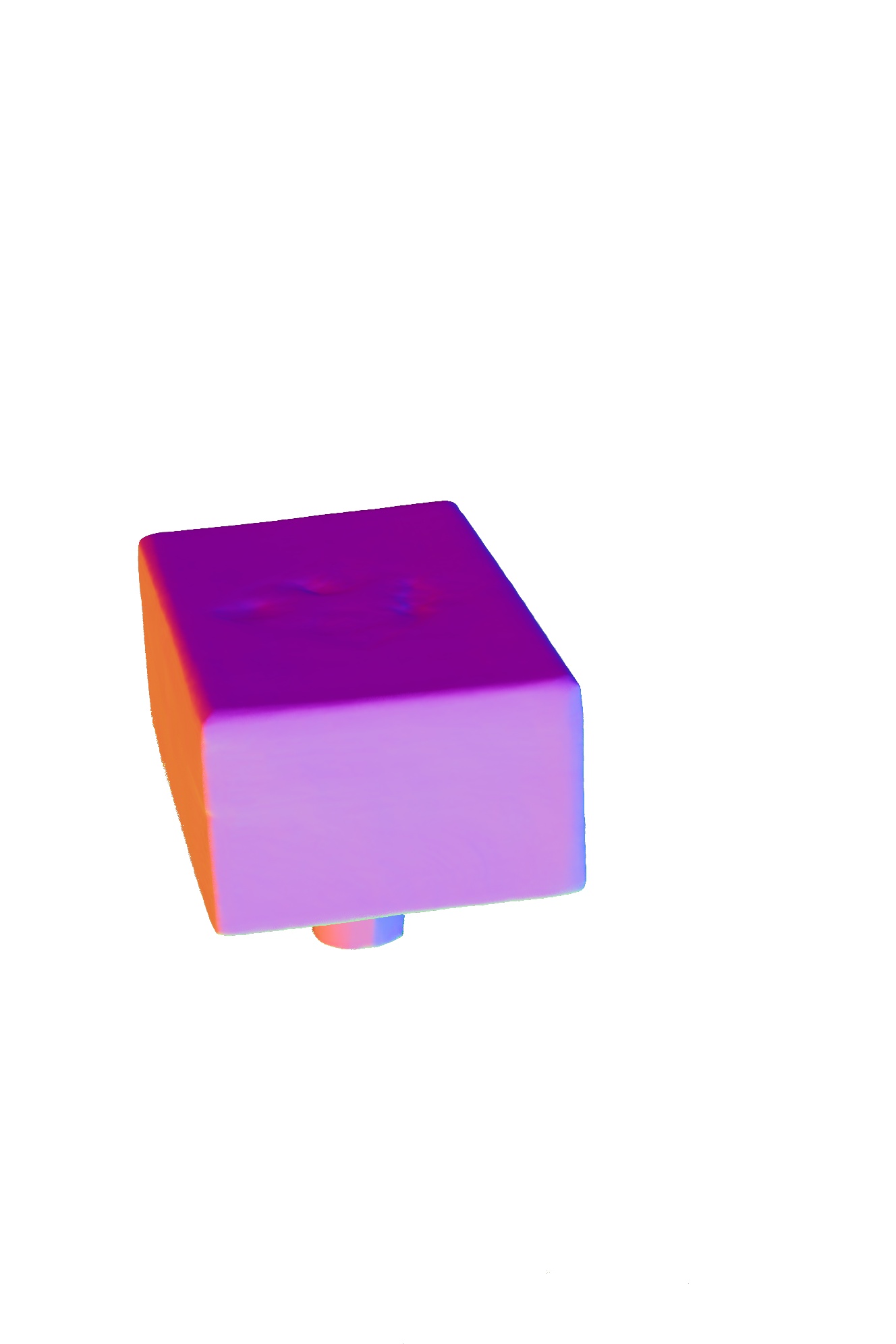} & \adjincludegraphics[width=\resultwidth, trim={0.0in 0.0in 0.0in 0.0in}, clip]{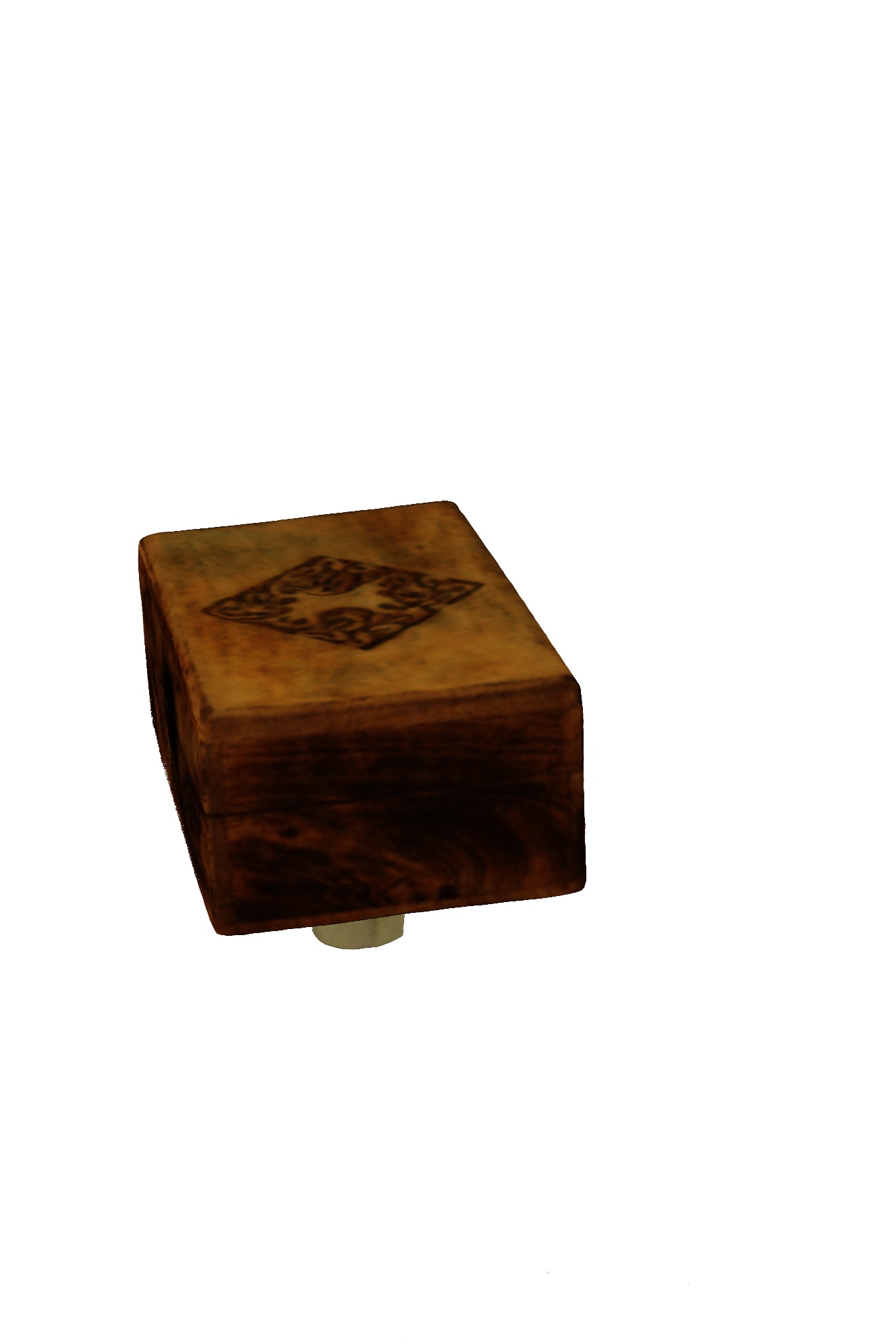} & \adjincludegraphics[width=\resultwidth, trim={0.0in 0.0in 0.0in 0.0in}, clip]{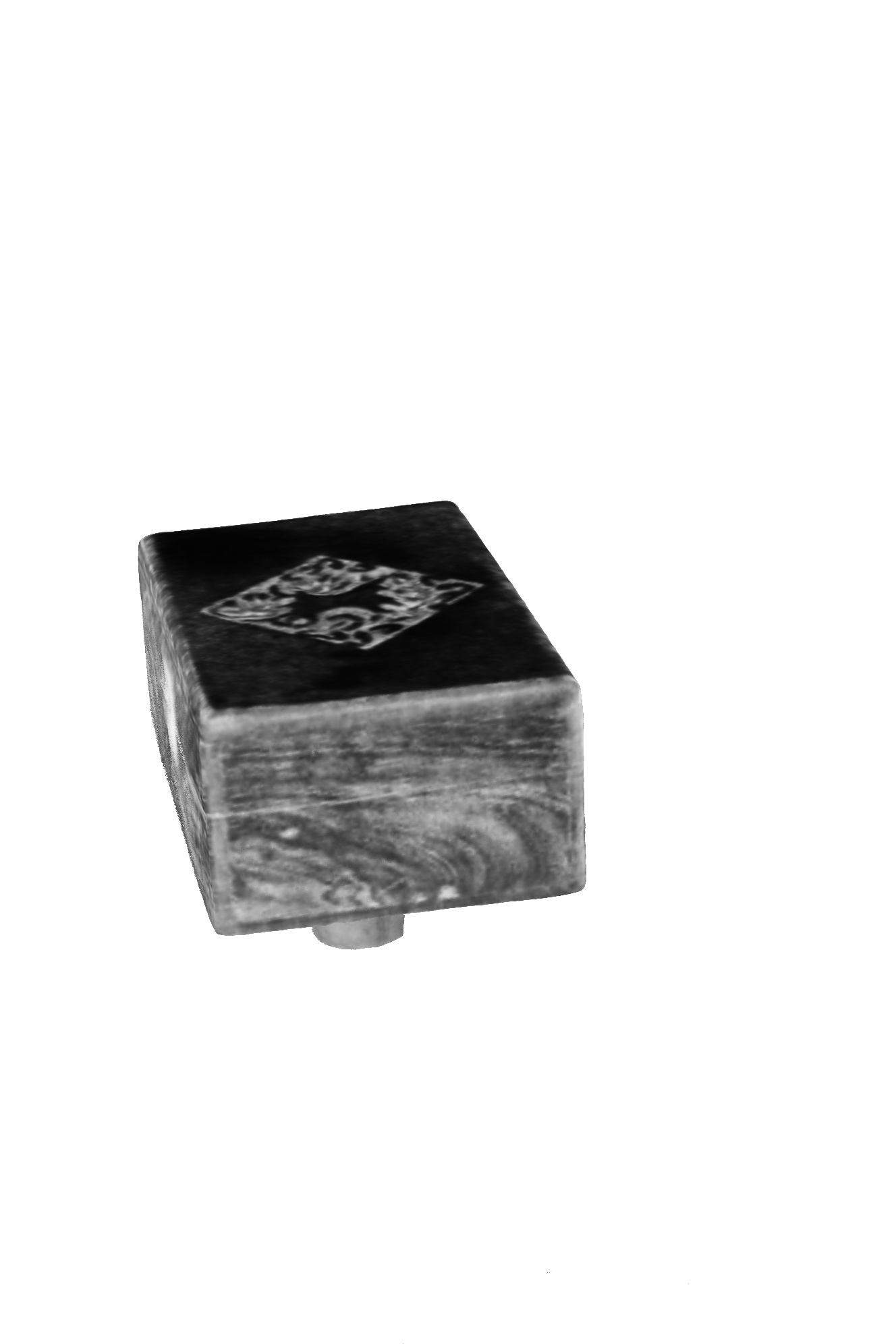} & \adjincludegraphics[width=\resultwidth, trim={0.0in 0.0in 0.0in 0.0in}, clip]{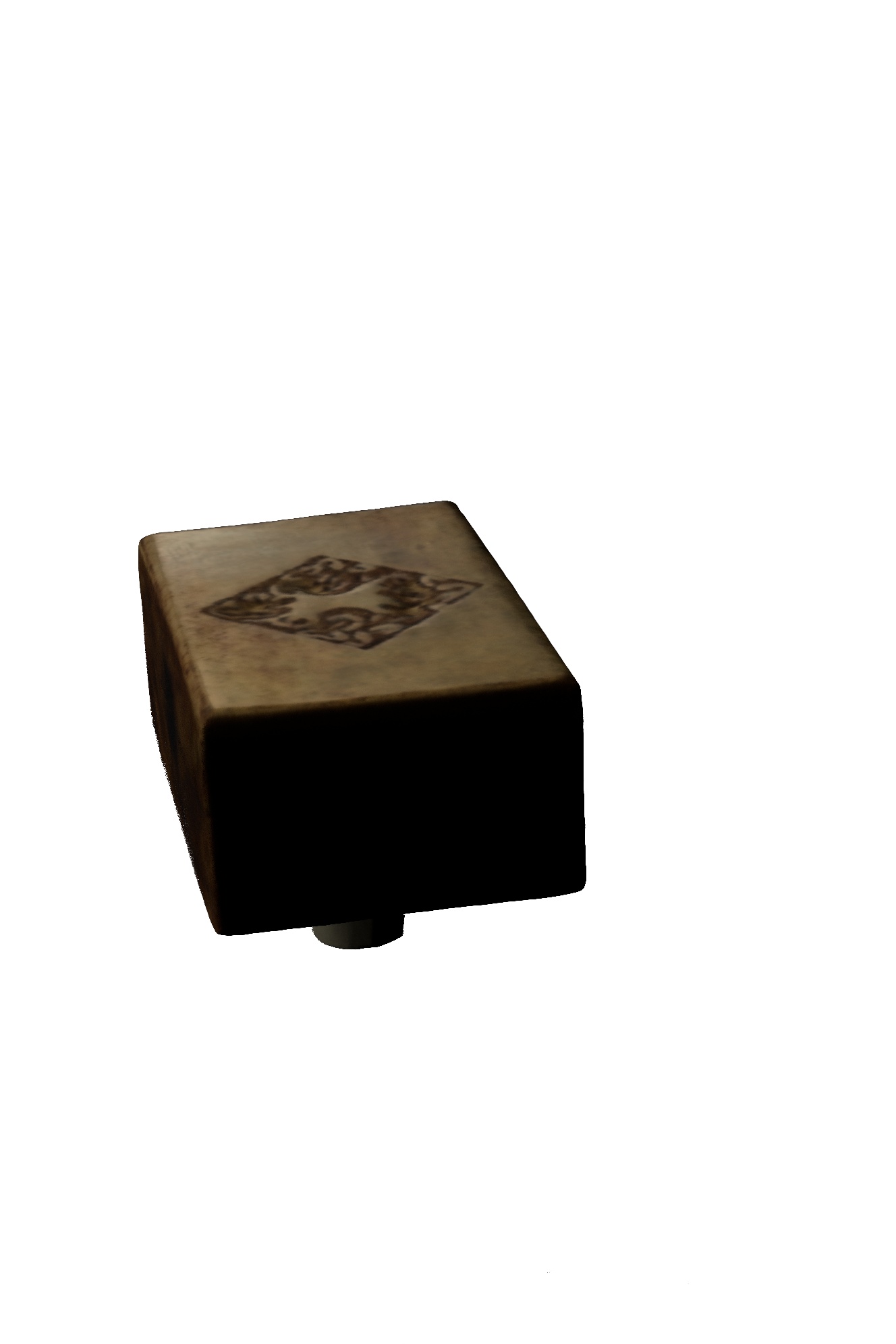} & \adjincludegraphics[width=\resultwidth, trim={0.0in 0.0in 0.0in 0.0in}, clip]{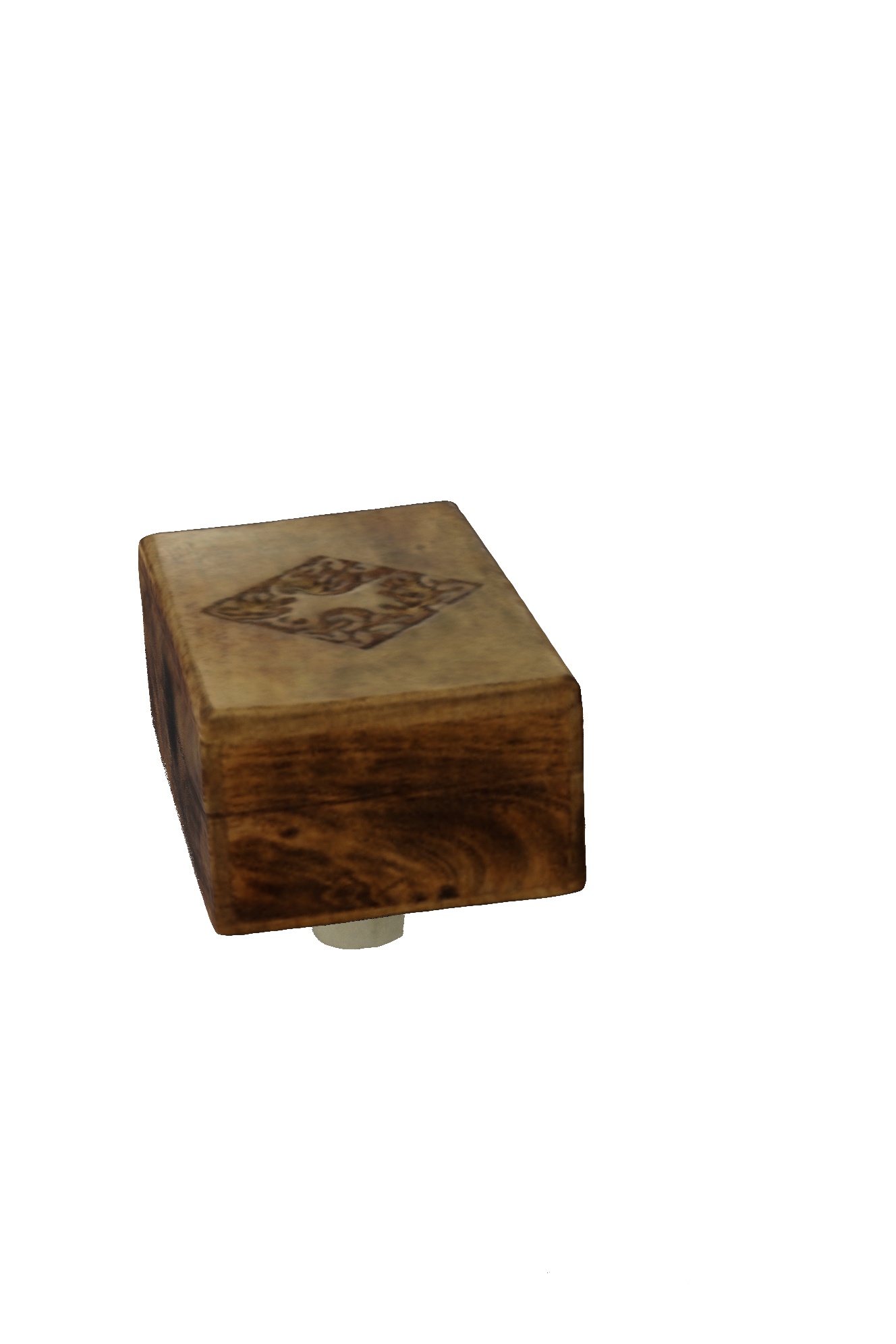} \\

\rotatebox{90}{\,\,\,\,\,\,\,\,\,\,\,\,\,\,\,\,\,\,\,\, Bucket} \hspace{1mm} & \adjincludegraphics[width=\resultwidth, trim={0.0in 0.0in 0.0in 0.0in}, clip]{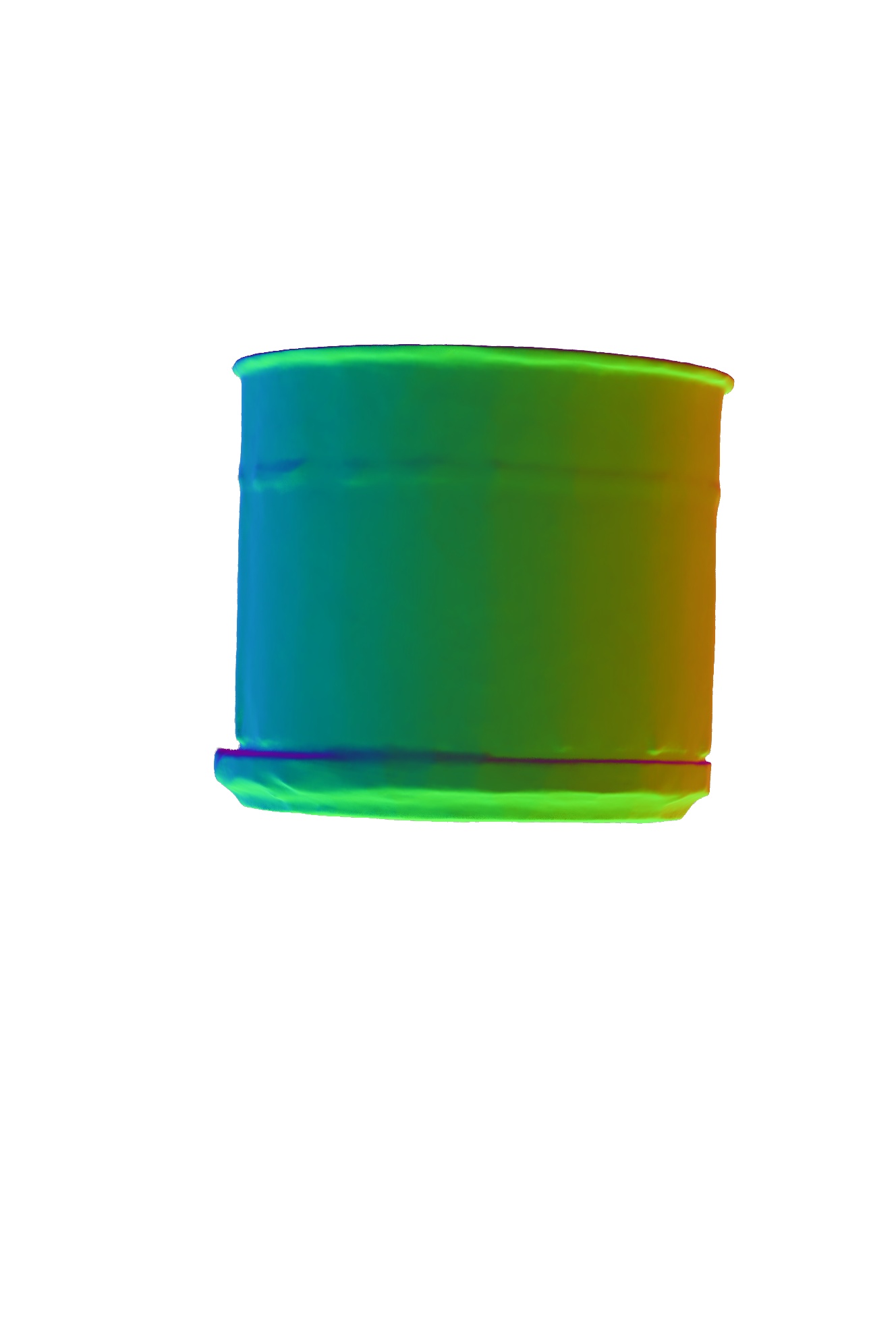} & \adjincludegraphics[width=\resultwidth, trim={0.0in 0.0in 0.0in 0.0in}, clip]{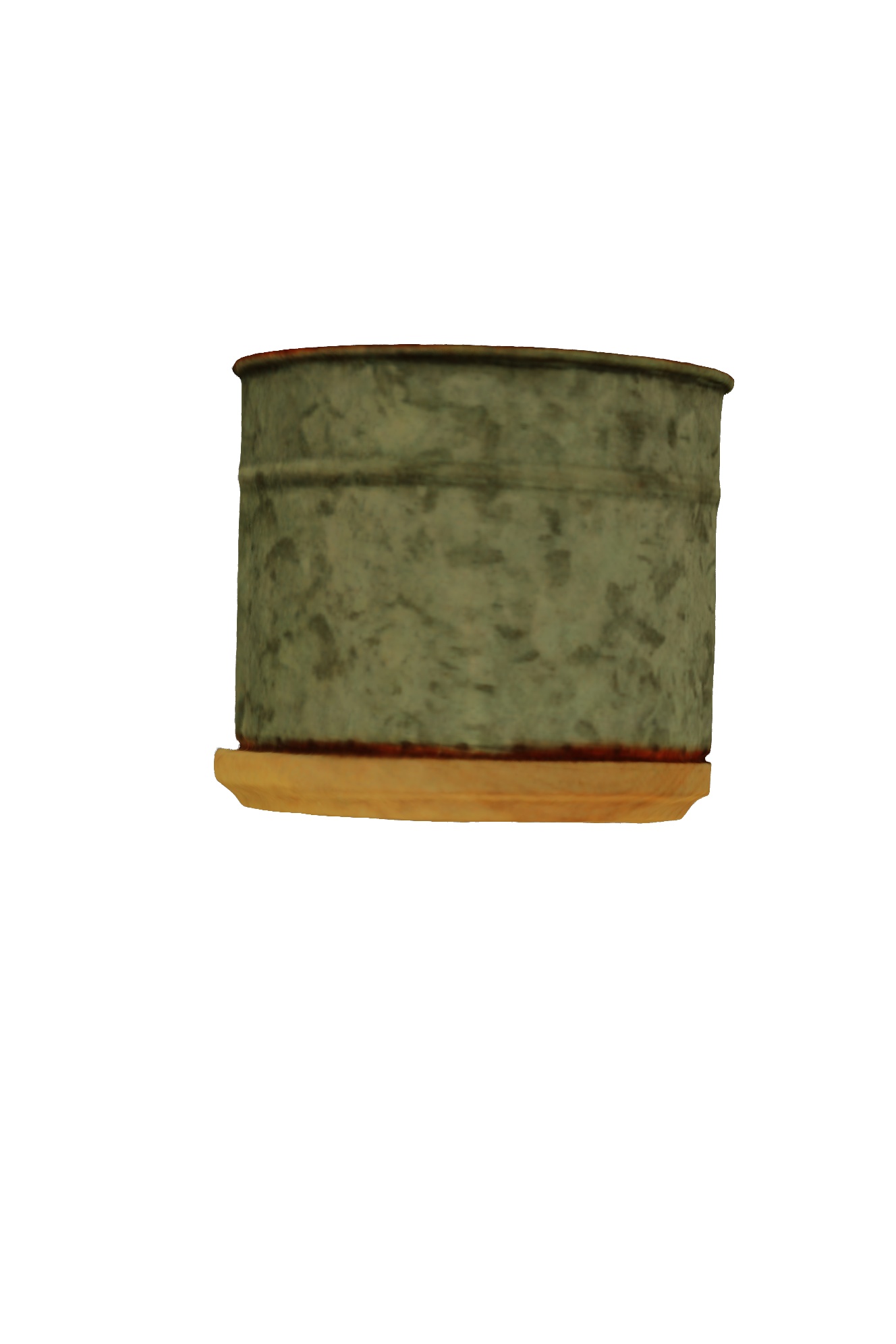} & \adjincludegraphics[width=\resultwidth, trim={0.0in 0.0in 0.0in 0.0in}, clip]{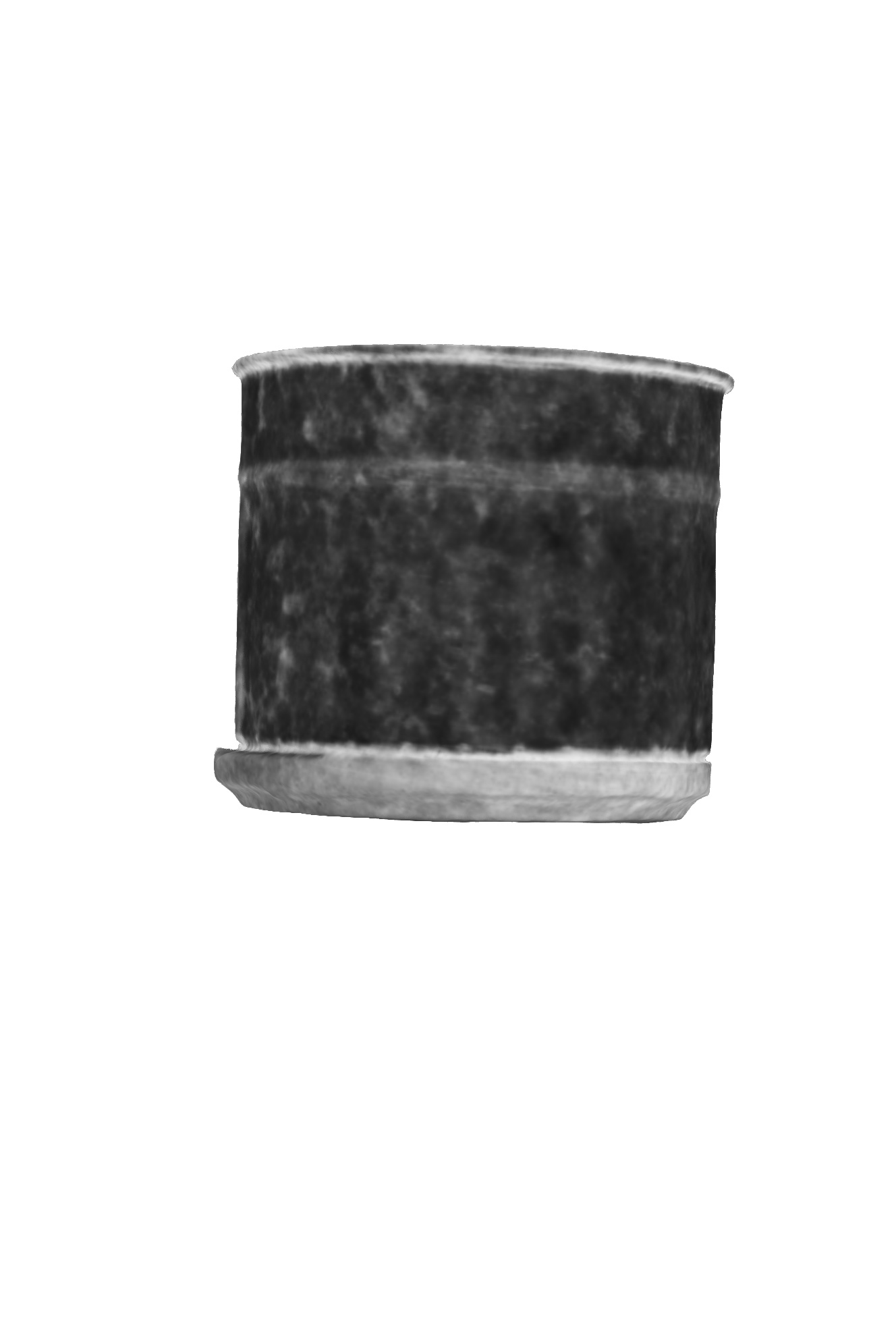} & \adjincludegraphics[width=\resultwidth, trim={0.0in 0.0in 0.0in 0.0in}, clip]{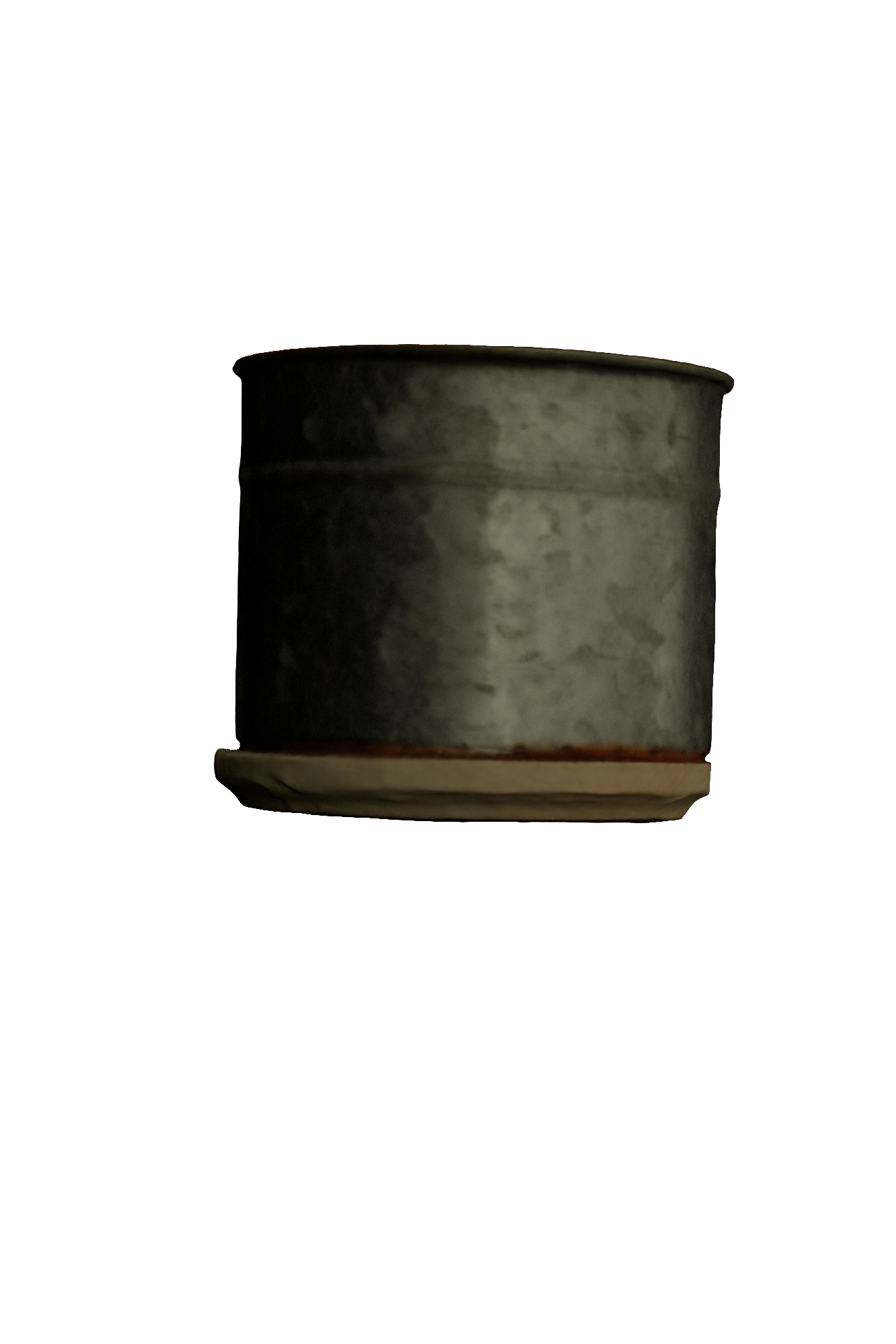} & \adjincludegraphics[width=\resultwidth, trim={0.0in 0.0in 0.0in 0.0in}, clip]{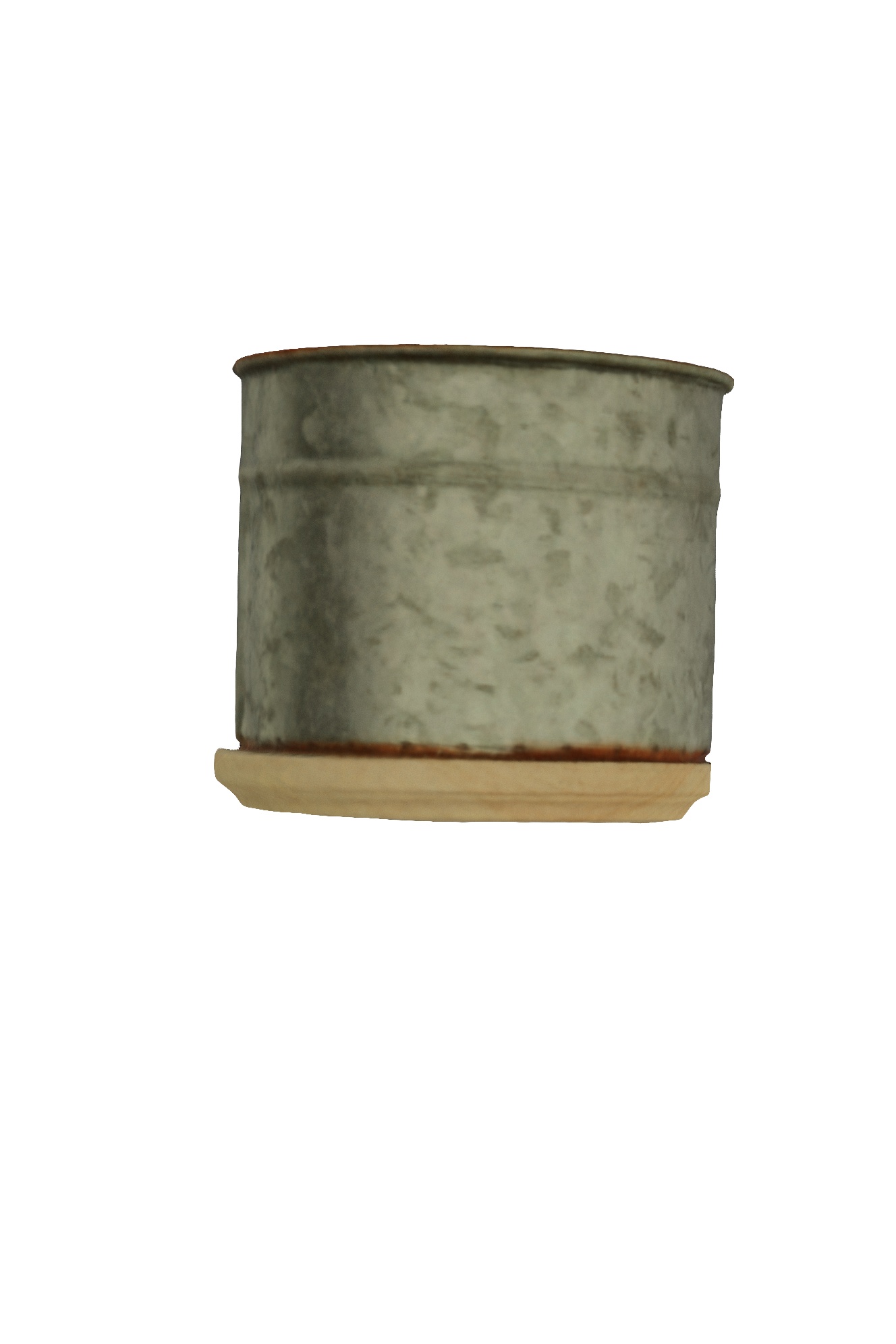} \\

\rotatebox{90}{\,\,\,\,\,\,\,\,\,\,\,\,\,\,\,\,\,\,\,\, Banana} \hspace{1mm} & \adjincludegraphics[width=\resultwidth, trim={0.0in 0.0in 0.0in 0.0in}, clip]{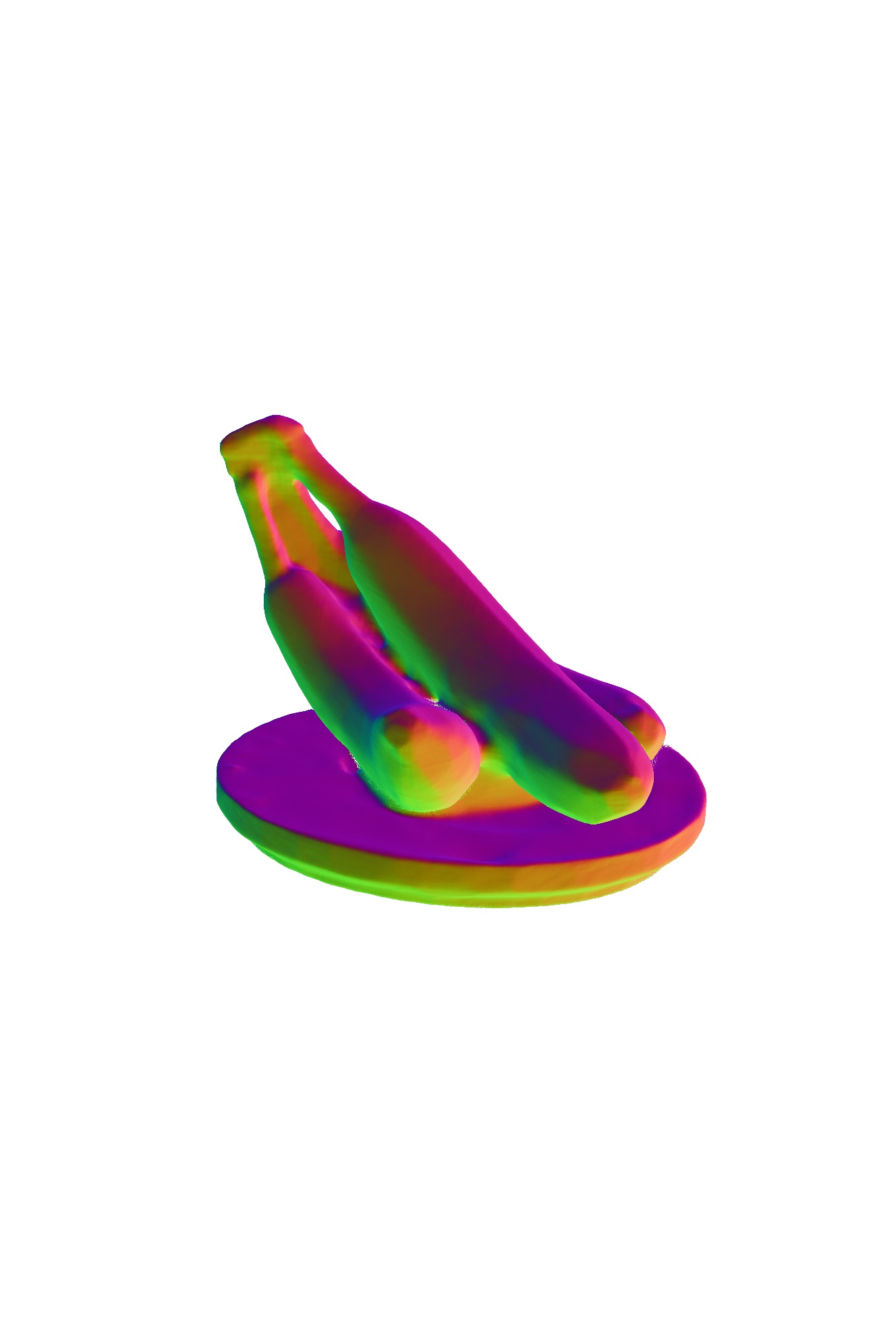} & \adjincludegraphics[width=\resultwidth, trim={0.0in 0.0in 0.0in 0.0in}, clip]{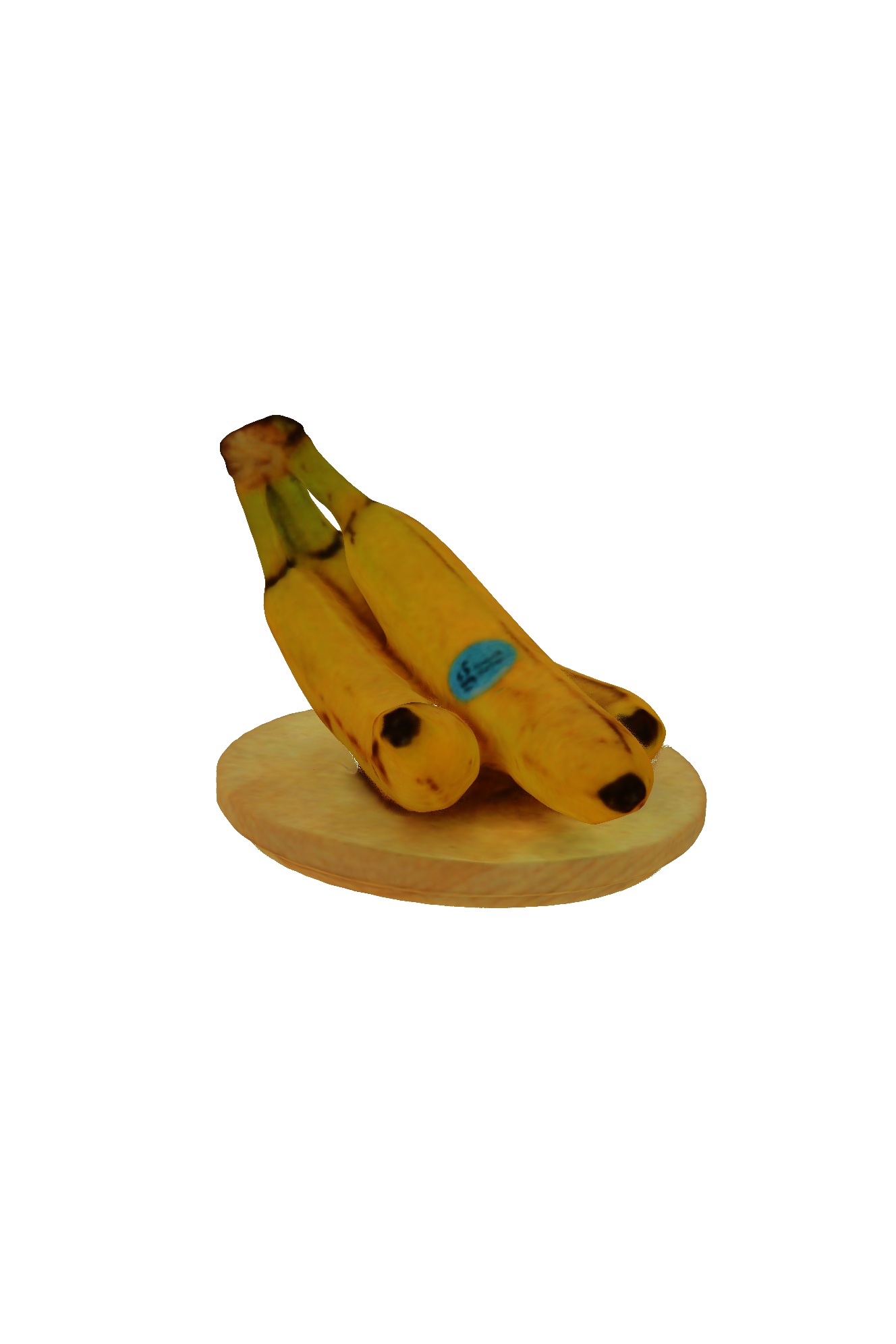} & \adjincludegraphics[width=\resultwidth, trim={0.0in 0.0in 0.0in 0.0in}, clip]{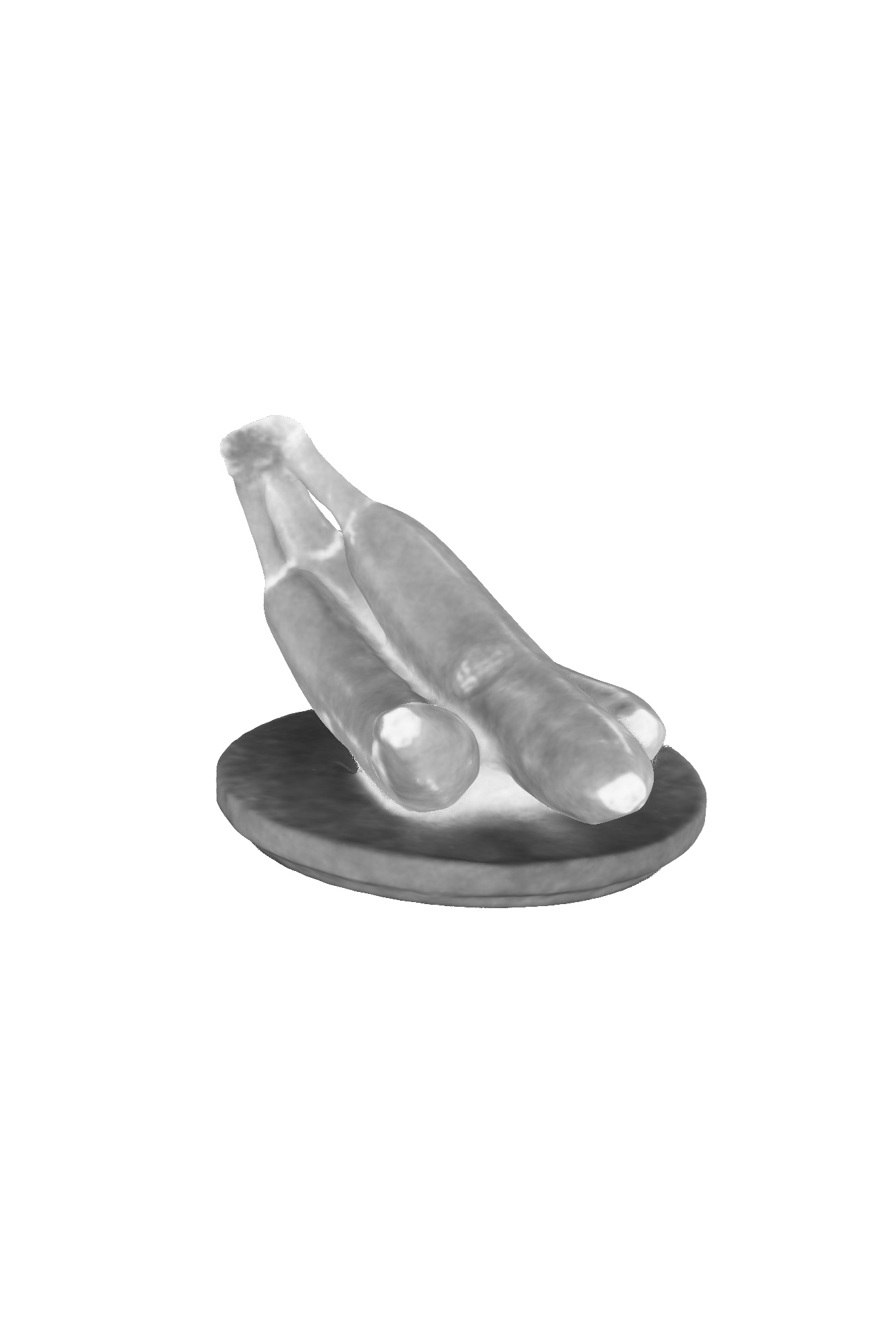} & \adjincludegraphics[width=\resultwidth, trim={0.0in 0.0in 0.0in 0.0in}, clip]{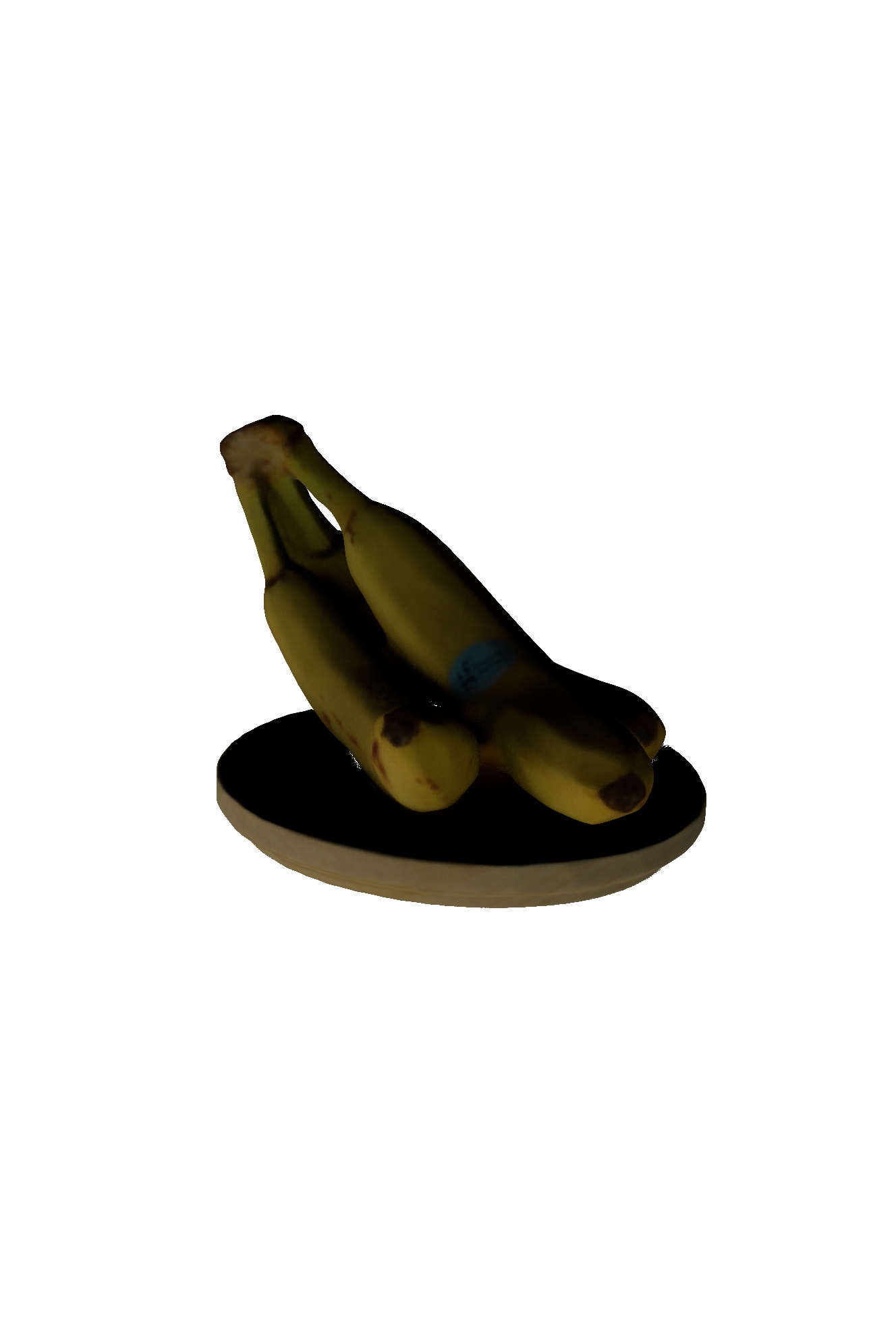} & \adjincludegraphics[width=\resultwidth, trim={0.0in 0.0in 0.0in 0.0in}, clip]{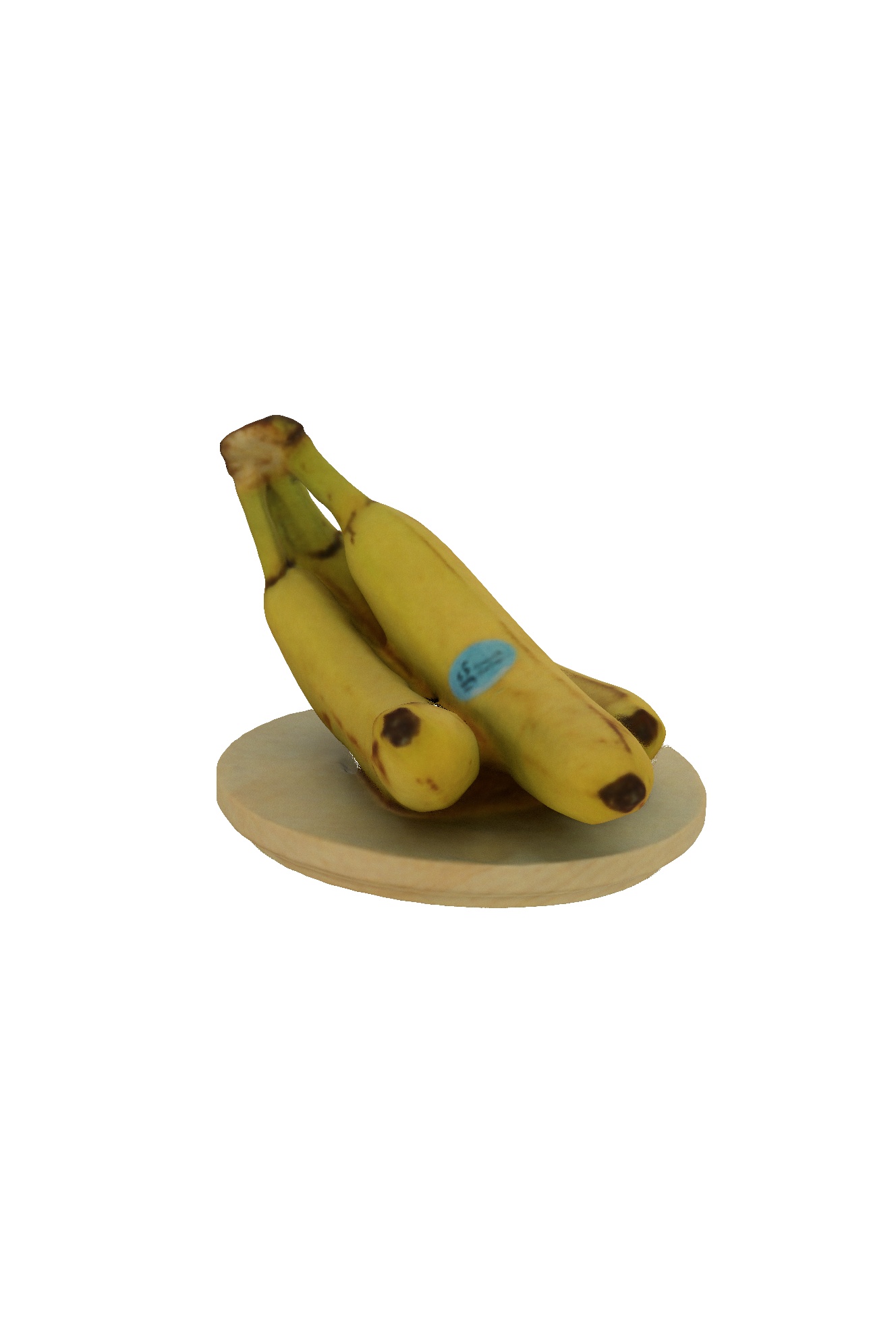} \\

\end{tabular}
\vspace{-0.1in}
\caption{%
\textbf{Additional Open Illumination~\cite{liu2024openillumination} Results.}
}%
\label{fig:qual-real}
\end{figure*}

\begin{figure*}[t!]
\footnotesize
\centering
\newcommand{\resultwidth}{0.92in}
\begin{tabular}{@{}c@{}c@{}c@{}c@{}c@{}c}
& GT & NDRMC & NeILF & NeRO & Ours \\

\rotatebox{90}{\,\,\,\,\,\,\,\,\,\,\,\,\,\,\,\,\, TBell} \hspace{1mm} & \adjincludegraphics[width=\resultwidth, trim={0.0in 0.0in 0.0in 0.0in}, clip]{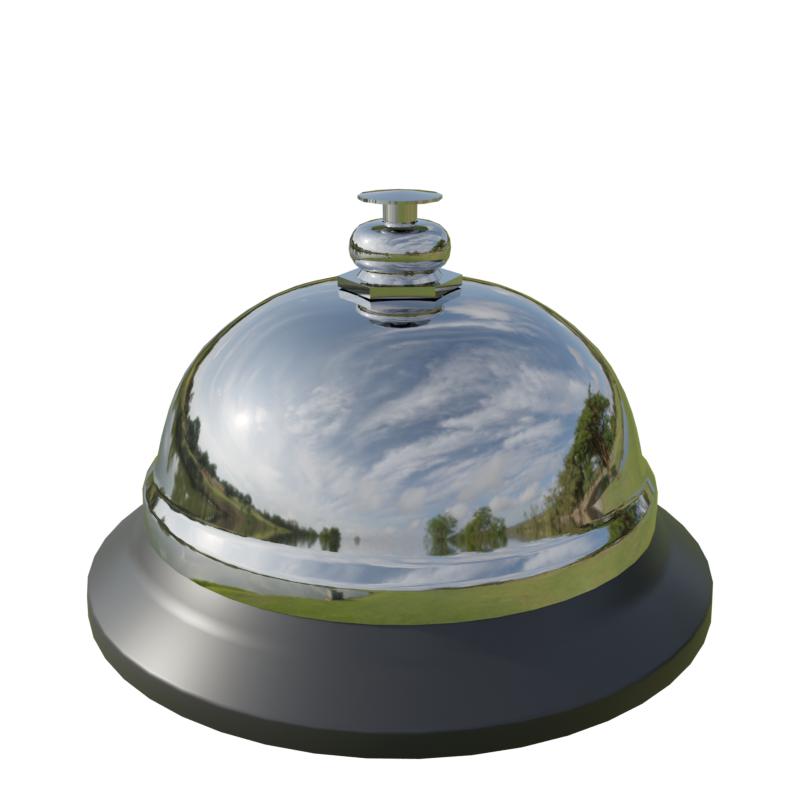} & \adjincludegraphics[width=\resultwidth, trim={0.0in 0.0in 0.0in 0.0in}, clip]{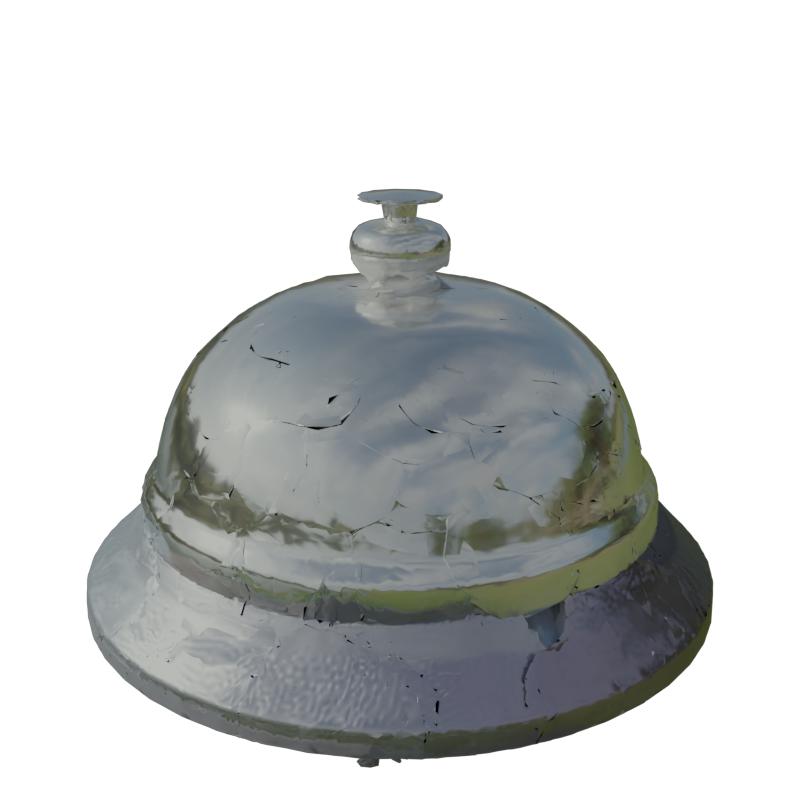} & \adjincludegraphics[width=\resultwidth, trim={0.0in 0.0in 0.0in 0.0in}, clip]{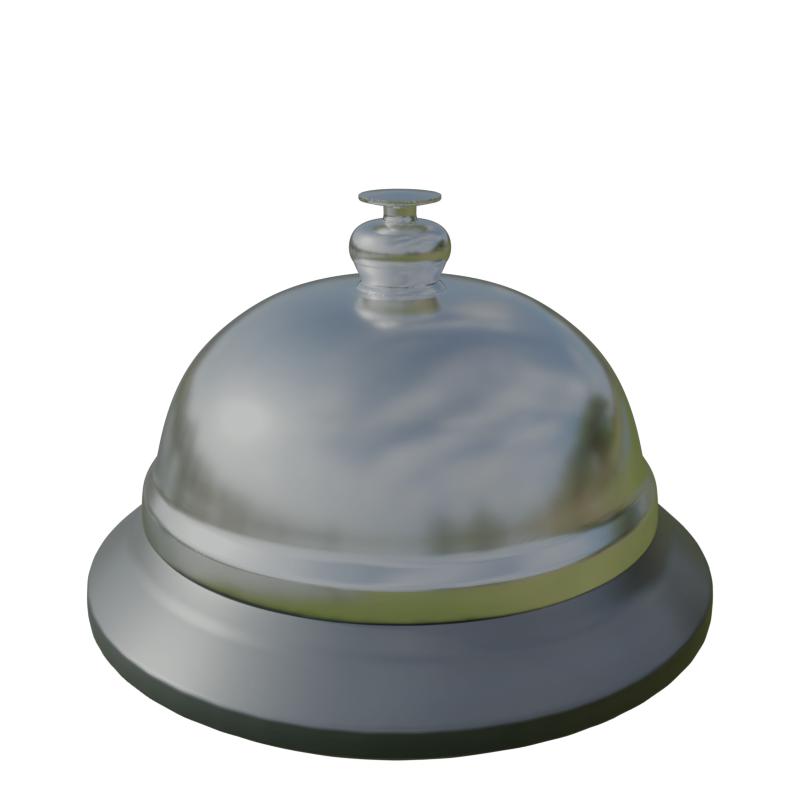} & \adjincludegraphics[width=\resultwidth, trim={0.0in 0.0in 0.0in 0.0in}, clip]{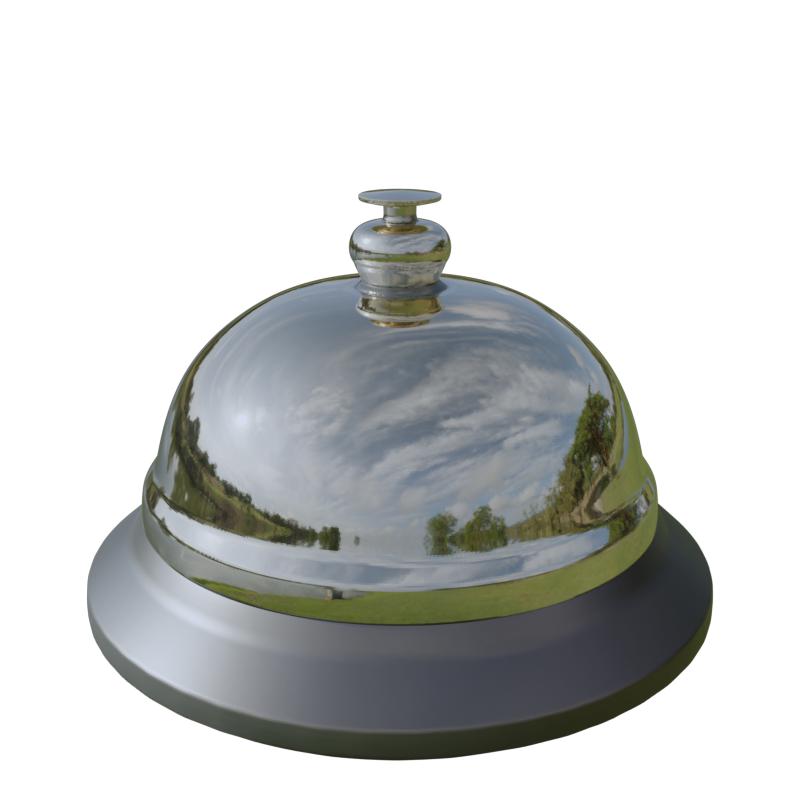} & \adjincludegraphics[width=\resultwidth, trim={0.0in 0.0in 0.0in 0.0in}, clip]{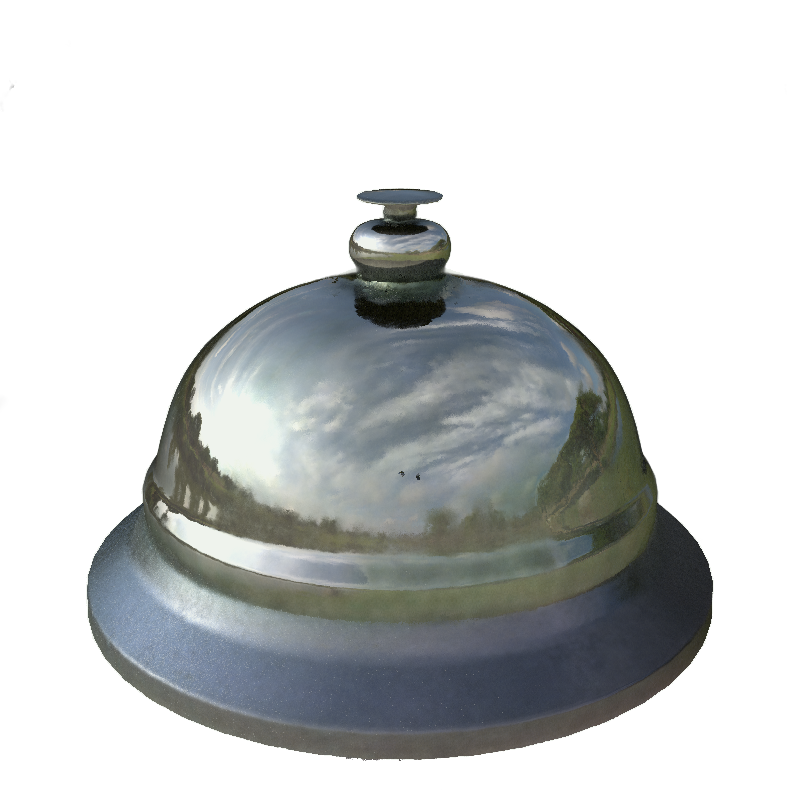} \\

\rotatebox{90}{\,\,\,\,\,\,\,\,\,\,\,\,\,\,\,\,\, Cat} \hspace{1mm} & \adjincludegraphics[width=\resultwidth, trim={0.0in 0.0in 0.0in 0.0in}, clip]{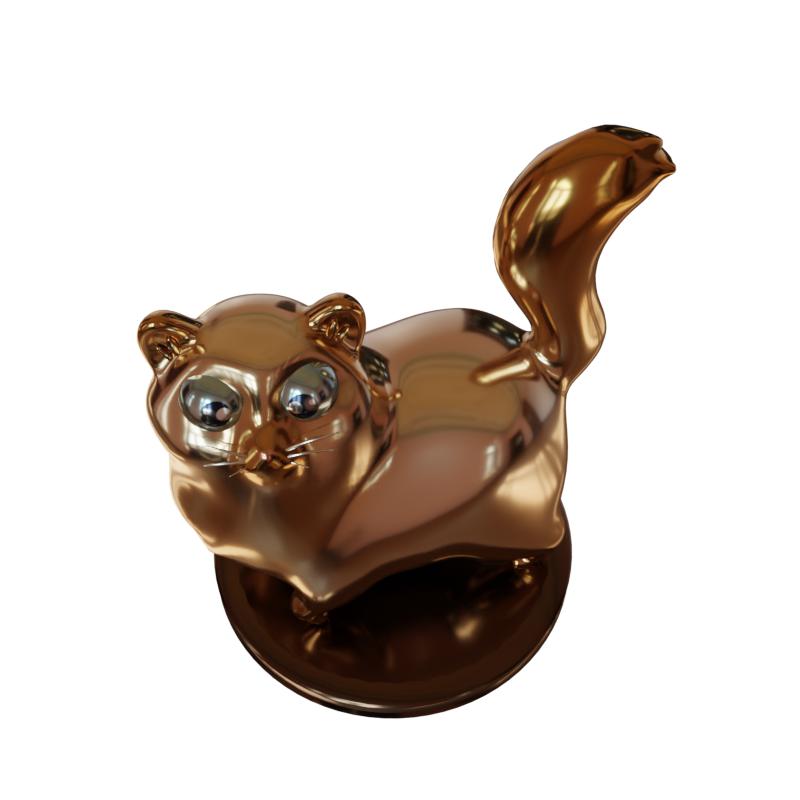} & \adjincludegraphics[width=\resultwidth, trim={0.0in 0.0in 0.0in 0.0in}, clip]{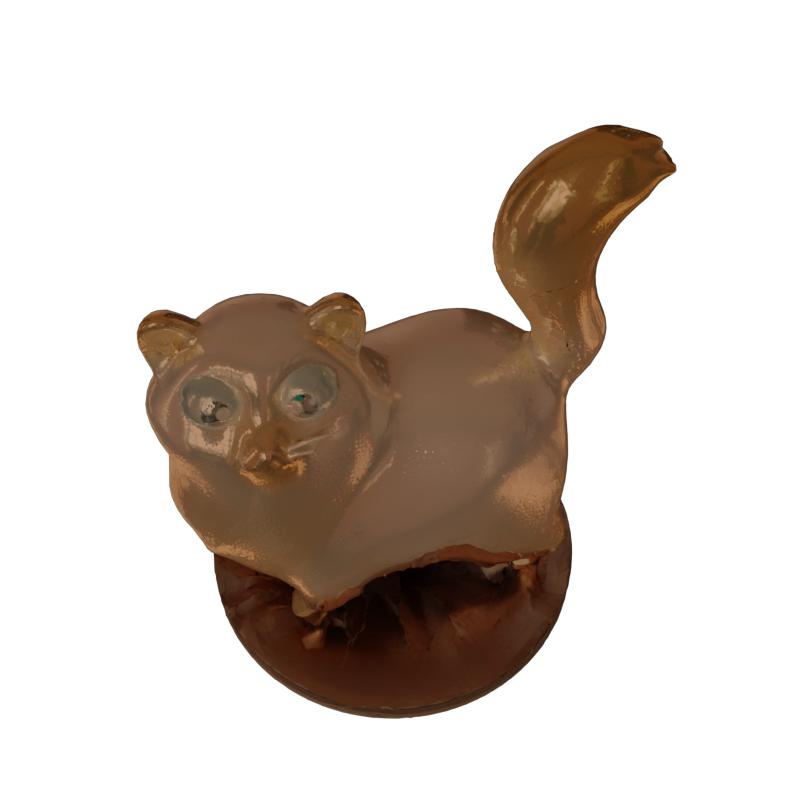} & \adjincludegraphics[width=\resultwidth, trim={0.0in 0.0in 0.0in 0.0in}, clip]{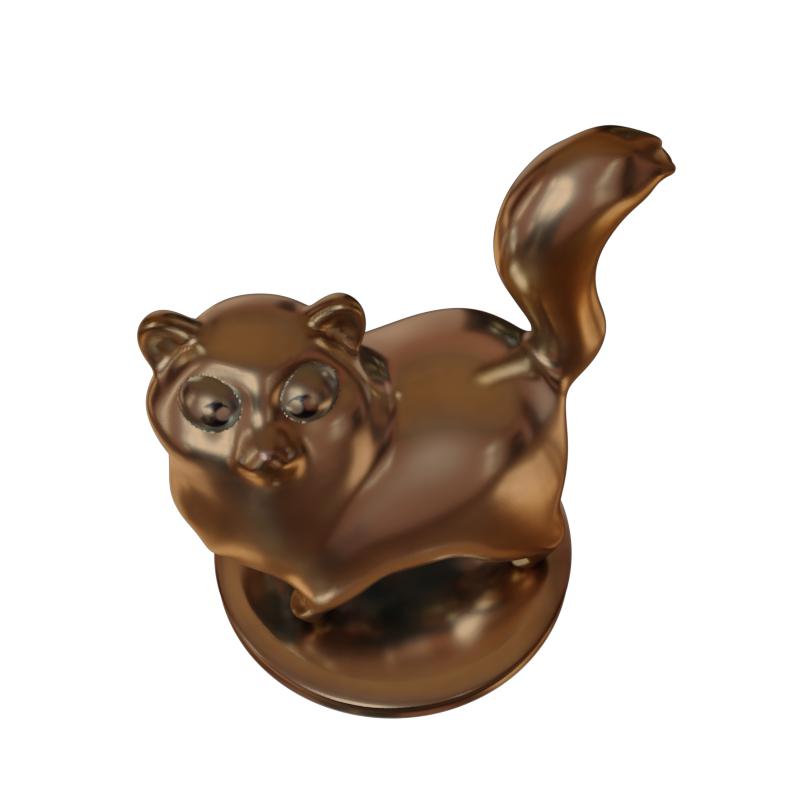} & \adjincludegraphics[width=\resultwidth, trim={0.0in 0.0in 0.0in 0.0in}, clip]{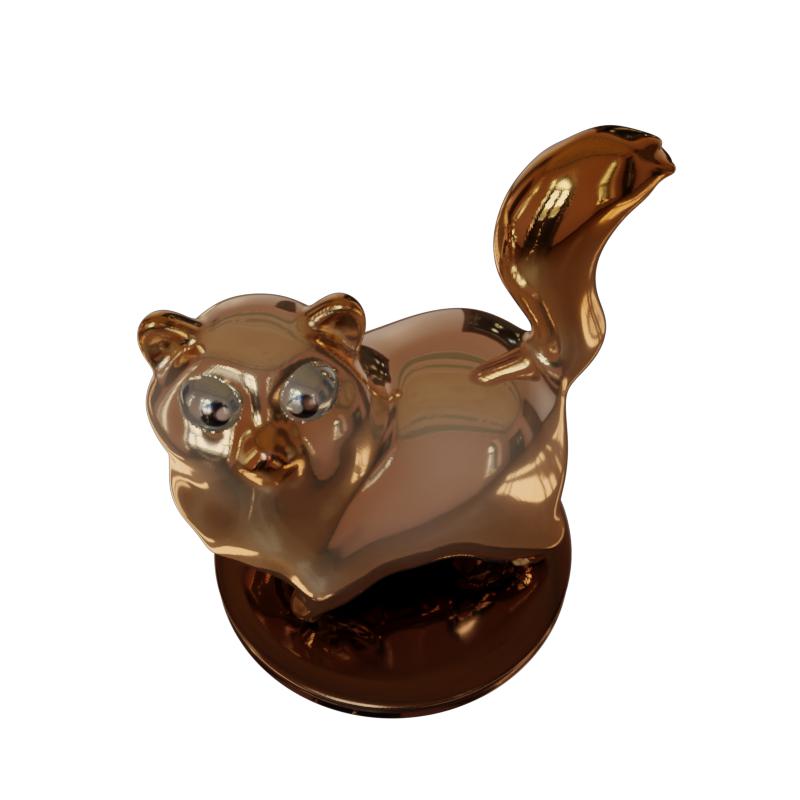} & \adjincludegraphics[width=\resultwidth, trim={0.0in 0.0in 0.0in 0.0in}, clip]{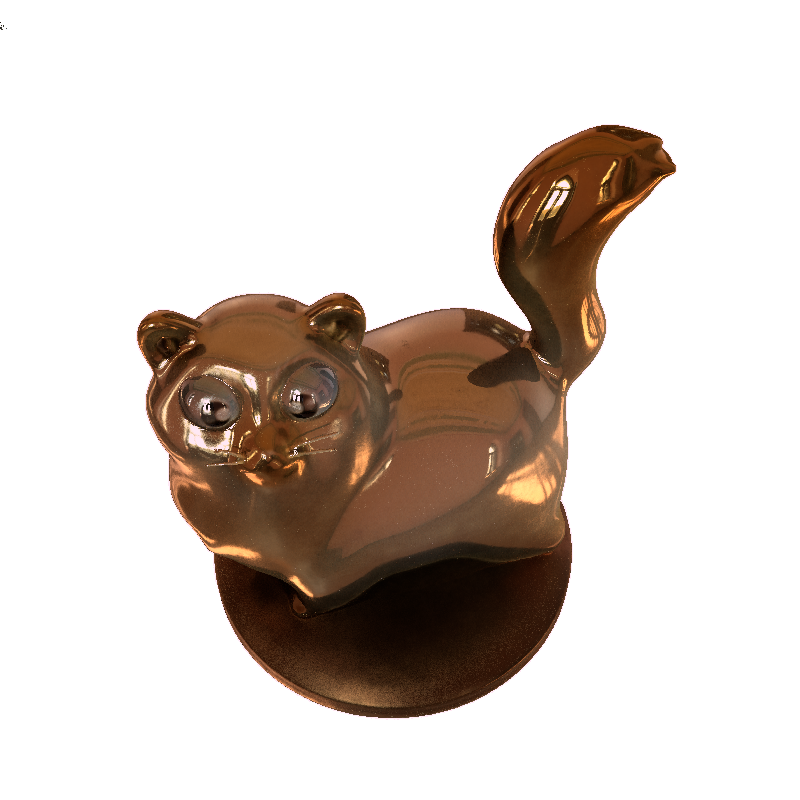} \\

\rotatebox{90}{\,\,\,\,\,\,\,\,\,\,\,\,\,\,\,\,\, Potion} \hspace{1mm} & \adjincludegraphics[width=\resultwidth, trim={0.0in 0.0in 0.0in 0.0in}, clip]{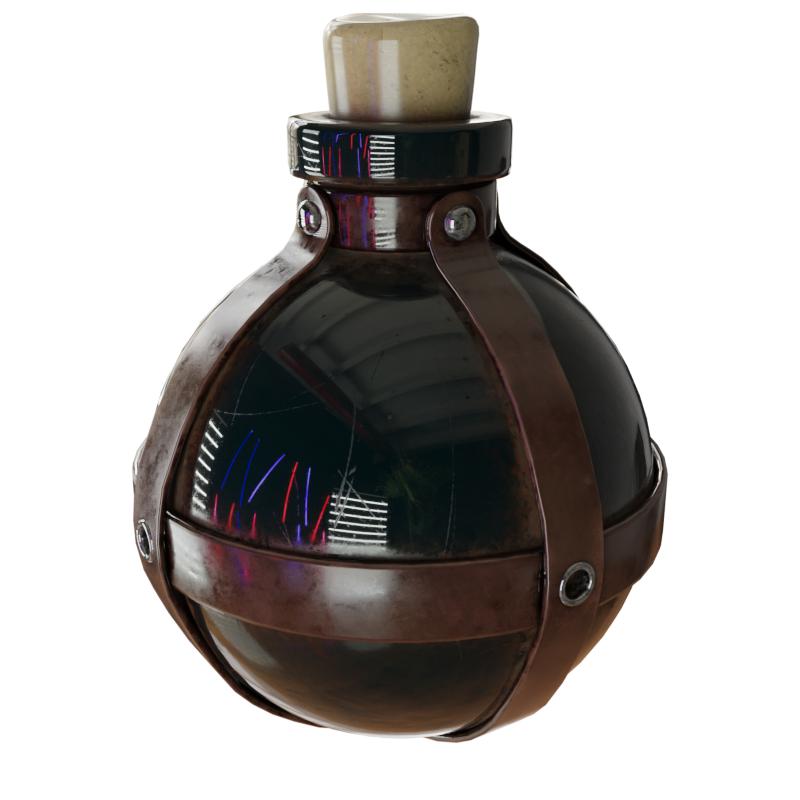} & \adjincludegraphics[width=\resultwidth, trim={0.0in 0.0in 0.0in 0.0in}, clip]{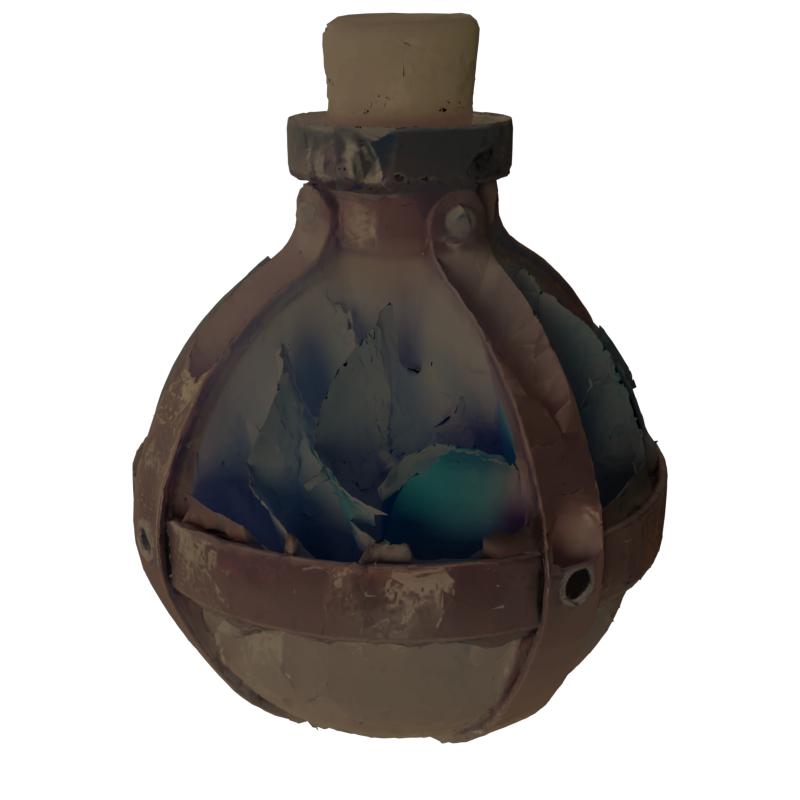} & \adjincludegraphics[width=\resultwidth, trim={0.0in 0.0in 0.0in 0.0in}, clip]{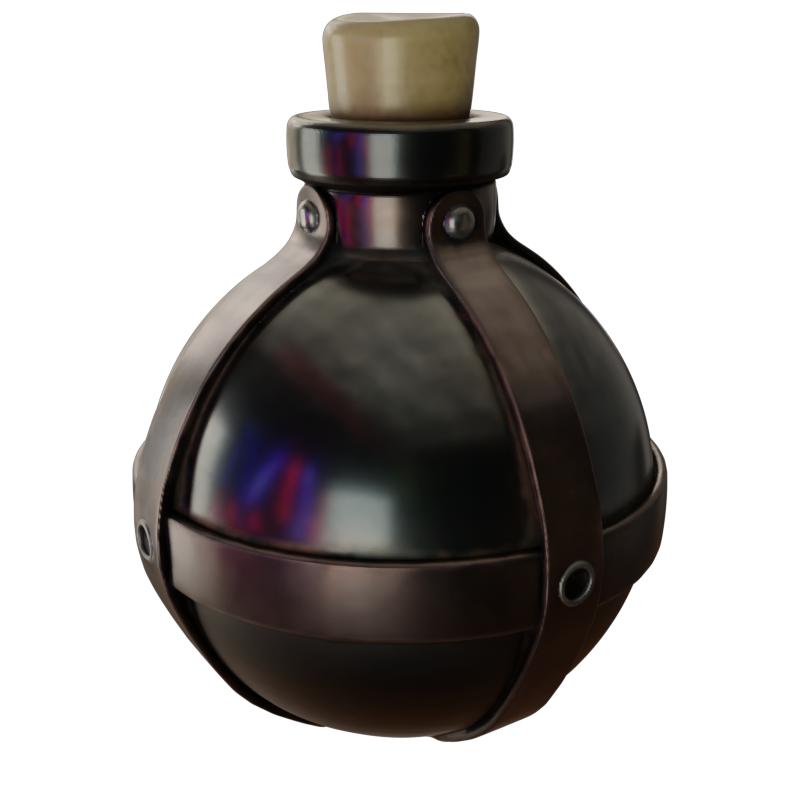} & \adjincludegraphics[width=\resultwidth, trim={0.0in 0.0in 0.0in 0.0in}, clip]{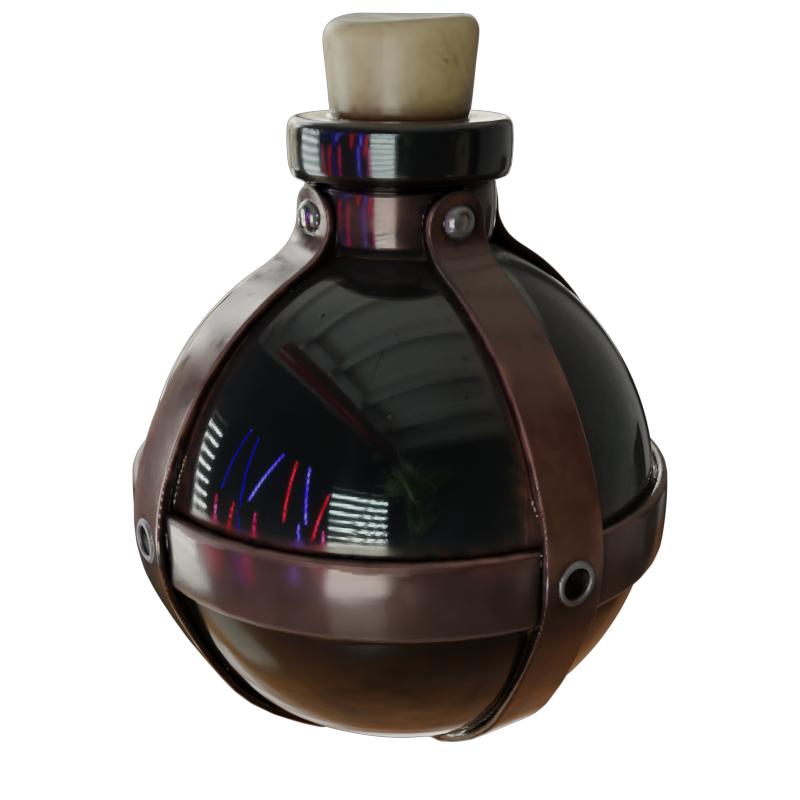} & \adjincludegraphics[width=\resultwidth, trim={0.0in 0.0in 0.0in 0.0in}, clip]{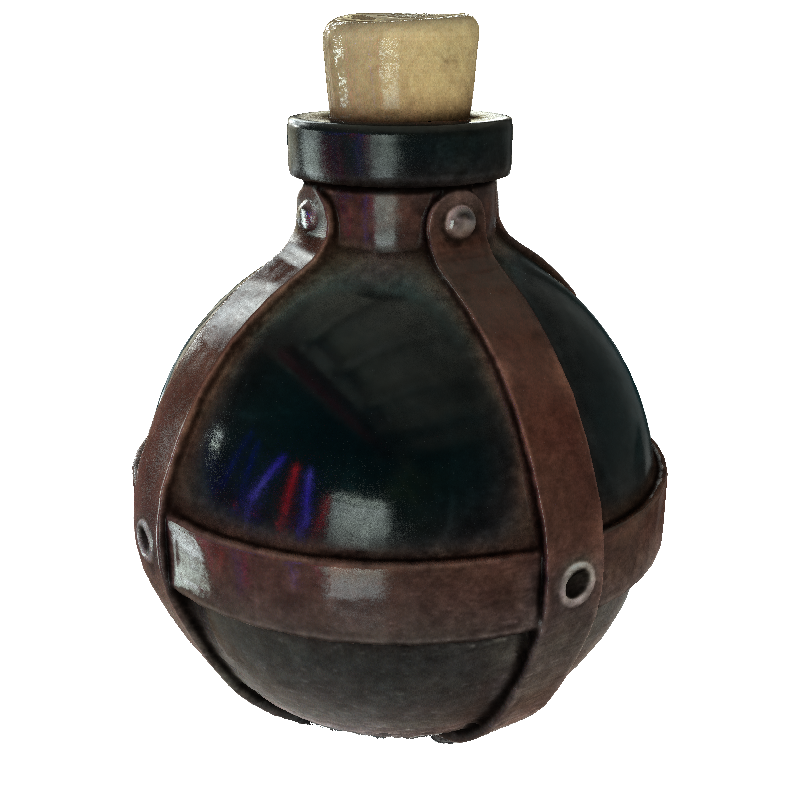} \\

\rotatebox{90}{\,\,\,\,\,\,\,\,\,\,\,\,\,\,\,\,\, Luyu} \hspace{1mm} & \adjincludegraphics[width=\resultwidth, trim={0.0in 0.0in 0.0in 0.0in}, clip]{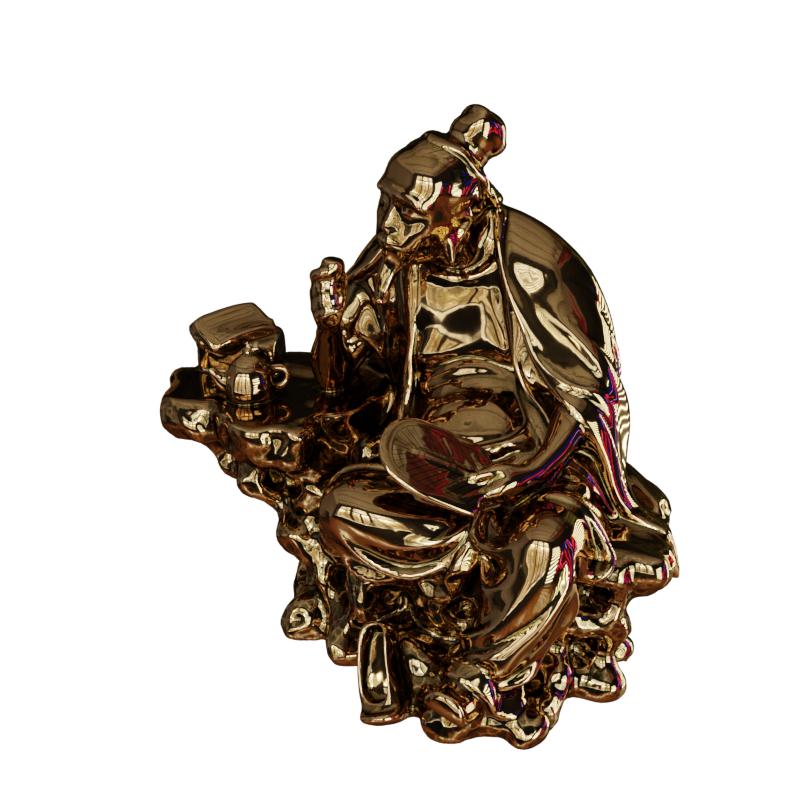} & \adjincludegraphics[width=\resultwidth, trim={0.0in 0.0in 0.0in 0.0in}, clip]{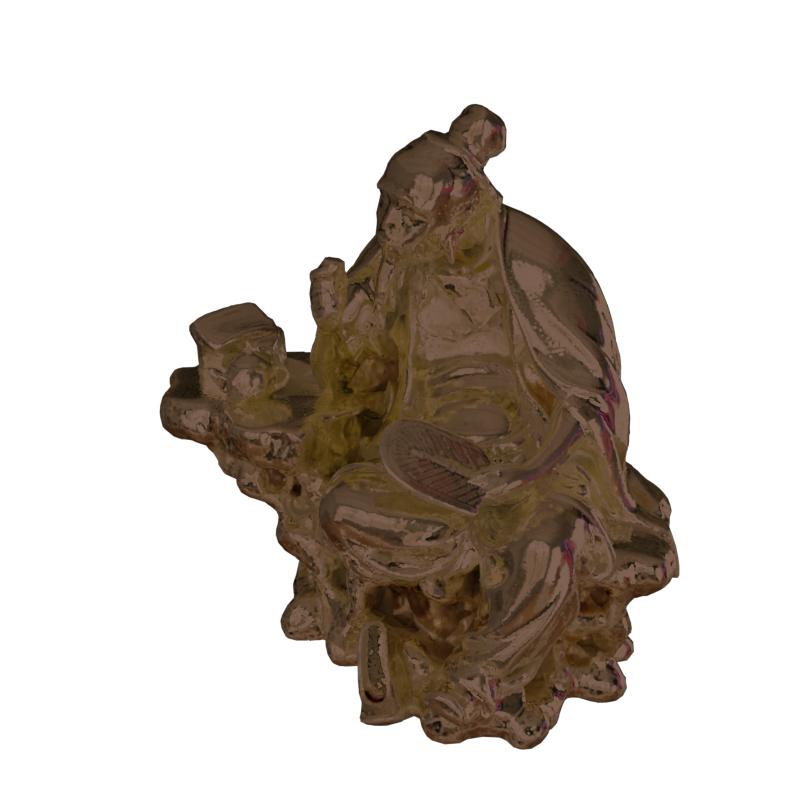} & \adjincludegraphics[width=\resultwidth, trim={0.0in 0.0in 0.0in 0.0in}, clip]{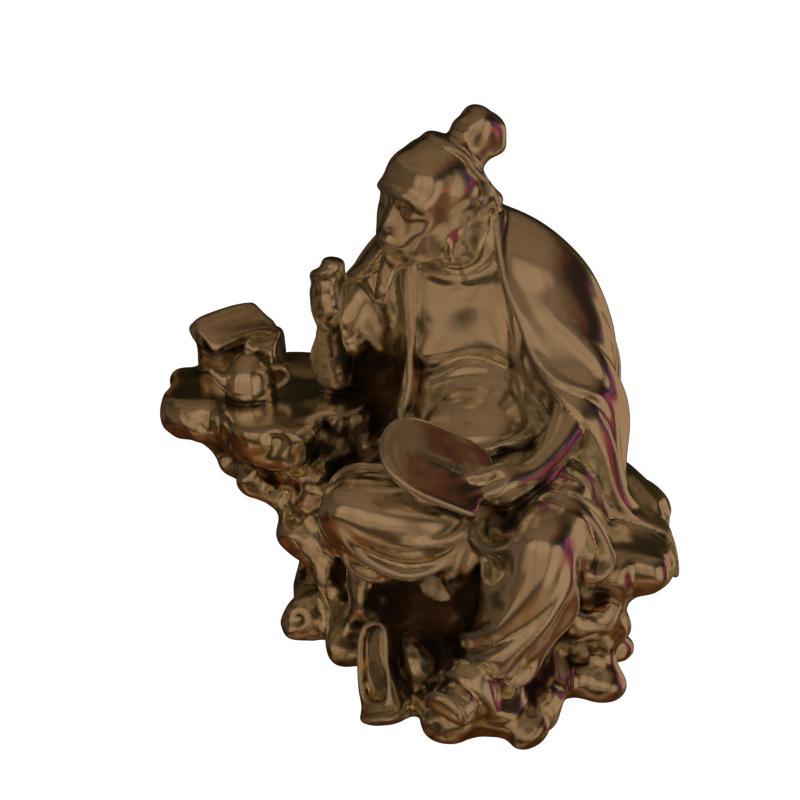} & \adjincludegraphics[width=\resultwidth, trim={0.0in 0.0in 0.0in 0.0in}, clip]{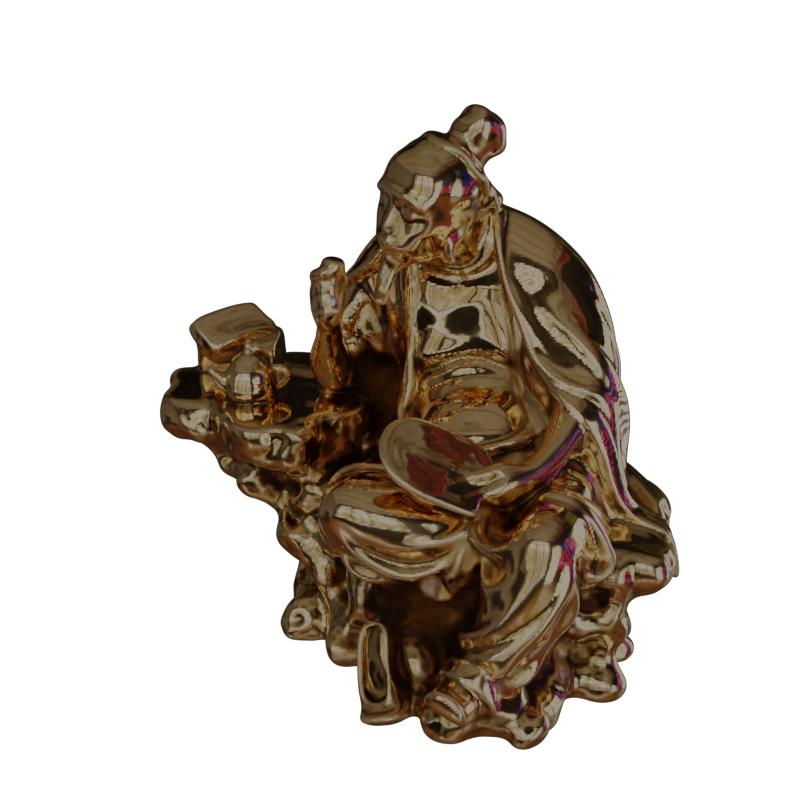} & \adjincludegraphics[width=\resultwidth, trim={0.0in 0.0in 0.0in 0.0in}, clip]{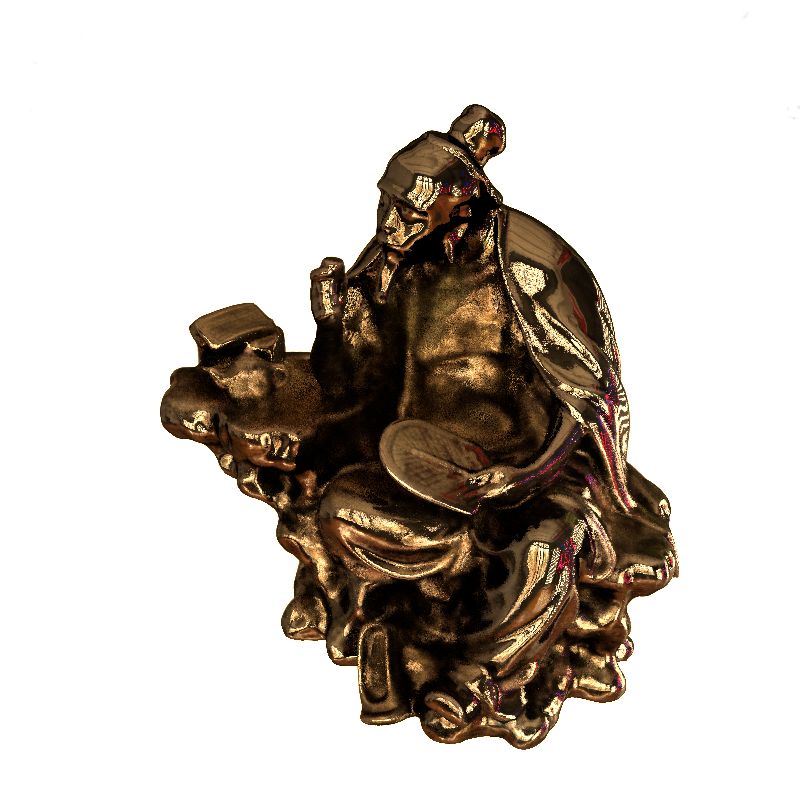} \\

\end{tabular}
\vspace{-0.1in}
\caption{%
\textbf{Glossy Synthetic~\cite{liu2023nero} Results.} Note that our results are relit with \textit{direct light only} (our codebase only supports direct relighting), while the other methods are relit using blender with exported meshes.
}%
\label{fig:qual-shiny}
\end{figure*}

Find additional qualitative results in Figures~\ref{fig:qual-tensoir},~\ref{fig:qual-real}, an ~\ref{fig:qual-shiny}as well as our videos on our  \href{https://benattal.github.io/flash-cache/}{supplemental website}.

\begin{figure*}[t!]
\footnotesize

\centering
\newcommand{\resultwidth}{0.95in}
\begin{tabular}{@{}c@{}c@{}c@{}c@{}c@{}c}
& Normal & Albedo & Roughness & Relighting 1 & Relighting 2 \\

\rotatebox{90}{\,\,\, Single Light} & \adjincludegraphics[width=\resultwidth, trim={0.0in 0.0in 0.0in 0.0in}, clip]{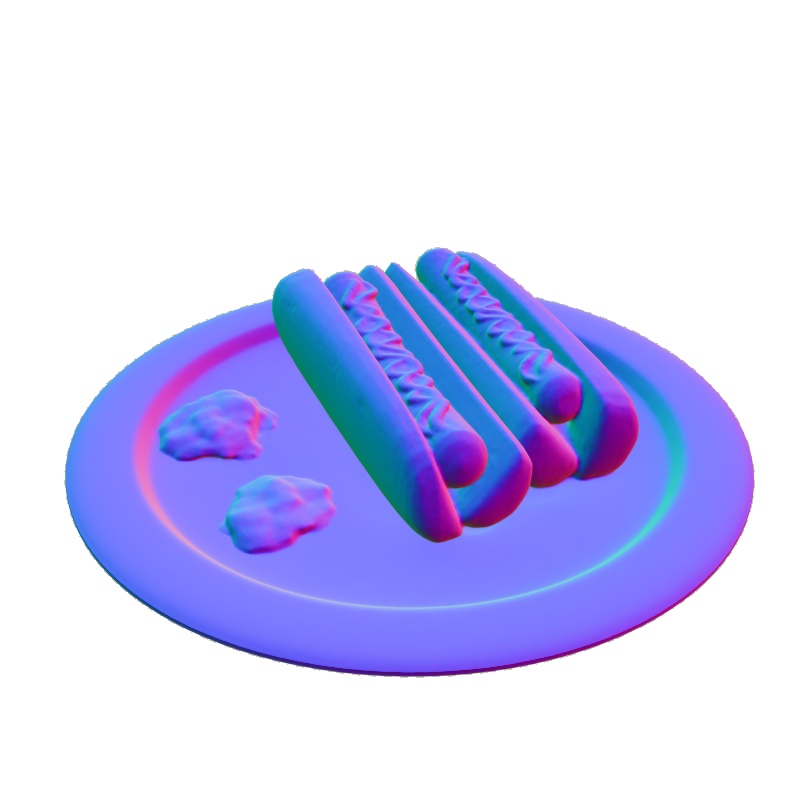} & \adjincludegraphics[width=\resultwidth, trim={0.0in 0.0in 0.0in 0.0in}, clip]{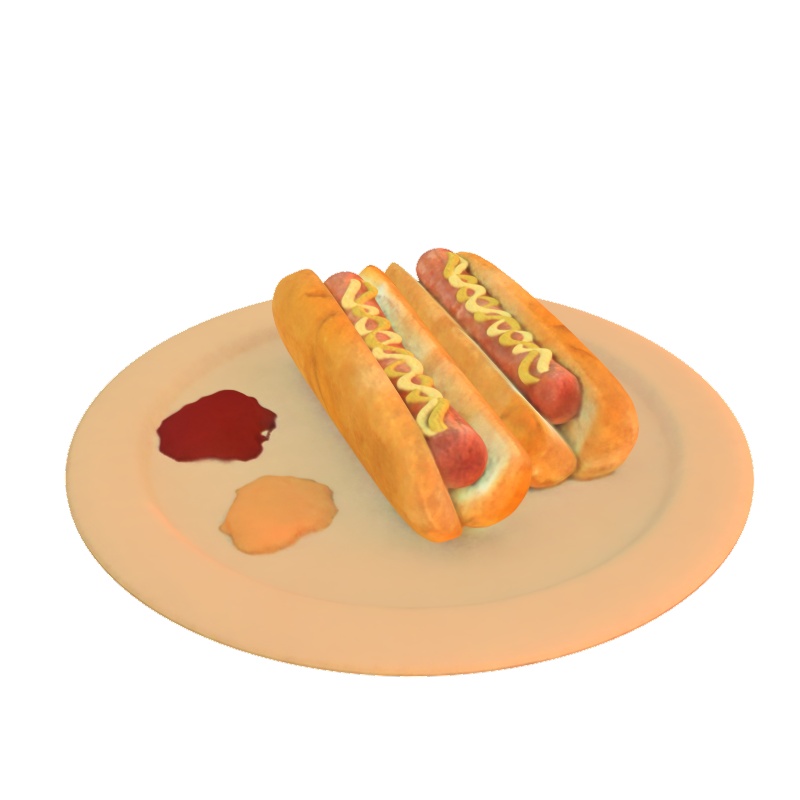} & \adjincludegraphics[width=\resultwidth, trim={0.0in 0.0in 0.0in 0.0in}, clip]{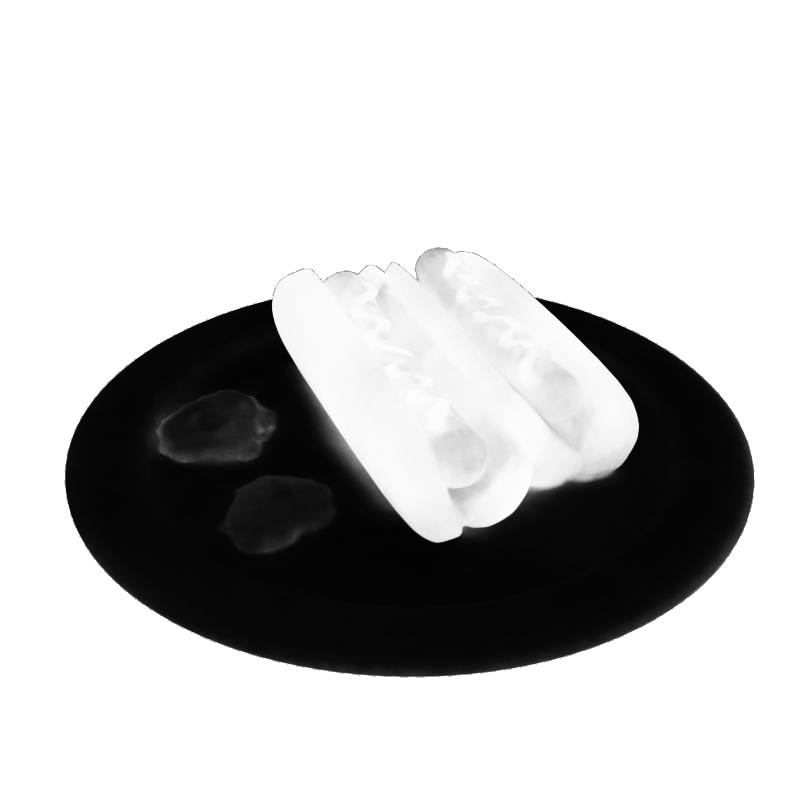} & \adjincludegraphics[width=\resultwidth, trim={0.0in 0.0in 0.0in 0.0in}, clip]{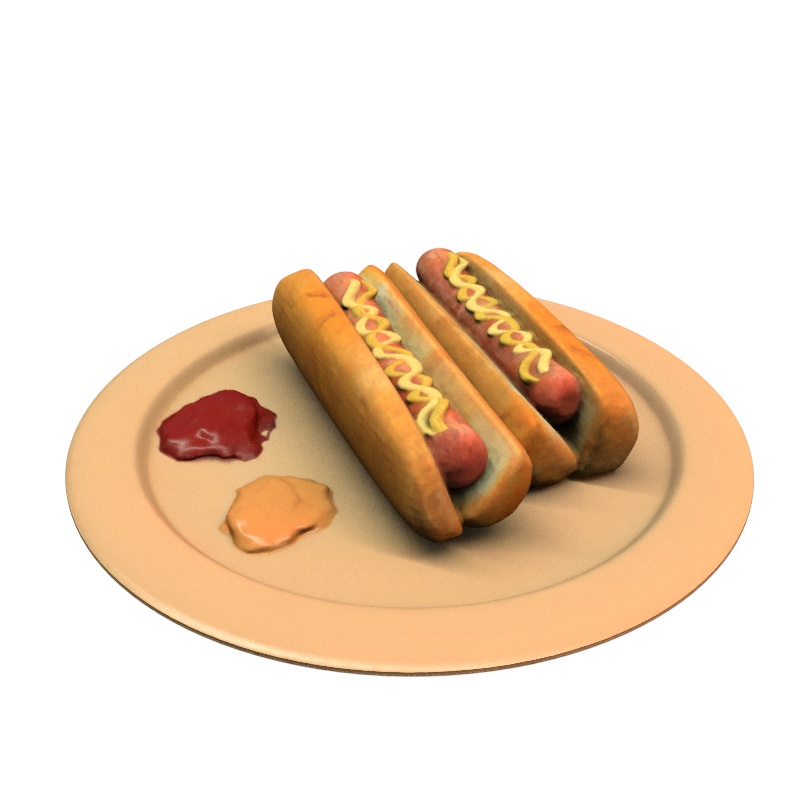} & \adjincludegraphics[width=\resultwidth, trim={0.0in 0.0in 0.0in 0.0in}, clip]{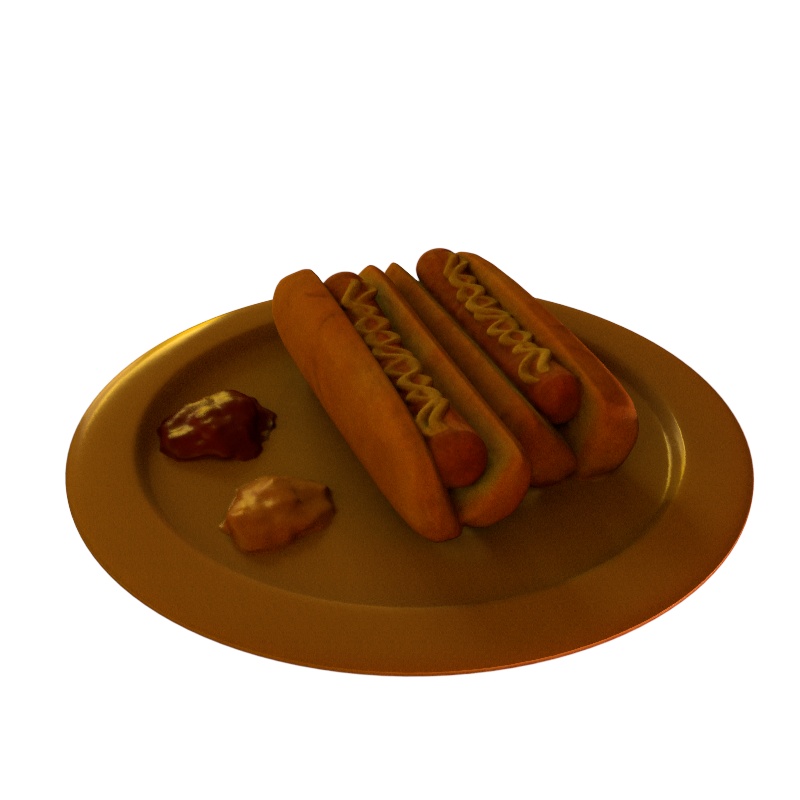} \\

\rotatebox{90}{\,\,\, Multi Light} & \adjincludegraphics[width=\resultwidth, trim={0.0in 0.0in 0.0in 0.0in}, clip]{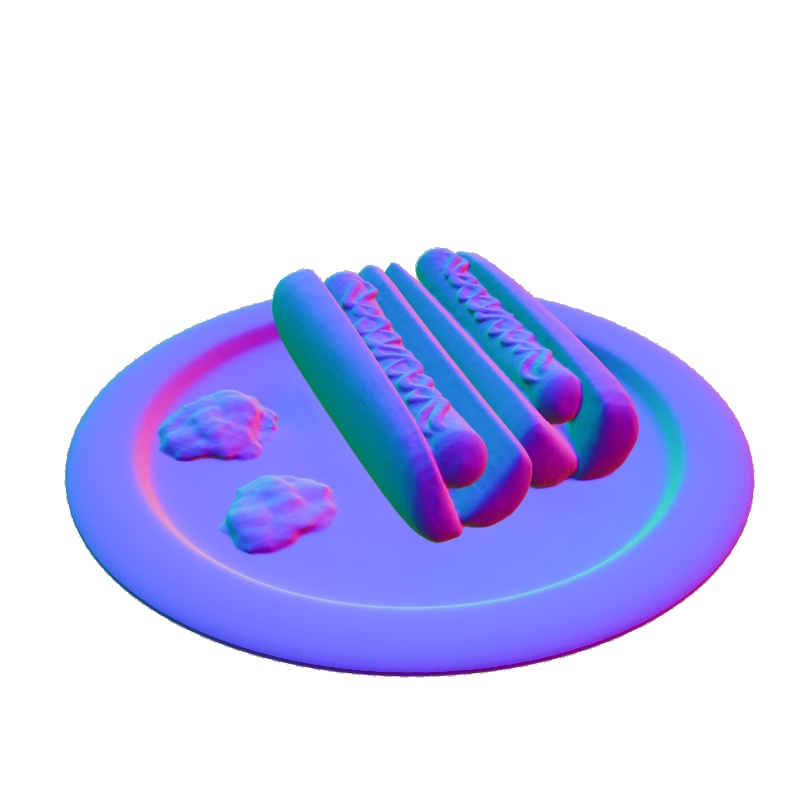} & \adjincludegraphics[width=\resultwidth, trim={0.0in 0.0in 0.0in 0.0in}, clip]{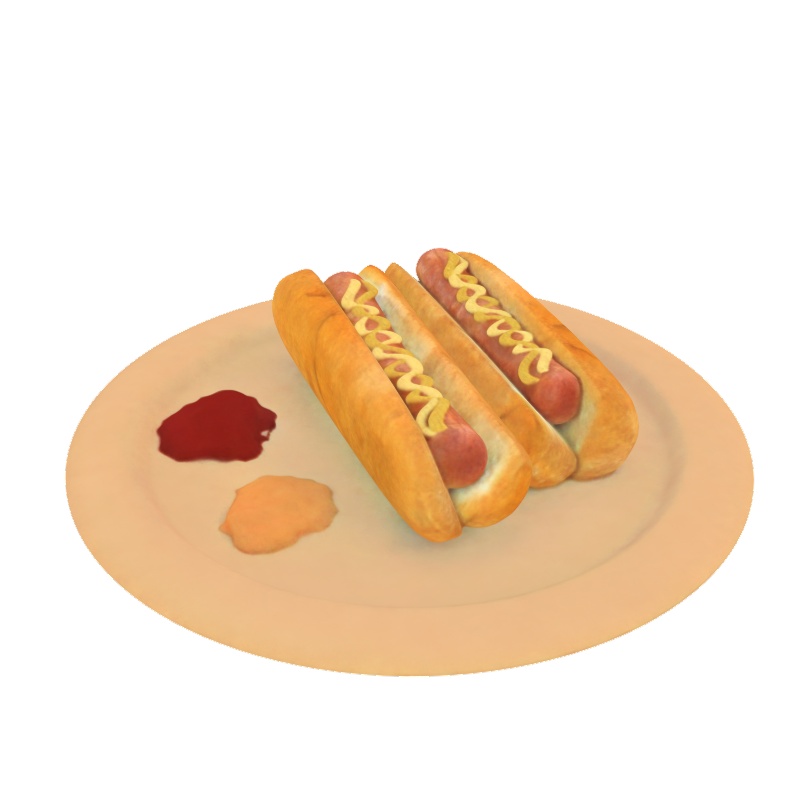} & \adjincludegraphics[width=\resultwidth, trim={0.0in 0.0in 0.0in 0.0in}, clip]{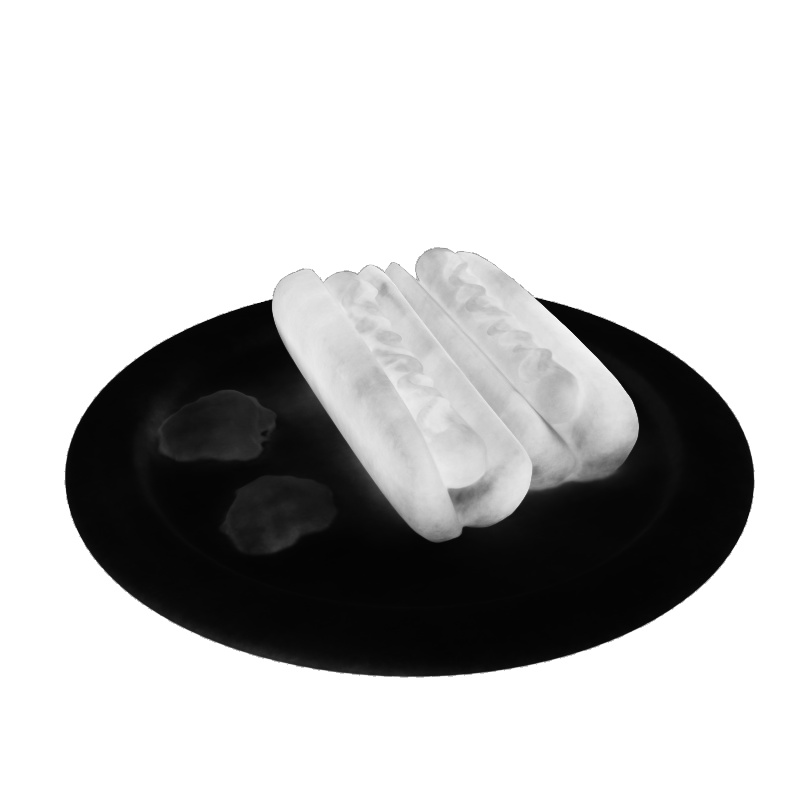} & \adjincludegraphics[width=\resultwidth, trim={0.0in 0.0in 0.0in 0.0in}, clip]{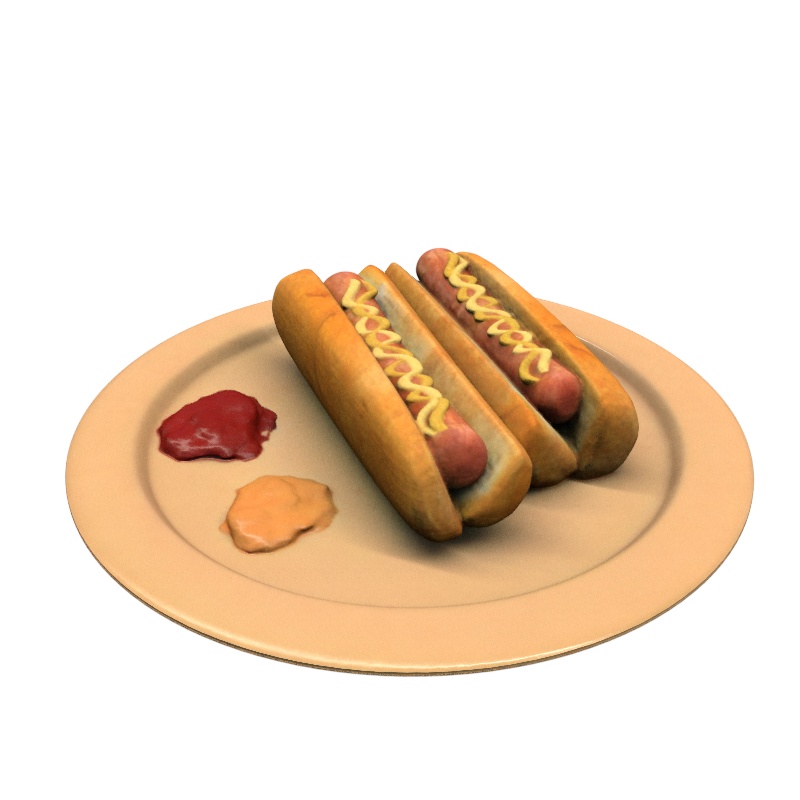} & \adjincludegraphics[width=\resultwidth, trim={0.0in 0.0in 0.0in 0.0in}, clip]{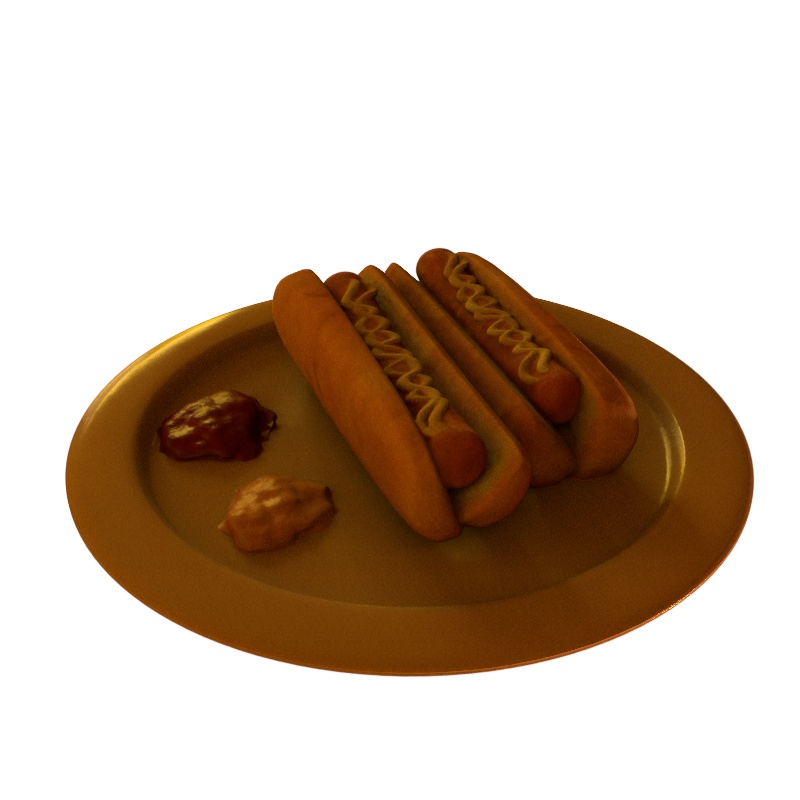} \\

\rotatebox{90}{\,\,\,\,\,\,\,\,\,\, GT} & \adjincludegraphics[width=\resultwidth, trim={0.0in 0.0in 0.0in 0.0in}, clip]{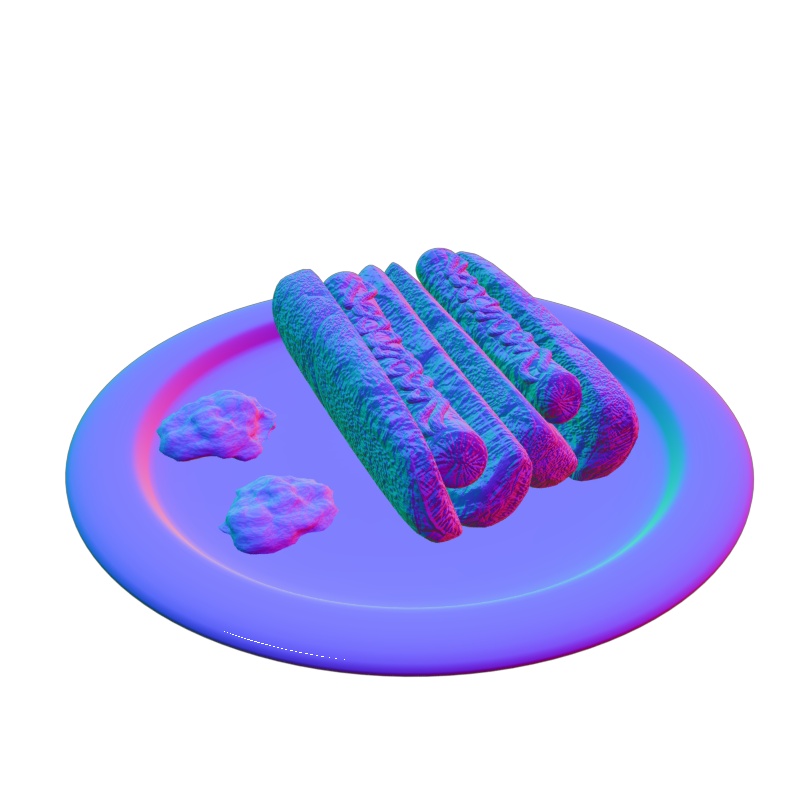} & \adjincludegraphics[width=\resultwidth, trim={0.0in 0.0in 0.0in 0.0in}, clip]{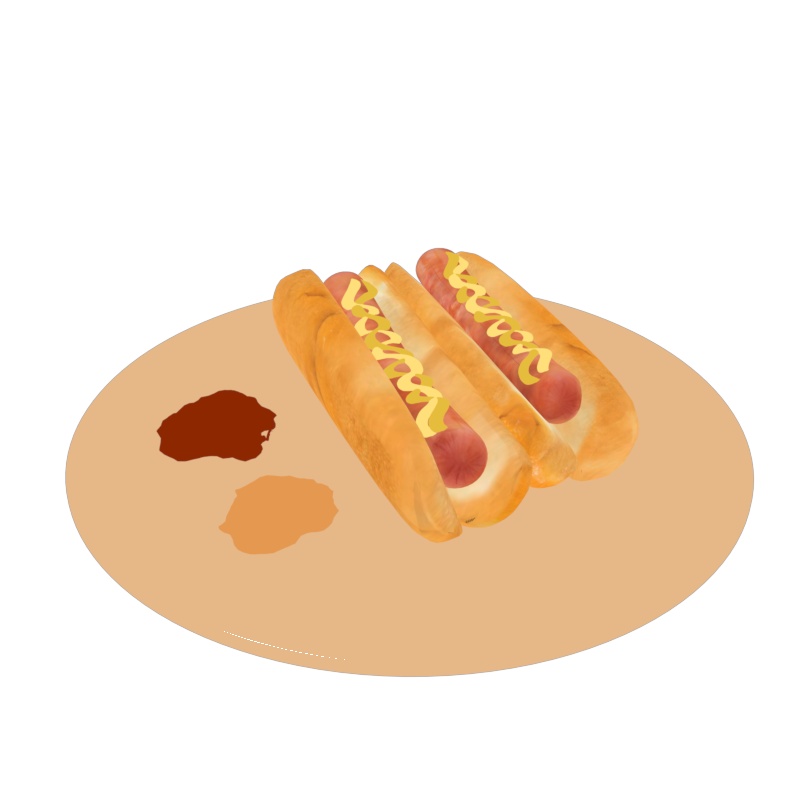} & & \adjincludegraphics[width=\resultwidth, trim={0.0in 0.0in 0.0in 0.0in}, clip]{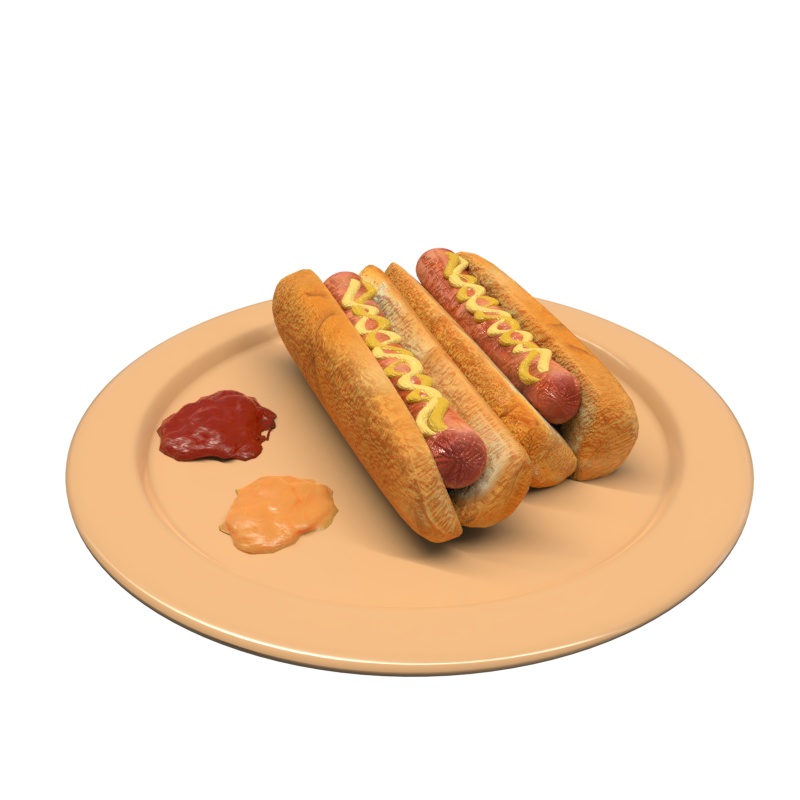} & \adjincludegraphics[width=\resultwidth, trim={0.0in 0.0in 0.0in 0.0in}, clip]{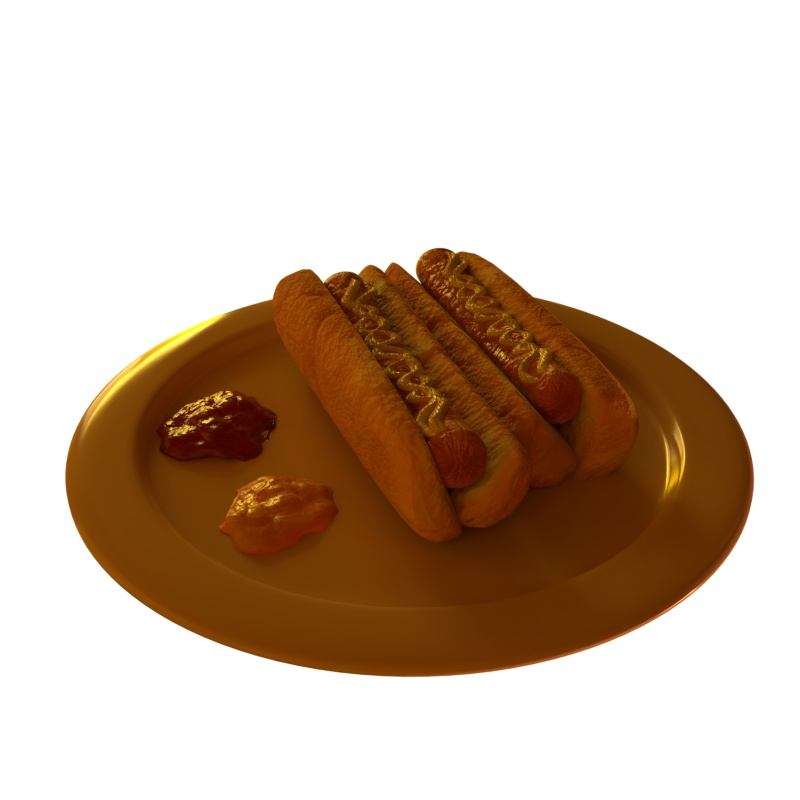} \\

\rotatebox{90}{\,\,\, Single Light} & \adjincludegraphics[width=\resultwidth, trim={0.0in 0.0in 0.0in 0.0in}, clip]{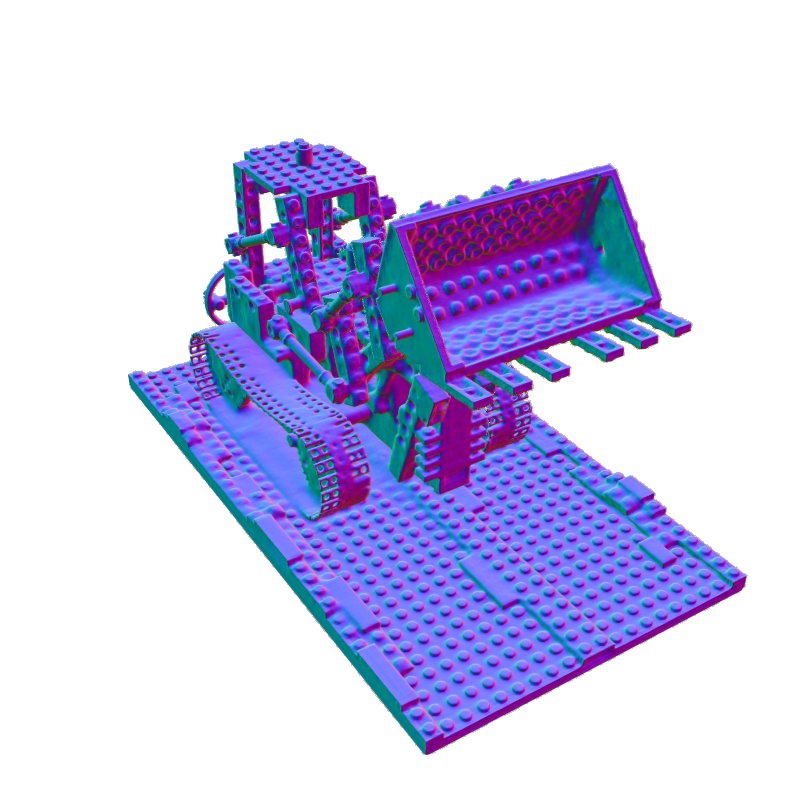} & \adjincludegraphics[width=\resultwidth, trim={0.0in 0.0in 0.0in 0.0in}, clip]{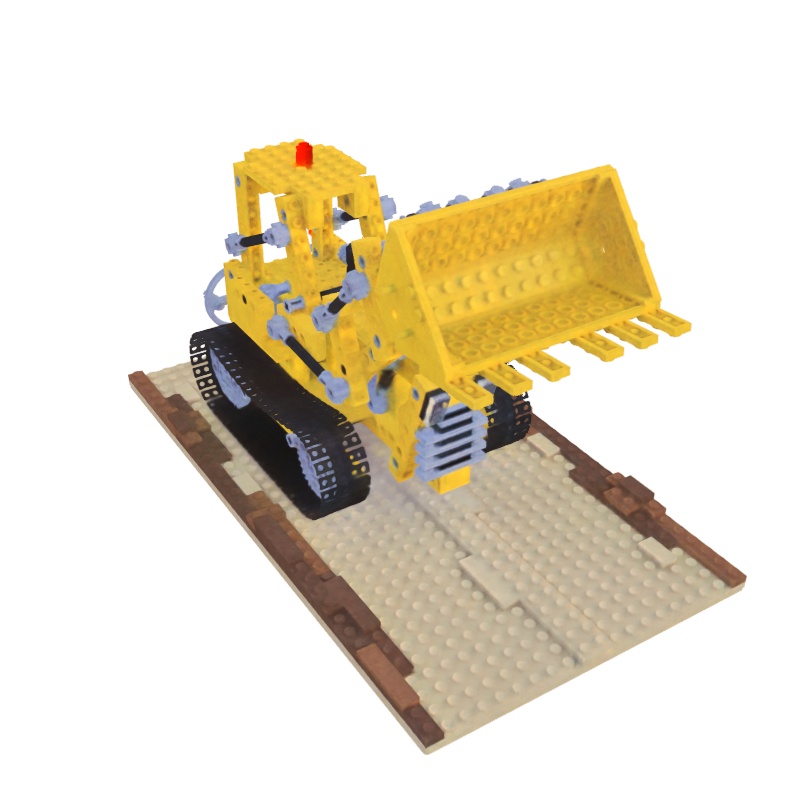} & \adjincludegraphics[width=\resultwidth, trim={0.0in 0.0in 0.0in 0.0in}, clip]{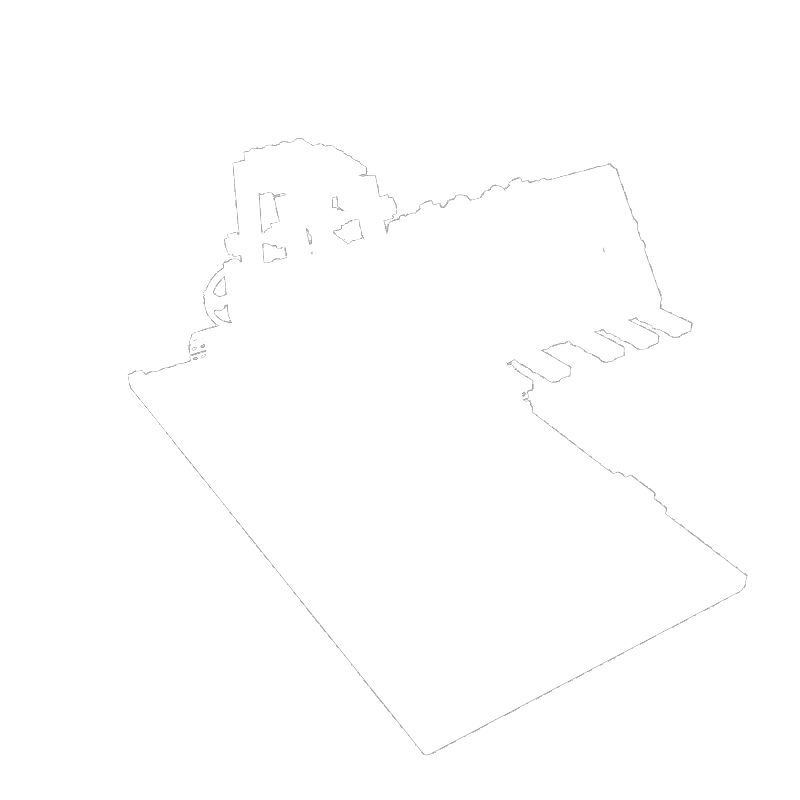} & \adjincludegraphics[width=\resultwidth, trim={0.0in 0.0in 0.0in 0.0in}, clip]{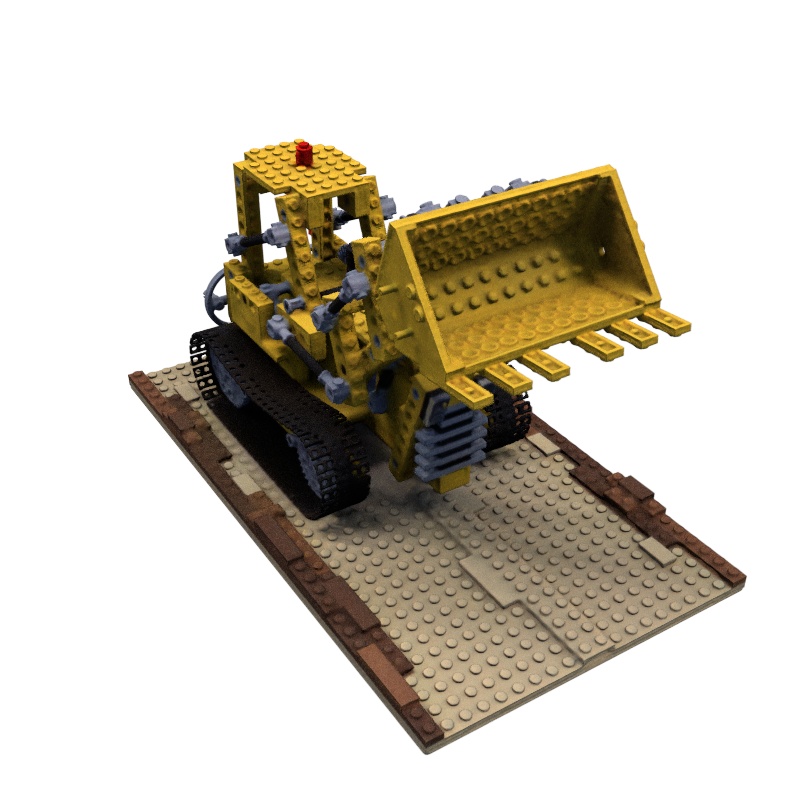} & \adjincludegraphics[width=\resultwidth, trim={0.0in 0.0in 0.0in 0.0in}, clip]{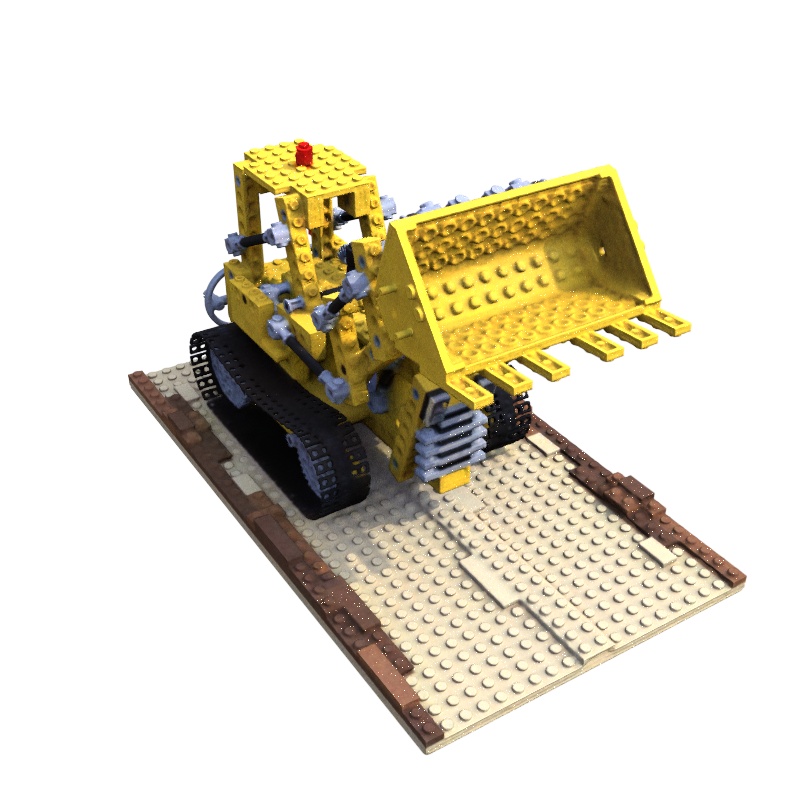} \\

\rotatebox{90}{\,\,\, Multi Light} & \adjincludegraphics[width=\resultwidth, trim={0.0in 0.0in 0.0in 0.0in}, clip]{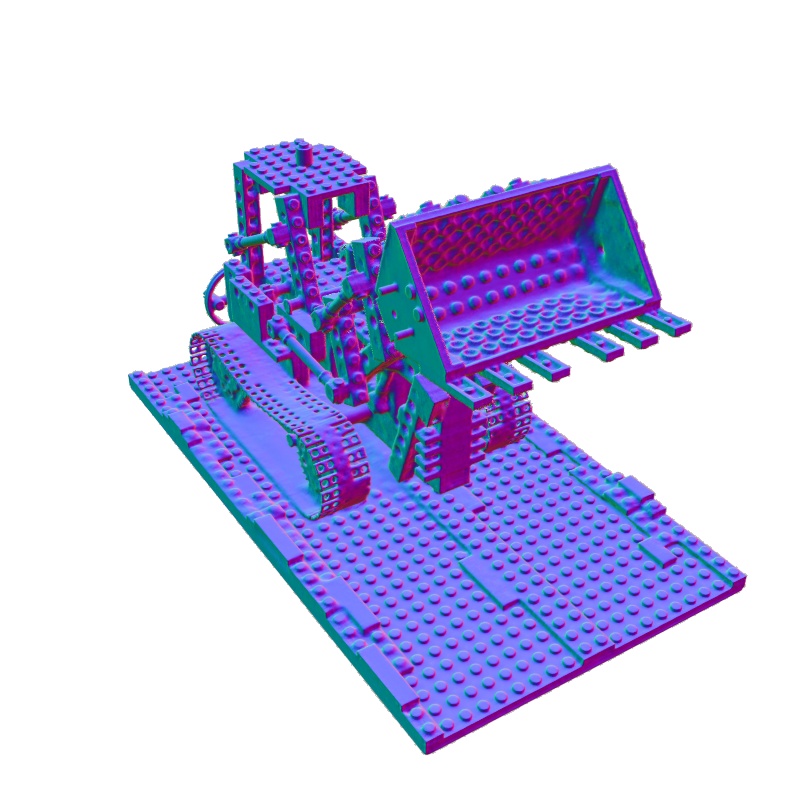} & \adjincludegraphics[width=\resultwidth, trim={0.0in 0.0in 0.0in 0.0in}, clip]{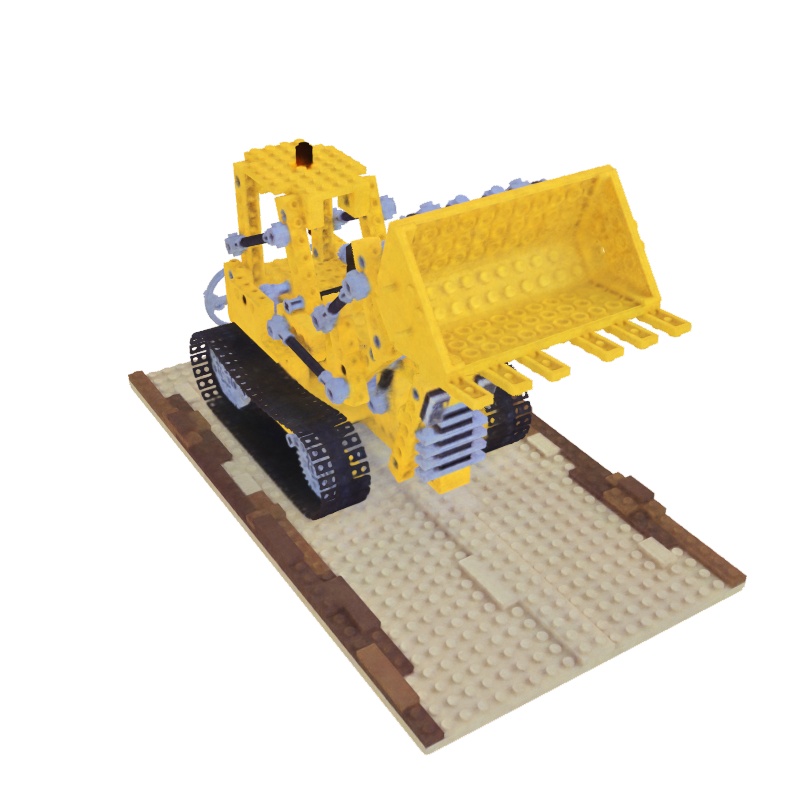} & \adjincludegraphics[width=\resultwidth, trim={0.0in 0.0in 0.0in 0.0in}, clip]{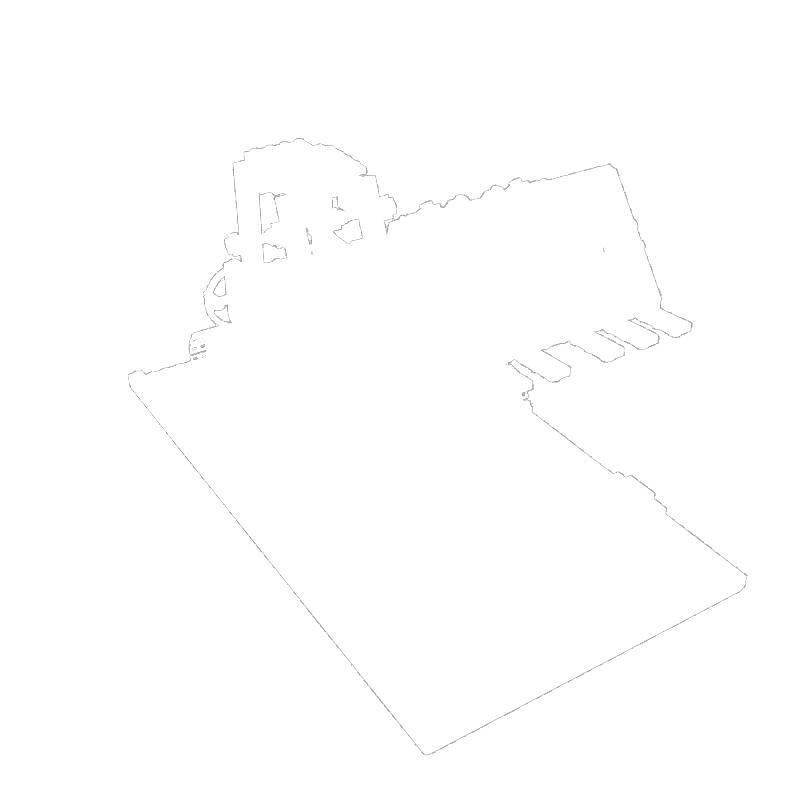} & \adjincludegraphics[width=\resultwidth, trim={0.0in 0.0in 0.0in 0.0in}, clip]{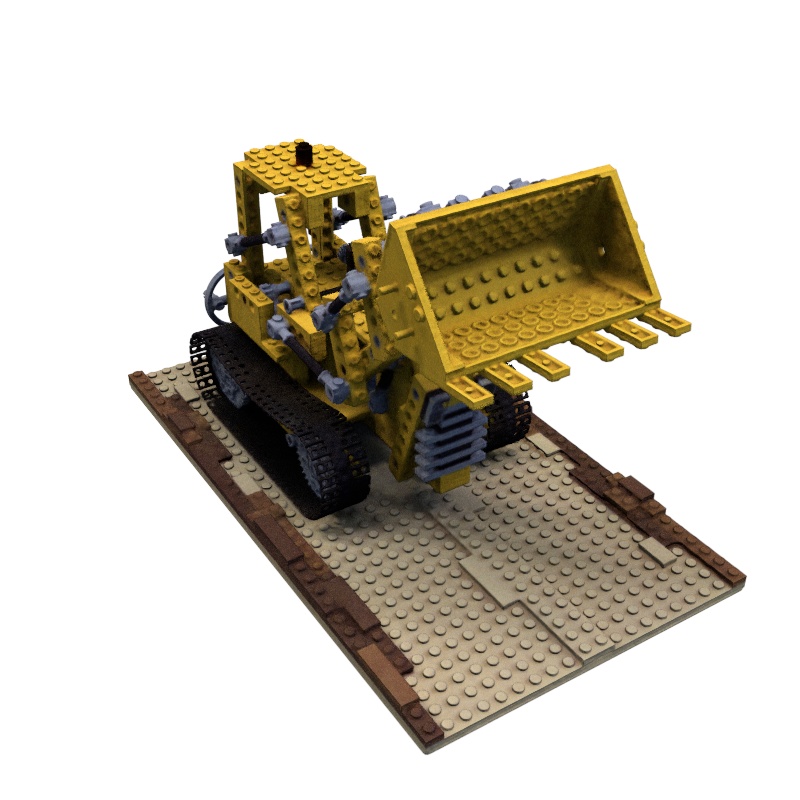} & \adjincludegraphics[width=\resultwidth, trim={0.0in 0.0in 0.0in 0.0in}, clip]{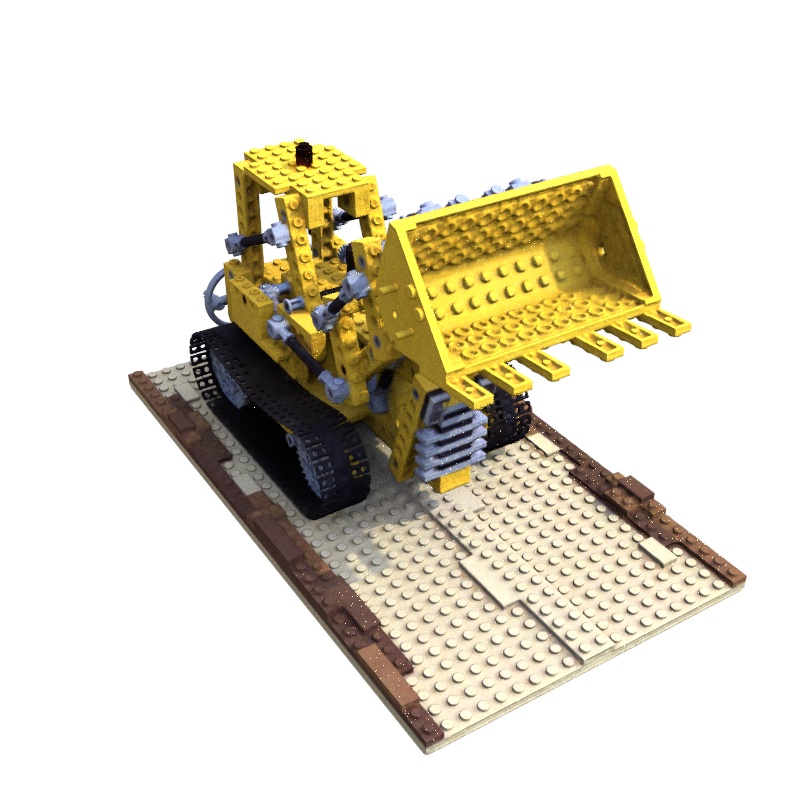} \\

\rotatebox{90}{\,\,\,\,\,\,\,\,\,\, GT} & \adjincludegraphics[width=\resultwidth, trim={0.0in 0.0in 0.0in 0.0in}, clip]{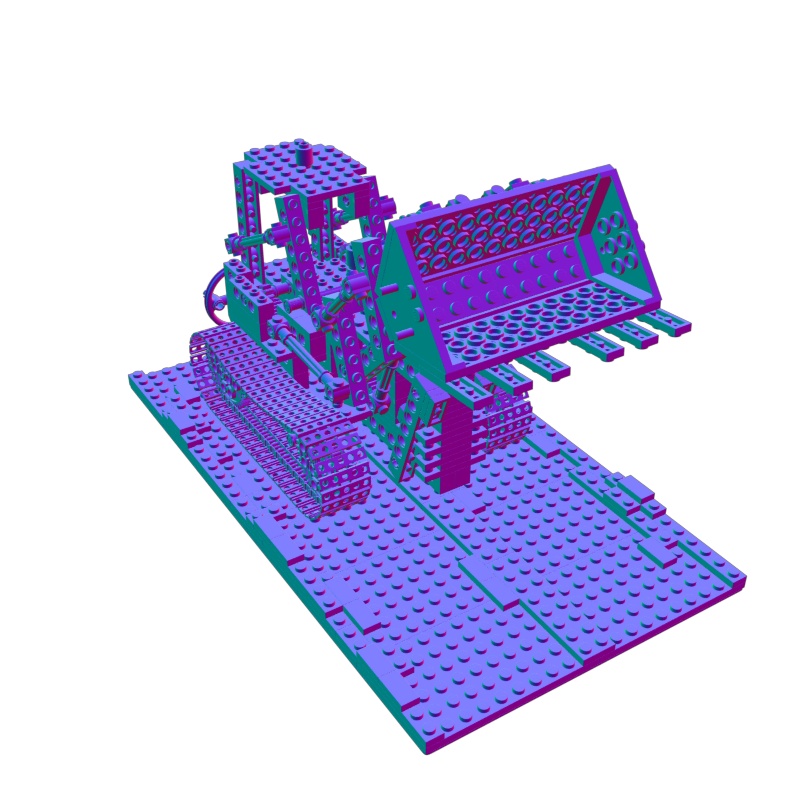} & \adjincludegraphics[width=\resultwidth, trim={0.0in 0.0in 0.0in 0.0in}, clip]{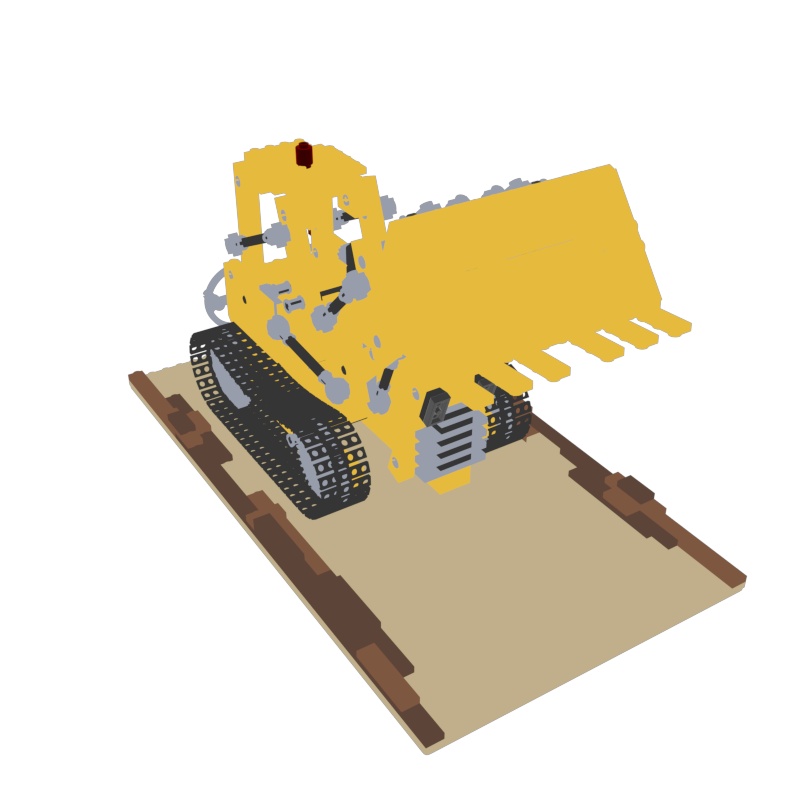} & & \adjincludegraphics[width=\resultwidth, trim={0.0in 0.0in 0.0in 0.0in}, clip]{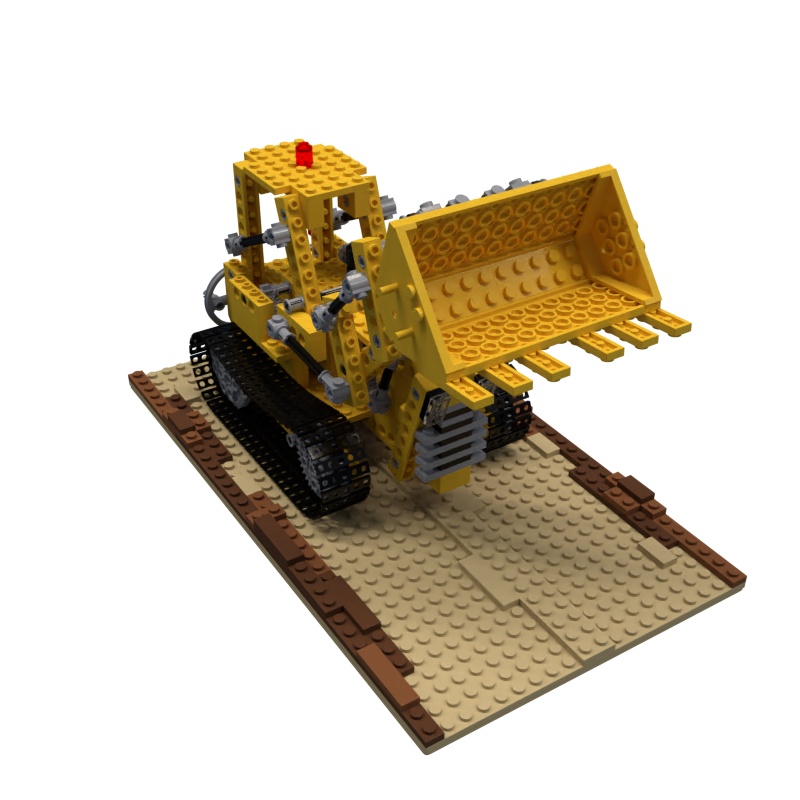} & \adjincludegraphics[width=\resultwidth, trim={0.0in 0.0in 0.0in 0.0in}, clip]{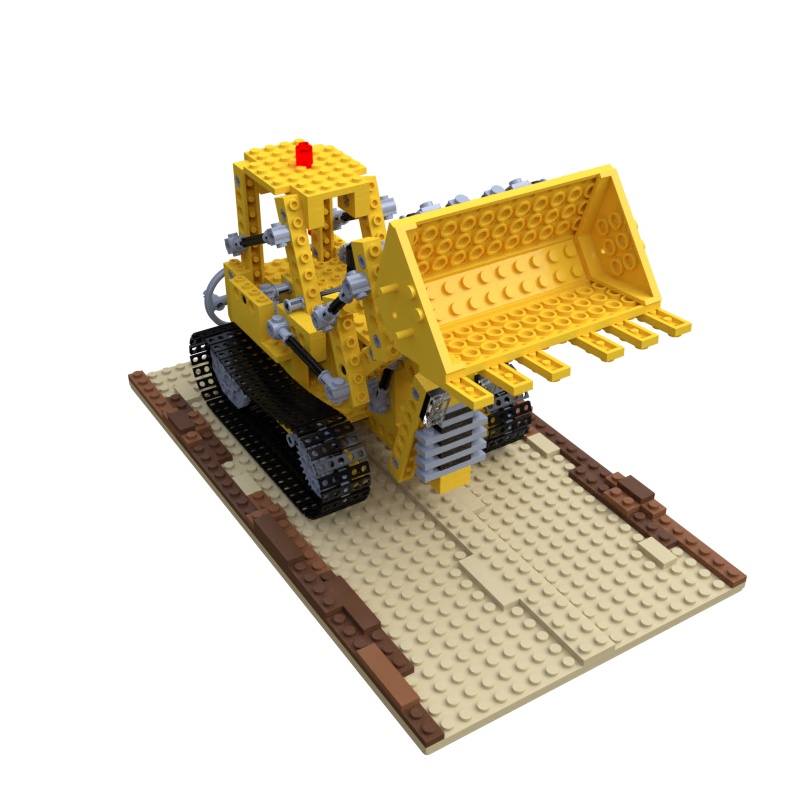}

\end{tabular}
\vspace{-0.1in}
\caption{%
\textbf{Additional TensoIR-Synthetic Results~\cite{jin2023tensoir} Results.}
}%
\label{fig:qual-tensoir-2}
\end{figure*}

\end{document}